\documentclass[11pt,a4paper]{article}
\usepackage[textwidth=6.8in,top=0.65in,bottom=0.65in,headheight=13pt,headsep=12pt]{geometry}
\usepackage{fontspec}
\usepackage{amsmath,amssymb,unicode-math}
\usepackage{graphicx,xcolor,booktabs,array,longtable,listings}
\AtBeginDocument{}
\usepackage{caption,placeins,needspace,titlesec,fancyhdr,enumitem,float}
\usepackage[numbers,square]{natbib}
\usepackage{xurl}
\usepackage[hidelinks,unicode]{hyperref}
\hypersetup{pdftitle={A3P5 NEMESIS Integrated Rover Design for Environmental Reconnaissance and Robotic Sampling with Reproducible Mobility Analysis and an External Data Machine Learning Calibration Benchmark},pdfauthor={Shafi Bin Sultan; Sabik Bin Sultan; Safwan Sadad}}
\titleformat{\section}{\normalfont\fontsize{13}{15}\selectfont}{\thesection}{0.6em}{}
\titleformat{\subsection}{\normalfont\fontsize{12}{14}\selectfont}{\thesubsection}{0.6em}{}
\titlespacing*{\section}{0pt}{10pt plus 2pt minus 2pt}{5pt}
\titlespacing*{\subsection}{0pt}{8pt plus 2pt minus 2pt}{4pt}
\setlist[itemize]{leftmargin=1.5em,itemsep=2pt,topsep=3pt}
\begin{document}
\thispagestyle{fancy}
\begin{center}
{\fontsize{18.5}{22}\selectfont\bfseries A3P5 NEMESIS Integrated Rover Design for Environmental Reconnaissance and Robotic Sampling with Reproducible Mobility Analysis and an External Data Machine Learning Calibration Benchmark\par}
\vspace{7pt}
{\normalsize\bfseries Shafi Bin Sultan\textsuperscript{1,*}\quad Sabik Bin Sultan\textsuperscript{2,*}\quad Safwan Sadad\textsuperscript{3}\par}
\vspace{4pt}
{\small\textsuperscript{1}St. Joseph Higher Secondary School\par
\textsuperscript{2}BAF Shaheen College Kurmitola\par
\textsuperscript{3}Greenland Residential School\par}
\vspace{3pt}
{\footnotesize\textsuperscript{1}\href{mailto:shafibinsultan0207@gmail.com}{shafibinsultan0207@gmail.com}\quad
\textsuperscript{2}\href{mailto:sabikbinsultan@gmail.com}{sabikbinsultan@gmail.com}\quad
\textsuperscript{3}\href{mailto:Avoidsafwan@gmail.com}{Avoidsafwan@gmail.com}\par}
\end{center}
\FloatBarrier

\section*{Abstract}

A3P5 NEMESIS is a four-wheel rover intended to combine remote inspection, environmental observation and lightweight manipulation within one serviceable platform. This study develops a photo-constrained geometric reconstruction, a subsystem architecture and a reproducible analytical assessment while distinguishing physical prototype evidence from proposed functions. An exploratory search retrieved 5,000 bibliographic records across ten queries, yielding 4,897 distinct DOI records and 1,212 metadata candidates; selected primary studies and technical documents informed the design. The reconstructed configuration retains the carbon-pattern enclosure, independently steered wheel assemblies, folded manipulator, inclined camera mast and side sampling equipment. A declared 24 kg scenario predicts 3.28 N\ensuremath{\cdot}m of gearbox-output torque per wheel on a 20\textdegree{} grade under equal load sharing; a separate static model shows how a 2 kg forward payload reduces the geometric front-tipping bound from 38.1\textdegree{} to 32.7\textdegree{}. These are design screens, not measured operating limits. A public-data calibration benchmark uses 7,344 eligible hourly observations, eight sensor/environmental predictors and chronological training, validation and test partitions. Validation-selected ridge regression achieves a held-out CO root-mean-square error of 0.502 mg/m\textsuperscript{3}, with a 95\% daily-block bootstrap interval of 0.435--0.569 mg/m\textsuperscript{3}. This result concerns an external sensor array and cannot establish NEMESIS accuracy. The combined analysis identifies priority measurements, proposed control interfaces and mission-specific validation requirements. The contribution is a traceable engineering design study and evaluation framework for a prototype whose integrated field performance remains to be established.

Keywords: environmental robotics; four-wheel steering; mobile manipulation; low-cost sensing; sensor calibration; temporal validation; inspection rover

\FloatBarrier

\section{Introduction}\label{sec:1}

Environmental inspection requires more than collecting a large number of sensor channels. A mobile platform must place those sensors in a useful location, remain mechanically stable while observing or manipulating a sample, preserve the integrity of each measurement and return information that an operator can interpret. These requirements are coupled. Wheel motion can disturb dust near an air inlet; motors can alter electrical noise and temperature; a manipulator changes the distribution of mass; and communications delays affect the distance needed to stop. A rover may therefore appear capable at the component level while still lacking a defensible system-level operating envelope. The reconstructed configuration is illustrated in Figure 1.

A3P5 NEMESIS---Autonomous Advanced Protection \& Patrol Platform---is motivated by environmental monitoring and remote inspection in settings where access is inconvenient or human exposure should be reduced. The prototype photographs show a rectangular carbon-pattern enclosure, four wheel modules, a folded multi-joint manipulator, an inclined mast and side-mounted sampling equipment. The accompanying design description identifies Arduino Mega hardware control, an ESP32/NodeMCU communication layer, radio control, an 11.1 V nominal battery architecture and several environmental sensing channels. Autonomous mapping, advanced planning and machine-learning interpretation are development objectives rather than demonstrated outcomes of the photographic record.

The central research question is how this existing physical arrangement can support a coherent environmental-inspection architecture without treating missing measurements as established performance. This requires three separations. Geometry reconstructed from photographs must be distinguished from dimensions established by metrology. An engineering calculation under assumed inputs must be distinguished from a test of the assembled vehicle. A model evaluated on a public dataset must be distinguished from a calibration deployed on the rover. Maintaining these boundaries makes the design more useful: readers can identify which conclusions are reproducible now and which depend on future experiments.

The paper makes four contributions. It documents the relationship between the visible prototype and a detailed Blender reconstruction with managed cable routing. It develops compatible low-level control, sensing, communications and protection interfaces, with additions marked where they are proposed. It presents reproducible calculations for wheel coordination, grade demand, static stability, energy, sensor response and stopping clearance. Finally, it evaluates a transparent temporal calibration benchmark on public data and defines a validation programme that connects the analytical results to measurable rover-level outcomes. It does not introduce a new SLAM algorithm or claim a new sensing principle.

The manuscript is organized around these interfaces and evidence levels. The literature search establishes a broad, auditable context; the mechanical sections identify constraints that software must respect; the sensing sections explain why calibration and sampling procedure matter; and the final sections connect control, learning and experimental design. Original analytical plots, published empirical figures and Blender mission illustrations each carry explicit provenance. This supports comparison without suggesting that a rendered mission is an observed deployment.

\begin{figure}[!htbp]
\centering
\includegraphics[width=3.874in,height=4.600in,keepaspectratio]{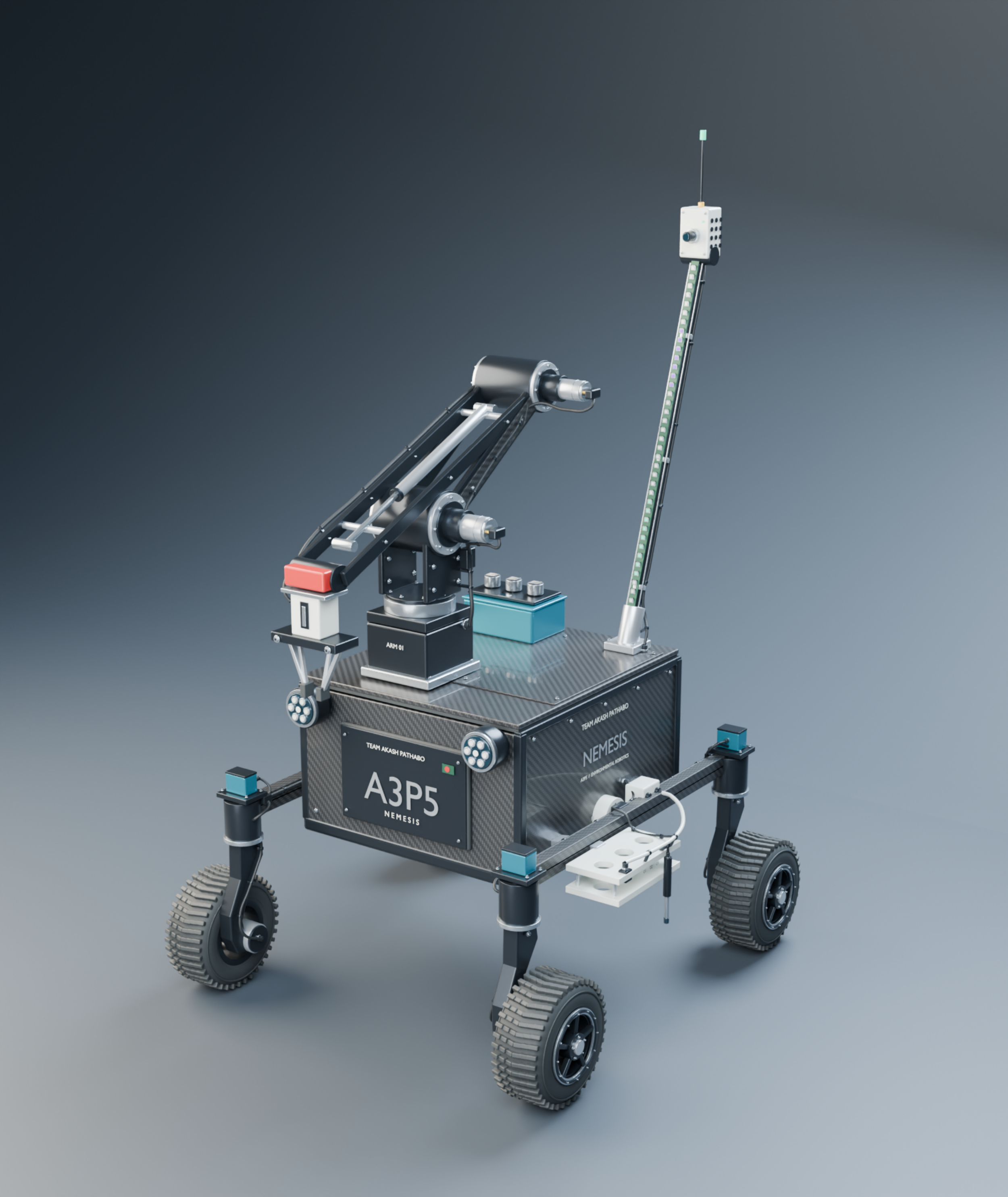}
\caption{Photo-based industrial reconstruction of A3P5 NEMESIS in Blender. The principal body, wheel, manipulator and mast locations follow the supplied photographs. Carbon-pattern panels, fasteners, connector boots and secured cable sleeves represent detailed geometric design. Dimensions and material properties are not experimentally verified.}
\label{fig:1}
\end{figure}

\FloatBarrier

\section{Evidence, literature search and reproducibility}\label{sec:2}

\FloatBarrier

\subsection{Evidence hierarchy}\label{sec:2.1}

The study uses six distinct evidence classes. Photographs establish visible exterior arrangement and appearance. The project description supplies intended hardware roles and mission objectives. The Blender model provides an editable geometric interpretation of the photographs. Published research and manufacturer documents support design principles and calibration requirements. Analytical scenarios produce values from declared inputs. A separate public dataset supports an empirical machine-learning benchmark. No rover test log, measured mass distribution, motor torque curve, sensor co-location record or autonomy trajectory was available for this study.

The distinction between presence and capability is particularly important. A camera in a photograph supports the statement that a camera is mounted; it does not establish visual localization accuracy. A pH probe supports the presence of a measurement interface; it does not establish buffer calibration, temperature compensation or uncertainty. A steering actuator supports a steerable wheel assembly; it does not reveal its angular limits, backlash or closed-loop tracking performance. Accordingly, the term prototype refers to the photographed system, while proposed architecture refers to the implementation developed here.

\FloatBarrier

\begingroup

\fontsize{9.5}{11.2}\selectfont

\renewcommand{\arraystretch}{1.16}

\setlength{\tabcolsep}{5pt}

\begin{longtable}{>{\raggedright\arraybackslash}p{\dimexpr 0.25\linewidth-2\tabcolsep\relax}>{\raggedright\arraybackslash}p{\dimexpr 0.29\linewidth-2\tabcolsep\relax}>{\raggedright\arraybackslash}p{\dimexpr 0.46\linewidth-2\tabcolsep\relax}}
\caption{Evidence sources and permitted interpretations.}\label{tab:1}\\
\toprule
\textbf{Evidence class} & \textbf{Available basis} & \textbf{Permitted interpretation} \\
\midrule
\endfirsthead
\multicolumn{3}{l}{\small Table \thetable\ continued}\\
\toprule
\textbf{Evidence class} & \textbf{Available basis} & \textbf{Permitted interpretation} \\
\midrule
\endhead
\midrule
\multicolumn{3}{r}{\footnotesize Continued on the next page}\\
\endfoot
\bottomrule
\endlastfoot
Prototype appearance & Four distinct photographic viewpoints & Exterior geometry and relative placement \\
Project description & Architecture and mission brief & Reported components and intended functions \\
Geometric reconstruction & Blender assemblies and mission scenes & Spatial arrangement and design illustration \\
Published evidence & Primary papers and technical documents & Source-specific mechanisms, results and limits \\
Analytical results & Declared equations and parameter sweeps & Conditional design screening \\
External-data benchmark & Archived UCI sensor-array observations & Dataset-specific predictive performance \\
\end{longtable}

\endgroup

\FloatBarrier

\subsection{Exploratory metadata census}\label{sec:2.2}

Queries were executed on 15 September 2026, with publication dates restricted to 1 January 1990--15 September 2026. Ten focused Crossref queries covered robotic environmental monitoring, air-quality rovers, gas mapping, particle calibration, gas-sensor calibration, robotic water assessment, four-wheel steering, outdoor mapping and fusion, robotic sampling, and environmental-robotics datasets. Each query retained up to 500 relevance-ranked records. The resulting 5,000 occurrences were normalized by DOI, leaving 4,897 unique records after 103 repeated occurrences were removed. Of the unique records, 1,291 contained an available abstract and 3,606 did not. Missing abstracts were retained as missing rather than inferred from titles. Figures 2 and 3 summarize retrieval and screening counts, publication-year coverage and the availability of abstracts in this sample.

\begin{figure}[!htbp]
\centering
\includegraphics[width=6.522in,height=3.000in,keepaspectratio]{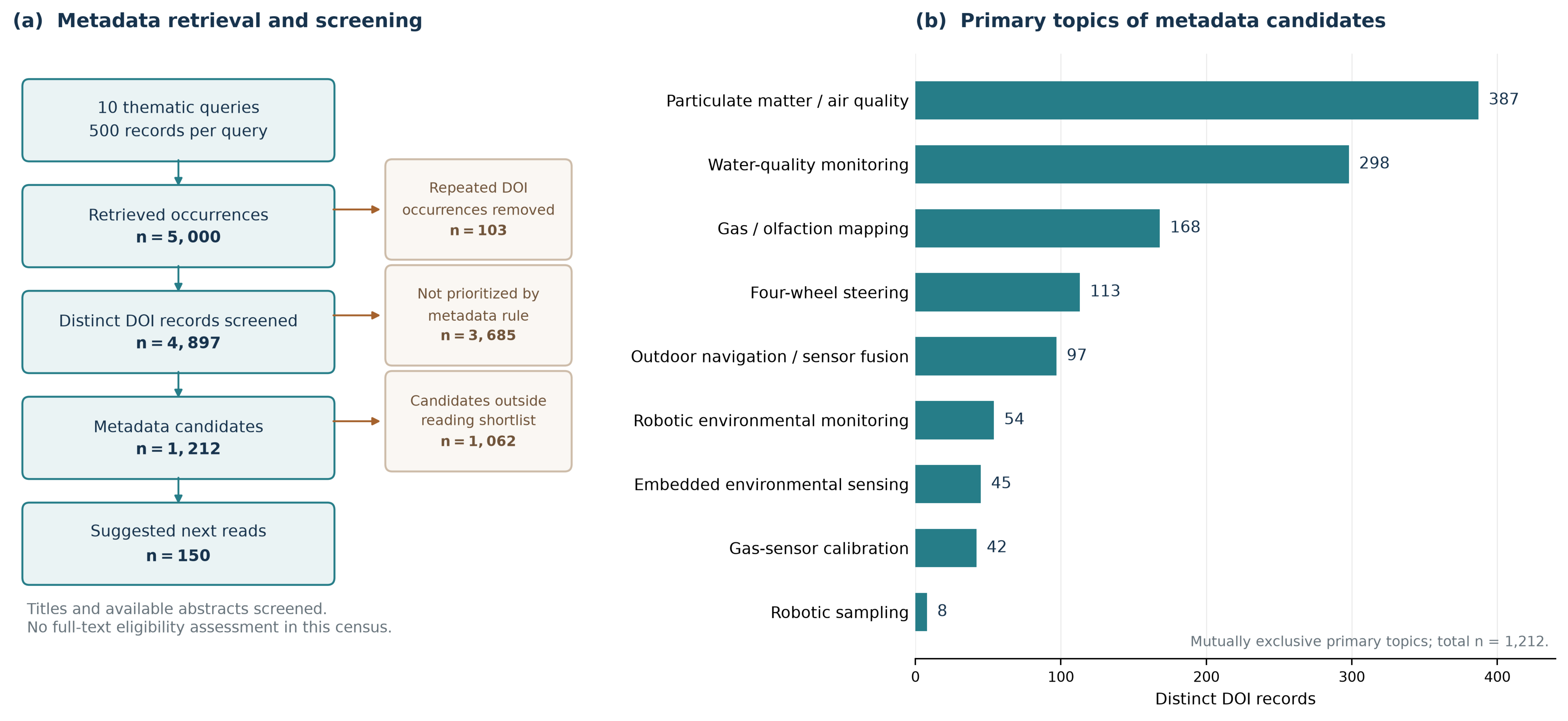}
\caption{Exploratory metadata census and topic distribution of the 1,212 prioritized candidates. Counts describe the retrieved sample after the documented screening rule; they are not field-wide publication totals or a systematic-review flow. Raw requests, timestamps, hashes and screening outputs are retained in the supplement.}
\label{fig:2}
\end{figure}

A documented rule combined topic conjunctions and keyword scores to prioritize 1,212 metadata candidates. The final screening version includes an explicit adjustment for concise four-wheel-steering and robotic-sampling titles that otherwise receive fewer generic keyword matches. It excludes visible administrative material, peer-review reports and retraction/correction notices from the shortlist. A 150-record reading shortlist contains 71 records with abstracts and suppresses exact normalized-title duplicates. This shortlist is a navigation aid, not a declaration that 150 complete papers were critically appraised.

The census is exploratory and is not a systematic review. Search terms, relevance ordering, a fixed per-query cap, Crossref coverage and abstract missingness all affect the retrieved sample. Topic and year counts therefore describe this sample, not the prevalence or growth of the entire research field. No full-text reading is counted in the automated census. The separate source registry records the actual depth of examination for selected papers and technical documents, ranging from publisher metadata or abstracts to inspected full-text sections and original figure captions.

\begin{figure}[!htbp]
\centering
\includegraphics[width=6.800in,height=2.410in,keepaspectratio]{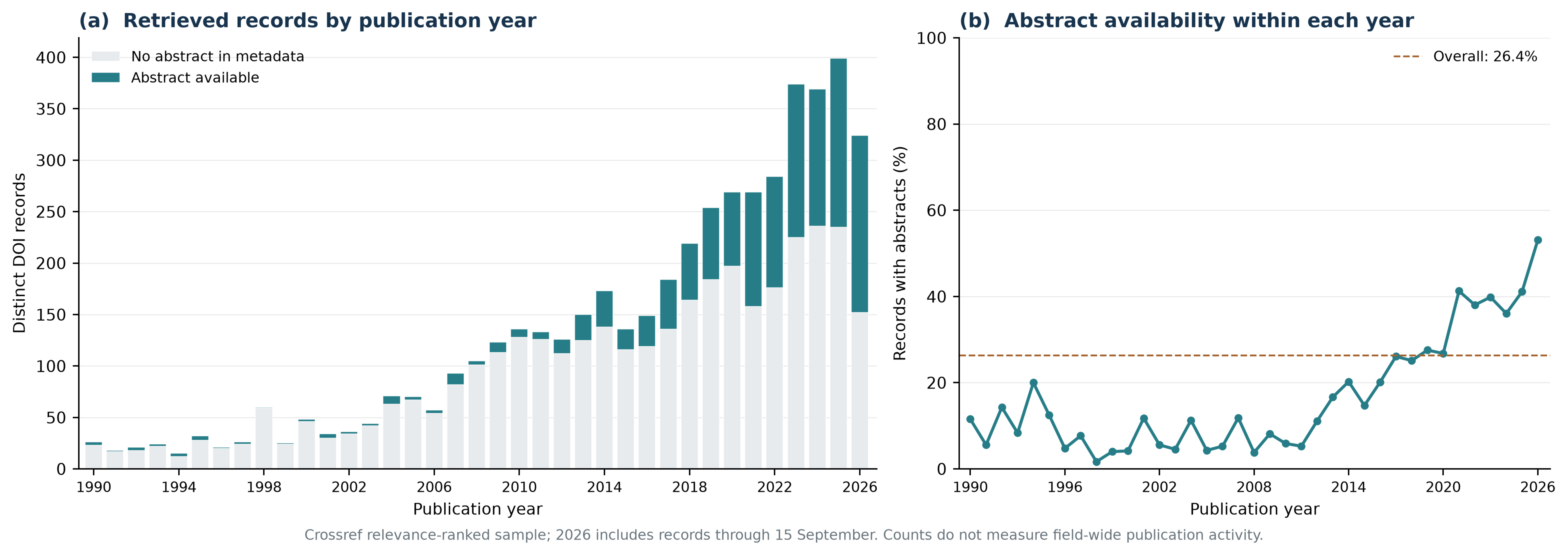}
\caption{Publication-year distribution and abstract availability within the retrieved metadata sample. Missing abstracts limit screening depth. Relevance-ranked retrieval and query caps prevent interpretation as an unbiased estimate of research activity.}
\label{fig:3}
\end{figure}

\FloatBarrier

\subsection{Reproducible artifacts and provenance}\label{sec:2.3}

The supplementary package contains request logs, original response hashes, DOI-level provenance, the retrieval and screening scripts, candidate tables and the reading shortlist. Abstract-presence flags are retained in the distributable metadata; publisher abstract text is not redistributed. The engineering calculations are deterministic parameter sweeps with exported numerical tables. The machine-learning archive contains cleaning decisions, split assignments, model-selection results, frozen fitted pipelines, test predictions and bootstrap draws. Dataset and retrieval files are identified by cryptographic hashes so that later changes to a remote resource do not silently alter the analysis.

All original scientific graphs were generated from the accompanying numerical outputs. The fourteen subsystem diagrams are complete editable Canva pages with figure numbers, titles and arrow legends. Full pages from the Canva PDF export were rendered for publication; editable designs, the source PDF and provenance records are included in the supplementary package. The three-dimensional figures were rendered in Blender from a separately saved mission-study file. Photographic prototype views remain identified as photographs, and reused literature figures retain source attribution and license information in their captions. This separation makes the document auditable at the level of each visual and result.

\FloatBarrier

\section{Related work}\label{sec:3}

\FloatBarrier

\subsection{Scope of comparison}\label{sec:3.1}

NEMESIS combines a four-wheel independently steered chassis, a deck-mounted manipulator, an elevated sensor mast, and air- and water-sensing interfaces within the photographed rover layout. Its relevance is therefore best assessed across mobility, manipulation, environmental measurement, and data interpretation. These areas impose different requirements: a vehicle can navigate successfully while producing poorly calibrated measurements, and a useful sensor can become difficult to interpret when moved through a changing environment. The following comparison separates these requirements and distinguishes the present design from experimentally demonstrated capabilities in prior work.

The review by Francis et al. \citep{francis2022_99d7b7} organizes robotic gas sensing around source localization and distribution mapping, with different platforms and probabilistic approaches serving those objectives. The review by Concas et al. \citep{concas2021_57ba0a} addresses a complementary problem: obtaining dependable information from low-cost air sensors despite cross-sensitivity, changing ambient conditions, and drift. Together, these reviews support treating mobility and measurement as coupled design problems. They do not establish that combining familiar components makes NEMESIS novel or quantitatively reliable. The accompanying metadata census provides a reproducible reading context, while the comparisons below rely on individually checked sources.

\FloatBarrier

\subsection{Steering, terrain interaction, and manipulator stability}\label{sec:3.2}

The wheel-compatibility analysis of Alexander and Maddocks \citep{alexander1989_c3c6fd} provides a foundation for coordinated wheeled motion: wheel orientations and rolling velocities must correspond to a realizable body motion. For four-wheel independent steering and driving, Lee and Li \citep{lee2015_5bfa55} develop a coordinated kinematic, dynamic, and tracking-control formulation, assessed through simulation. NEMESIS shares the need to translate desired chassis motion into compatible steering and drive commands. Its visible steering modules establish a mechanical arrangement, but their existence does not demonstrate steering-angle feedback, synchronized control, or a particular tracking accuracy. These quantities require identification and testing on the assembled platform.

Suspension geometry introduces a separate set of constraints. Reina and Foglia \citep{reina2010_3d6119} propose a variable-camber mechanism for rocker-type planetary-rover suspensions, intended to preserve an upright wheel attitude as the suspension follows uneven terrain. NEMESIS retains its photographed side rails and wheel yokes; it does not reproduce that camber-correction mechanism. The study by Reina and Foglia therefore illustrates how a suspension mechanism can be evaluated against a specific terrain-contact objective. Applying its conclusions to NEMESIS would require identification and testing of an equivalent mechanism. Its mechanism, assumptions, and results should remain distinct from the visual reconstruction.

Traction also depends on contact conditions and applied loads. Iagnemma and Dubowsky \citep{iagnemma2004_9f7418} study terrain-aware traction control, while Papadopoulos and Rey \citep{papadopoulos1996_3a8aa9} develop a tipover-stability measure that incorporates loading and contact geometry for mobile manipulators. These approaches explain why tire appearance and a wide stance are insufficient measures of capability. For NEMESIS, the folded arm, elevated mast, and alternative sampling poses make configuration-dependent loading relevant. A subsequent evaluation should identify component masses, payload positions, wheel contacts, and actuator limits before interpreting calculated stability margins or traction bounds as physical performance.

\FloatBarrier

\subsection{Integrated platforms and task-specific mechanisms}\label{sec:3.3}

Existing integrated robots show that mission requirements strongly influence mechanical architecture. The Karo rescue platform links its tracked, four-flipper arrangement to manipulator design and actuator selection \citep{habibian2021_bdac2f}. ResQbot 2.0 instead incorporates a dedicated casualty-handling system, including an inflatable neck-securing device, and reports a design assessment with simulation-based validation \citep{saputra2021_e81c43}. NEMESIS differs from both: its conventional tires, compact manipulator, and environmental interfaces support a different design scope. Neither rescue platform provides evidence that NEMESIS can climb the same obstacles, handle equivalent loads, or perform casualty extraction. The contrasting rescue and telescopic-leg platforms in Figures 4 and 5 show how different mission requirements produce different mechanical arrangements.

\begin{figure}[!htbp]
\centering
\includegraphics[width=2.598in,height=2.000in,keepaspectratio]{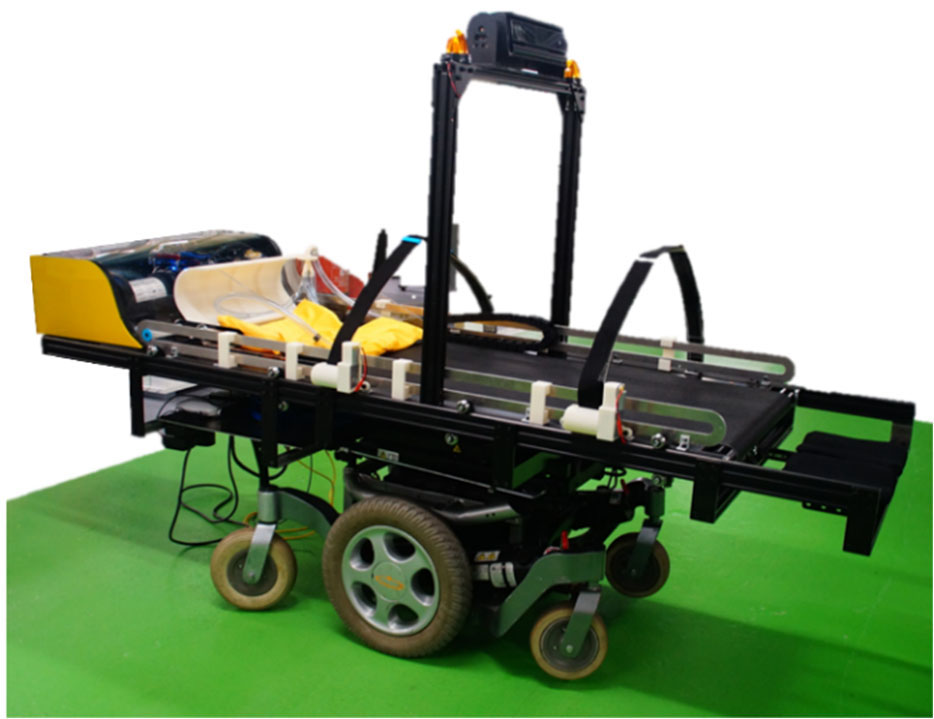}
\caption{ResQbot 2.0 in compact configuration. Reproduced from Saputra et al. \citep{saputra2021_e81c43}, Fig. 10(a), CC BY 4.0; panel isolated from the original composite. Dedicated casualty extraction is a different scope from NEMESIS inspection assistance. Source: \url{https://doi.org/10.3390/app11125414}. Licence: \url{https://creativecommons.org/licenses/by/4.0/}}
\label{fig:4}
\end{figure}

The telescopic-leg robot of Mohamed et al. \citep{mohamed2025_7d0710} offers another instructive contrast because it combines omni-wheels with adjustable leg geometry and documents the transition from design to prototype. Its wheel-contact model and additional mechanisms differ from NEMESIS’s conventional treaded tires and retained layout. The comparison emphasizes the value of documenting mechanical configuration, controller interfaces, and physical assembly together. NEMESIS’s added housings, fasteners, and managed wiring improve the explicitness of its design representation; they remain separate from measurements of stiffness, durability, maneuverability, or service life.

\begin{figure}[!htbp]
\centering
\includegraphics[width=3.204in,height=2.450in,keepaspectratio]{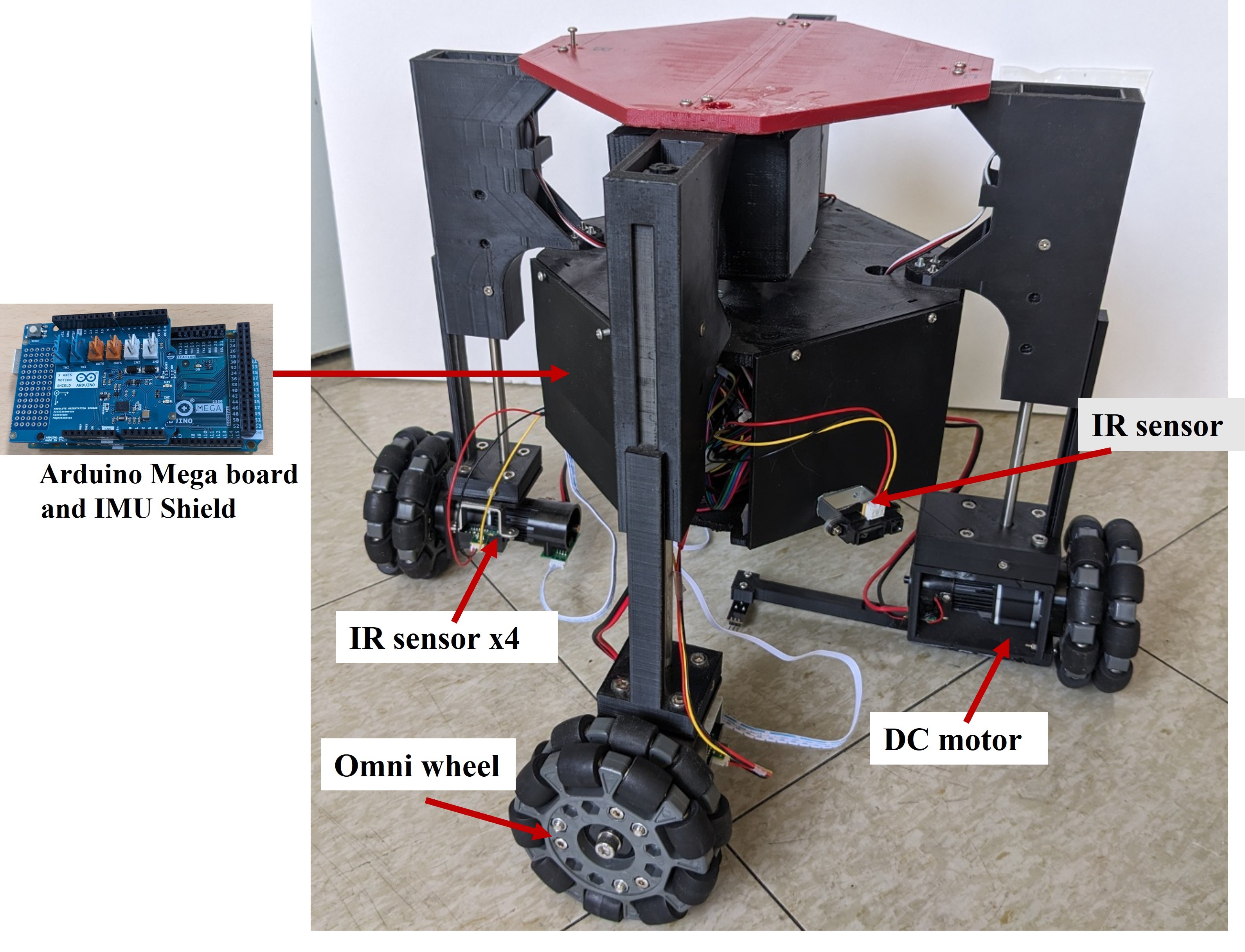}
\caption{Comparator with telescopic legs and omni wheels. Reproduced from Mohamed et al. \citep{mohamed2025_7d0710}, Fig. 1(c), CC BY 4.0; panel isolated from the original composite. This is not A3P5 NEMESIS. Source: \url{https://doi.org/10.3390/machines13040292}. Licence: \url{https://creativecommons.org/licenses/by/4.0/}}
\label{fig:5}
\end{figure}

\FloatBarrier

\subsection{Air sensing and calibration}\label{sec:3.4}

Low-cost gas sensing requires careful interpretation at the component level. Documentation from Zhengzhou Winsen Electronics Technology Co., Ltd. (Winsen) distinguishes the broad response of MQ135 from the methane-sensitive MQ4 and specifies conditioning and heater requirements \citep{winsen2021_36e9d6}, \citep{winsen2021_dd85bc}. NEMESIS should therefore represent these channels as sensor responses pending calibration, rather than assume selective pollutant concentrations from a generic conversion curve. The description identifies an MQ7, whereas the checked carbon-monoxide manual concerns MQ-7B; its operating sequence cannot be assigned to an unidentified board without confirming the variant \citep{winsen2021_174511}. Similarly, the BMP180 supplies pressure and temperature, not relative humidity \citep{boschsensortec2013_0d84c1}.

Particulate measurements have different error mechanisms. Jayaratne et al. \citep{jayaratne2018_d2f9d3} examine humidity and fog effects using PMS1003 sensors. Zheng et al. \citep{zheng2019_9520c4} develop dynamic calibration for a Delhi network that actually used PMS7003 devices, making it especially relevant to NEMESIS’s stated particulate sensor. These studies support sheltered, unobstructed airflow and the collection of environmental covariates, but their sensor configurations and deployment settings differ. NEMESIS’s intake placement should also be evaluated against wheel-generated dust, nearby heat sources, and water-handling activity; the current arrangement has not been tested for these interactions.

Further studies show why a single published correction should not be treated as universal. Barkjohn et al. \citep{barkjohn2021_7ba5aa} evaluate a PurpleAir correction using held-out dates and geographic regions, and Patel et al. \citep{patel2024_4264a8} investigate hygroscopic-growth calibration for low-cost particulate sensing. A PMS7003 development and testing study also examines laboratory calibration and extended field operation \citep{bthory2022_c26801}. The practical comparison for NEMESIS is thus between transparent candidate corrections evaluated with its own reference observations. Raw measurements, correction versions, maintenance history, and validity flags should remain traceable so that apparent improvements can be checked across concentrations and environmental conditions.

\FloatBarrier

\subsection{Water sampling and the measurement interface}\label{sec:3.5}

Robotic access to water is not equivalent across platform types. The airboat developed by Melo et al. \citep{melo2019_023727} provides a precedent for online water-quality monitoring from an aquatic vehicle. NEMESIS is a ground rover, so its intended pH and turbidity functions depend on an accessible sampling position and a defined interface between the instrument and water. The manipulator and holders can represent how a probe is presented, stored, and serviced, but do not establish immersion depth, sampling reach, watertightness, or the absence of contamination between samples.

Turbidity research makes this interface dependence explicit. Trevathan et al. \citep{trevathan2020_3d3a03} investigate affordable light-attenuation sensing, including calibration against reference standards and practical enclosure considerations. Wang et al. \citep{wang2024_112661} examine a compact stormwater sensor and the effects of cleaning and deployment on its behavior. These are sensor-specific demonstrations, not validation of an unidentified NEMESIS turbidity module. For pH, the USGS measurement protocol emphasizes calibration, maintenance, stabilization, and quality assurance \citep{usgs2021_e859f3}. The appropriate comparison is therefore an entire sampling procedure, including standards, rinsing, and repeated observations, rather than the presence of a probe or a displayed numerical value.

\FloatBarrier

\subsection{Localization and environmental mapping}\label{sec:3.6}

Environmental mapping adds requirements beyond collecting sensor readings. Gongora et al. \citep{gongora2023_ca6a1d} combine gas and wind information with action selection based on expected information gain, supported by simulation and controlled experimental evaluation. NEMESIS could provide a physical platform for related methods, but a comparable implementation would need synchronized localization, an identified wind-measurement channel, and suitable computation. Its angled mast establishes a location for instruments; it does not by itself supply the pose accuracy, airflow observations, or measurement model needed to infer a concentration field.

Simulation can support that development when its scope is explicit. GADEN models gas dispersion and sensor responses in three-dimensional environments \citep{monroy2017_e93884}, while the VGR dataset provides simulated airflow and gas dispersion in detailed house models \citep{ojeda2023_b7b9b6}. These resources can support repeatable algorithm comparisons, but their simulated environments differ from outdoor rover operation. A Blender scenario illustrates component placement and mission intent. It becomes a transport simulation only when geometry, boundary conditions, physical models, and numerical procedures are separately specified and evaluated; visual plume effects should therefore remain illustrative.

Localization frameworks address another part of the problem. ORB-SLAM3 supports visual and visual-inertial mapping, while RTAB-Map provides lidar and visual mapping approaches \citep{campos2021_973bd6}, \citep{labb2019_d34c3f}. Either would require sensors, calibration, and processing resources appropriate to the selected configuration. NEMESIS’s described microcontroller-based control architecture does not establish that these stacks have been integrated. The relevant next comparison is whether a chosen implementation can produce sufficiently reliable, timestamped poses for the intended sampling task, rather than whether a rendered rover carries a camera or a proposed lidar housing.

\FloatBarrier

\subsection{Data analysis, external benchmarks, and validation}\label{sec:3.7}

Public datasets can make algorithm development reproducible without substituting for rover evidence. The UCI Air Quality dataset provides an external multisensor record \citep{vito2008_026a72}, associated with an electronic-nose calibration study that used reference observations for benzene estimation \citep{devito2008_1548b8}. These data are useful for checking preprocessing, regression, and evaluation procedures. Their sensor chemistry, installation, sampling history, and reference measurements differ from NEMESIS. Any experiment using them should be labeled an external-data benchmark; its prediction errors cannot be reported as calibration accuracy or detection performance of the rover.

Candidate analysis methods should also have clear roles. Ridge regression provides a regularized linear baseline for correlated predictors \citep{hoerl1970_31b919}, while random forests offer a nonlinear tree-ensemble baseline \citep{breiman2001_75dc04}. Gaussian-process regression provides conditional predictions and model-based uncertainty \citep{rasmussen2006_5c445c}. These alternatives should be compared using the same deployment-relevant partitions and reference targets. Anomaly detection addresses a different question: Isolation Forest identifies unusual observations, which may include faults, changing environmental conditions, or genuine events \citep{liu2012_0ef0af}. An anomaly score alone does not identify a pollutant or establish a hazardous condition.

Evaluation design is particularly important for repeated measurements. The methodological review by Roberts et al. \citep{roberts2017_8ac43f} explains why temporal, spatial, and hierarchical dependence should inform cross-validation. Kapoor and Narayanan \citep{kapoor2023_8dd715} examine leakage pathways that can make scientific machine-learning results appear more successful than their evaluation supports. For NEMESIS, neighboring readings from one mission should not automatically be treated as independent examples of generalization. Preprocessing and candidate models should be fitted on training data; configuration selection belongs on a separate validation partition within the development data. Held-out missions, days, sites or instruments should remain untouched until the final evaluation and should reflect the deployment claim. External-data results and future rover measurements should remain clearly distinguishable.

\FloatBarrier

\subsection{Position of the present design}\label{sec:3.8}

Taken together, the literature supports a staged assessment of NEMESIS: establish coordinated motion and load limits, characterize the installed sensors and sampling interfaces, then evaluate localization and data interpretation under defined missions. The NIST response-robot program provides a useful task-based framework spanning mobility, sensing, communications, manipulation, and energy-related performance \citep{nist2026_a8b13c}. It supplies a structure for selecting repeatable tests, not a certification conferred by visual similarity. The present model preserves the rover’s recognizable architecture and makes its component interfaces inspectable. Its contribution should be described at that design level until measured evidence supports stronger mechanical, sensing, or autonomous-performance claims.

\FloatBarrier

\section{Mechanical reconstruction and system architecture}\label{sec:4}

\FloatBarrier

\subsection{Geometry retained from the prototype}\label{sec:4.1}

The reconstruction preserves the arrangement that gives NEMESIS its identity: the central enclosure sits above four outboard wheel modules, the manipulator occupies the forward upper deck, and the camera mast rises obliquely from the rear of the deck. The arm remains folded in the photographed triangular configuration. The sampling rack remains in its photographed side position, and the blue enclosure remains behind the arm. No solar panel is included. The tall linear mechanism visible in some photographs is retained in an optional collection rather than silently merged into every configuration. Figure 6 collects the four photographic views used to constrain the reconstruction.

Four distinct photographs were used for geometric interpretation. The front and oblique views constrain the relative body, wheel and manipulator arrangement; the elevated view constrains the top deck and mast base; and the alternate side configuration helps identify sampling and optional lifting equipment. Perspective, lens distortion, occlusion and the absence of a scale reference prevent a unique dimensional solution. The mesh is therefore a visual engineering reconstruction, not a photogrammetric survey or a manufacturing drawing.

\FloatBarrier

\begingroup

\fontsize{9.5}{11.2}\selectfont

\renewcommand{\arraystretch}{1.16}

\setlength{\tabcolsep}{5pt}

\begin{longtable}{>{\raggedright\arraybackslash}p{\dimexpr 0.25\linewidth-2\tabcolsep\relax}>{\raggedright\arraybackslash}p{\dimexpr 0.29\linewidth-2\tabcolsep\relax}>{\raggedright\arraybackslash}p{\dimexpr 0.46\linewidth-2\tabcolsep\relax}}
\caption{Geometric quantities and analytical assumptions.}\label{tab:2}\\
\toprule
\textbf{Geometric quantity} & \textbf{Model value} & \textbf{Status and use} \\
\midrule
\endfirsthead
\multicolumn{3}{l}{\small Table \thetable\ continued}\\
\toprule
\textbf{Geometric quantity} & \textbf{Model value} & \textbf{Status and use} \\
\midrule
\endhead
\midrule
\multicolumn{3}{r}{\footnotesize Continued on the next page}\\
\endfoot
\bottomrule
\endlastfoot
Wheelbase L & 0.908 m & Estimated coordinate separation for calculations \\
Track W & 0.930 m & Estimated wheel-centre separation \\
Rolling radius r & 0.140 m & Approximate model radius, real loaded radius unmeasured \\
Enclosure width \ensuremath{\times} length & 0.600 \ensuremath{\times} 0.660 m & Photo-based model geometry \\
Overall width \ensuremath{\times} length & 1.086 \ensuremath{\times} 1.190 m & Configuration-dependent model envelope \\
Overall height & Approximately 1.95 m & Includes antenna in reconstructed pose \\
Base mass and CG height & 24 kg and 0.58 m & Analytical assumptions, not measurements \\
\end{longtable}

\endgroup

\begin{figure}[!htbp]
\centering
\includegraphics[width=3.536in,height=5.500in,keepaspectratio]{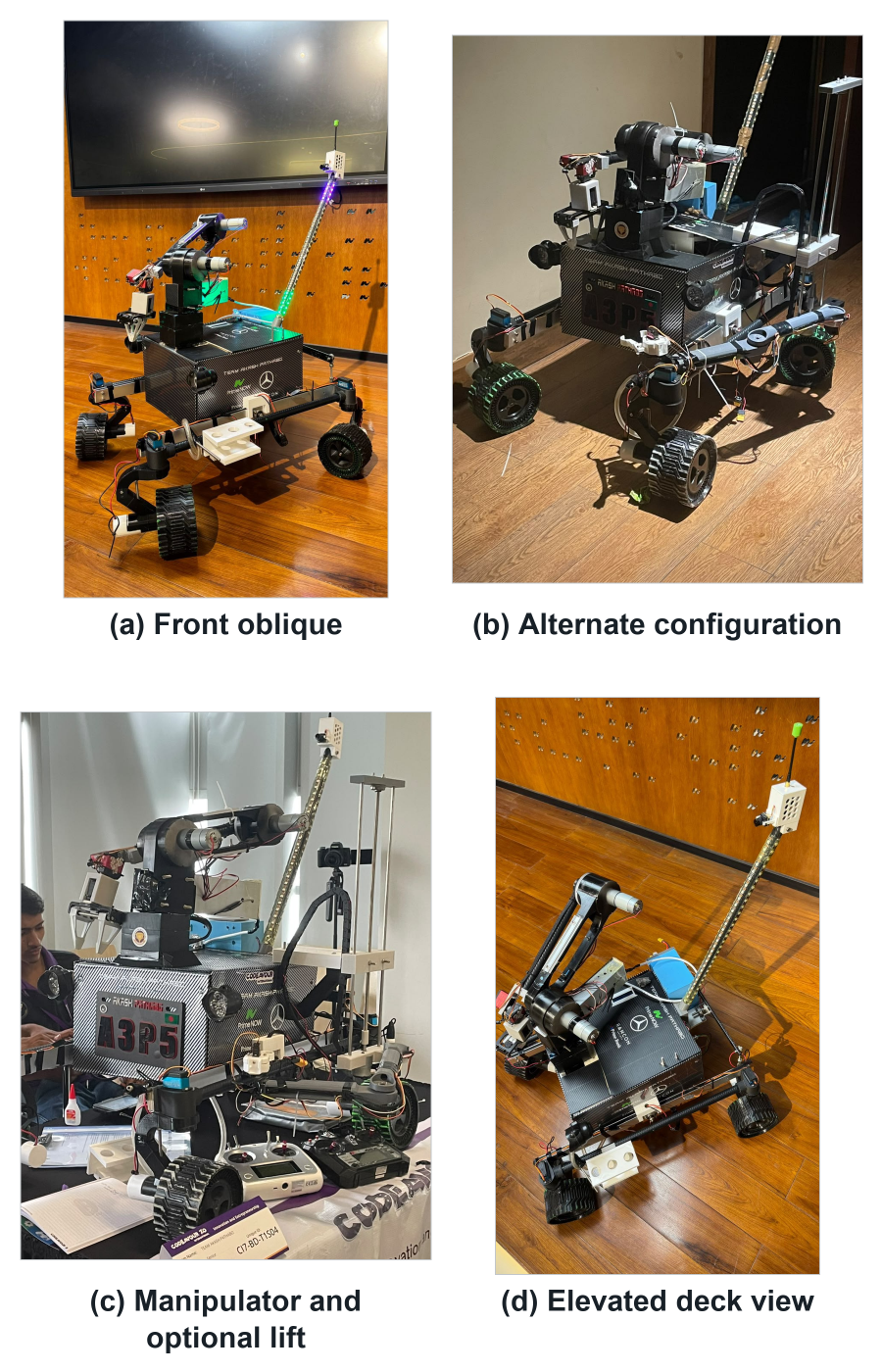}
\caption{Four distinct prototype photographs used for geometric reconstruction. The views constrain component arrangement and show an optional lift configuration; dimensions remain estimated without a scale reference.}
\label{fig:6}
\end{figure}

The analytical origin is the ground-plane projection of the chassis centre on a level reference support plane, with x forward, y left and z upward. Centre-of-mass height is measured above that support plane. The Blender asset uses front along negative Y, with the sampling rack on its positive-X side. A right-handed analytical frame maps to Blender as (X, Y, Z) = (y, \ensuremath{-}x, z). This frame conversion prevents a common integration error: a mathematically correct wheel command can still steer the wrong module if drawing, controller and navigation frames are inconsistent. Mechanical drawings and firmware should use explicit frame names and a verified sign convention.

\FloatBarrier

\subsection{Industrial detailing and cable management}\label{sec:4.2}

The visual refinement introduces fasteners, panel-edge protectors, bearing and joint housings, connector collars, cable sleeves and support brackets while maintaining the principal component positions. The carbon appearance is implemented as a surface material. Neither a rendered weave nor a black panel establishes a carbon-fibre laminate lay-up, stiffness, impact resistance or electrical insulation. Structural verification requires actual panel material, thickness, fastener spacing and load paths. The enclosure should be treated as a protective shell around a supporting frame until those properties are known. The secured sleeves, cable supports and connection details are visible in Figure 7.

The wiring revision replaces unsupported long runs with restrained branches along the chassis rails, steering housings, arm links and mast. The model includes 39 cable or hose curves and 62 mounted clamp assemblies. Connector boots provide visual strain relief; short service loops are placed near moving interfaces. Clamp collars were aligned with the cable centreline, and mounting feet were terminated against supporting meshes. These checks establish contact in the reconstructed static pose. They do not certify bend radius, fatigue life, connector retention or clearance over the full range of joint motion.

\begin{figure}[H]
\centering
\includegraphics[width=4.810in,height=4.300in,keepaspectratio]{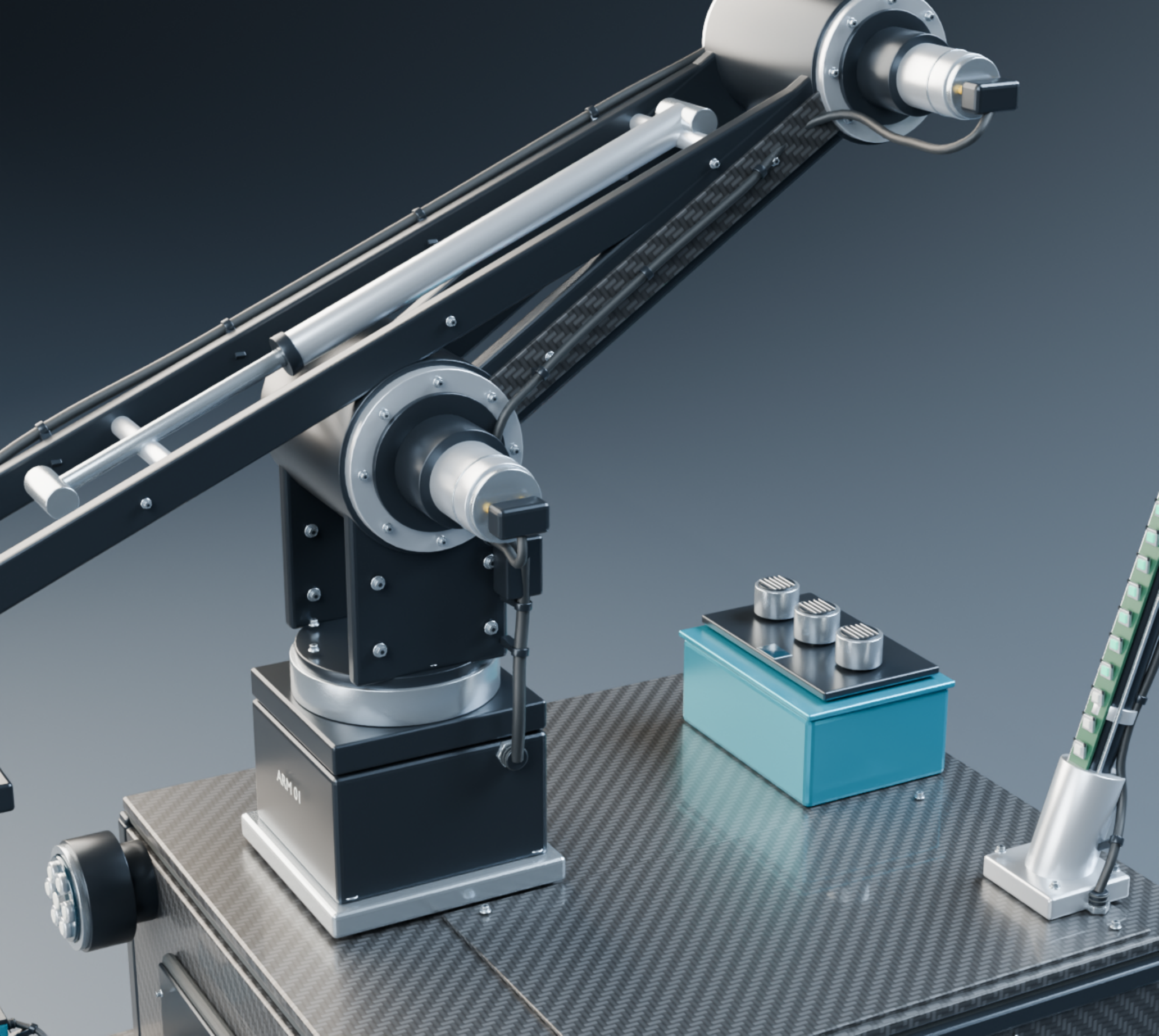}
\caption{Managed wiring in the Blender reconstruction. Sleeves follow structural members, connector boots support cable entry and clamps attach branches to nearby structure. The routing is checked for the displayed static configuration; steering and arm articulation require a full-travel cable-clearance and fatigue assessment.}
\label{fig:7}
\end{figure}

A buildable harness should divide power, motor switching, servo supply, digital communications and high-impedance analog sensing into named branches. Each branch needs a documented connector pinout, conductor size, current limit and termination. Shielding and return routing should be selected from the actual signal bandwidth and noise environment. The pH lead is especially sensitive to leakage and interference; a visually tidy cable is not sufficient if its analog interface is electrically unsuitable. Detachable panels need a service loop or disconnect that allows maintenance without pulling a sensor connector.

\FloatBarrier

\subsection{Functional partition}\label{sec:4.3}

The proposed architecture assigns time-critical acquisition and actuator supervision to the low-level controller. The communication processor handles local networking and telemetry. Computationally demanding localization, mapping or model evaluation belongs on a proposed companion processor unless an implementation-specific resource assessment demonstrates otherwise. This partition avoids assuming that a familiar development board can execute an entire autonomous-robotics stack simply because it can read sensors and drive motors. Figure 8 summarizes the proposed functional interfaces between measurement, control, communications and operator supervision.

Commands should cross these boundaries as bounded requests rather than raw, unqualified actuator values. A velocity request should identify its source, sequence number, timestamp, expiration time and motion mode. The receiving controller should validate limits and report what it actually applied. Sensor records should similarly identify units, validity, calibration version and acquisition time. This permits software replacement without changing the mechanical layout, while keeping the low-level controller responsible for rejecting stale or infeasible commands.

\begin{figure}[!htbp]
\centering
\includegraphics[page=1,width=6.800in,height=4.387in,keepaspectratio]{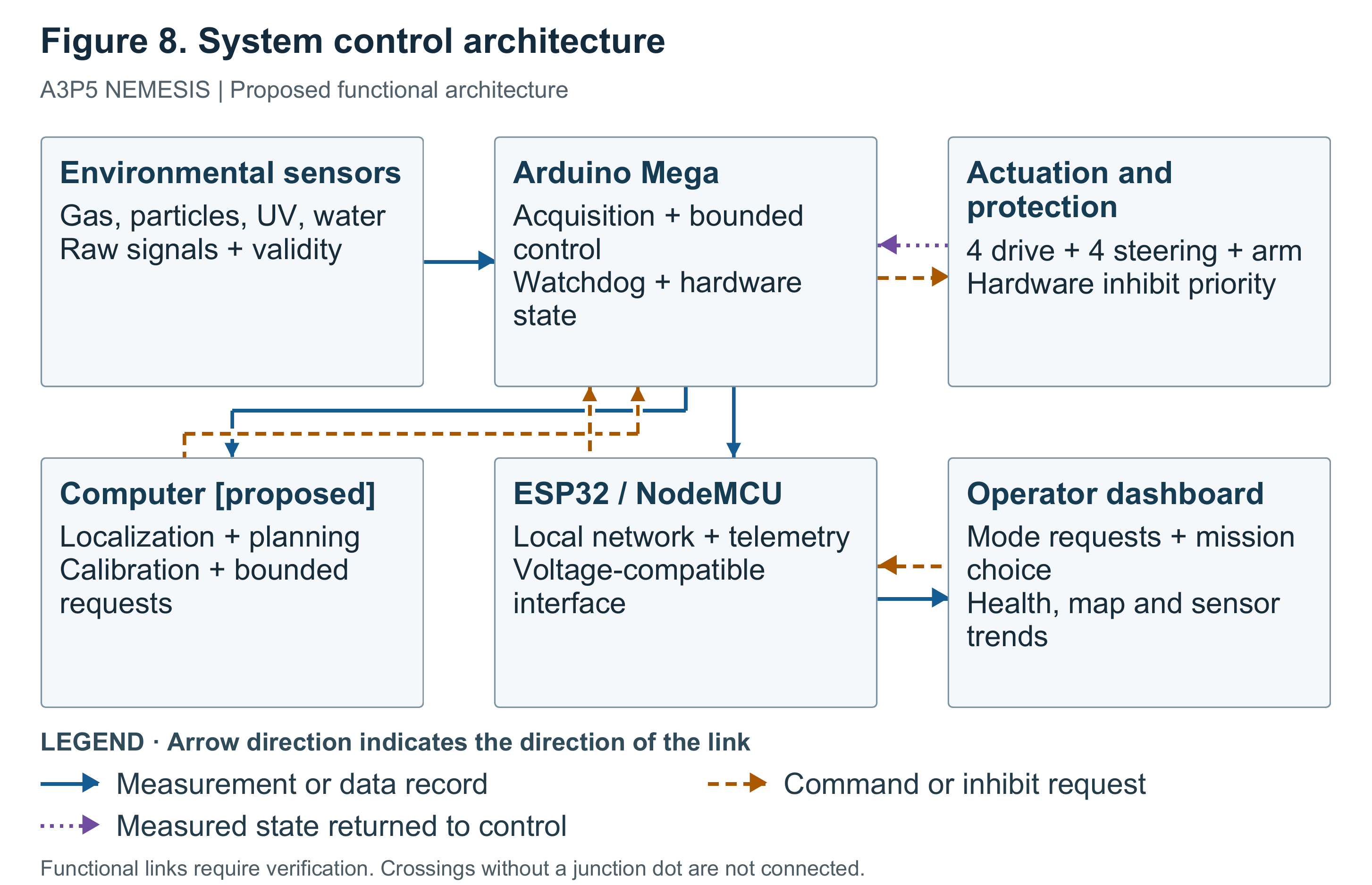}
\caption{Proposed system architecture connecting environmental measurement, hardware control, communications, operator supervision and high-level computation. Functional links summarize interfaces; detailed control authority is specified in the text.}
\label{fig:8}
\end{figure}

\FloatBarrier

\section{Mobility, traction and steering coordination}\label{sec:5}

\FloatBarrier

\subsection{Compatible wheel velocities}\label{sec:5.1}

Four independently steered wheels can support several motion modes, but their commands must be mutually compatible. Under a locally planar, rigid-body, no-slip approximation, the velocity at each wheel follows from the body translation and yaw rate.

Let wheel i lie at (\(x_i\), \(y_i\)), with body velocity components \(v_x\) and \(v_y\) and yaw rate \ensuremath{\omega}. The wheel velocity components and corresponding steering angle are given below. The formulation is consistent with the wheel-compatibility treatment of Alexander and Maddocks \citep{alexander1989_c3c6fd} and the coordinated architecture discussed by Lee and Li \citep{lee2015_5bfa55}.

\begin{equation}
\label{eq:1}
u_i=v_x-\omega y_i,\qquad w_i=v_y+\omega x_i
\end{equation}

\begin{equation}
\label{eq:2}
\delta_i=\operatorname{atan2}(w_i,u_i),\qquad \Omega_i=\frac{\sqrt{u_i^2+w_i^2}}{r_i}
\end{equation}

Here \(\delta_i\) is wheel heading, \(\Omega_i\) is angular wheel speed and \(r_i\) is loaded rolling radius. The equivalent command (\(\delta_i\) + \ensuremath{\pi}, \ensuremath{-}\(\Omega_i\)) may reduce steering travel, but only if the measured mechanism and cable routing permit it. At near-zero wheel velocity the heading is numerically indeterminate; the controller should retain a previous feasible angle or execute an explicit reorientation state. Small measurement noise must not cause an arbitrary steering reversal. Figures 9 and 10 connect the proposed wheel-command allocation with the calculated steering geometry.

Straight motion sets lateral velocity and yaw rate to zero. Crab motion uses a common heading with zero yaw rate. A coordinated turn requires every wheel velocity to be tangent to the same instantaneous rigid-body motion. Physical point-turn feasibility depends on verified steering travel, cable clearance, tire contact and actuator torque. Large heading changes should initially occur with traction inhibited or strongly limited, followed by a controlled speed ramp after convergence.

At 0.25 m/s and radius 0.140 m, ideal wheel speed is approximately 17.05 rpm. This is an output-wheel requirement, not a motor-shaft specification. Gear ratio, controller resolution and encoder counts must support observable low-speed motion. A motor with ample stall torque can still be unsuitable if it overheats during slow continuous operation or provides inadequate speed resolution near the operating point.

\begin{figure}[!htbp]
\centering
\includegraphics[page=3,width=6.355in,height=4.100in,keepaspectratio]{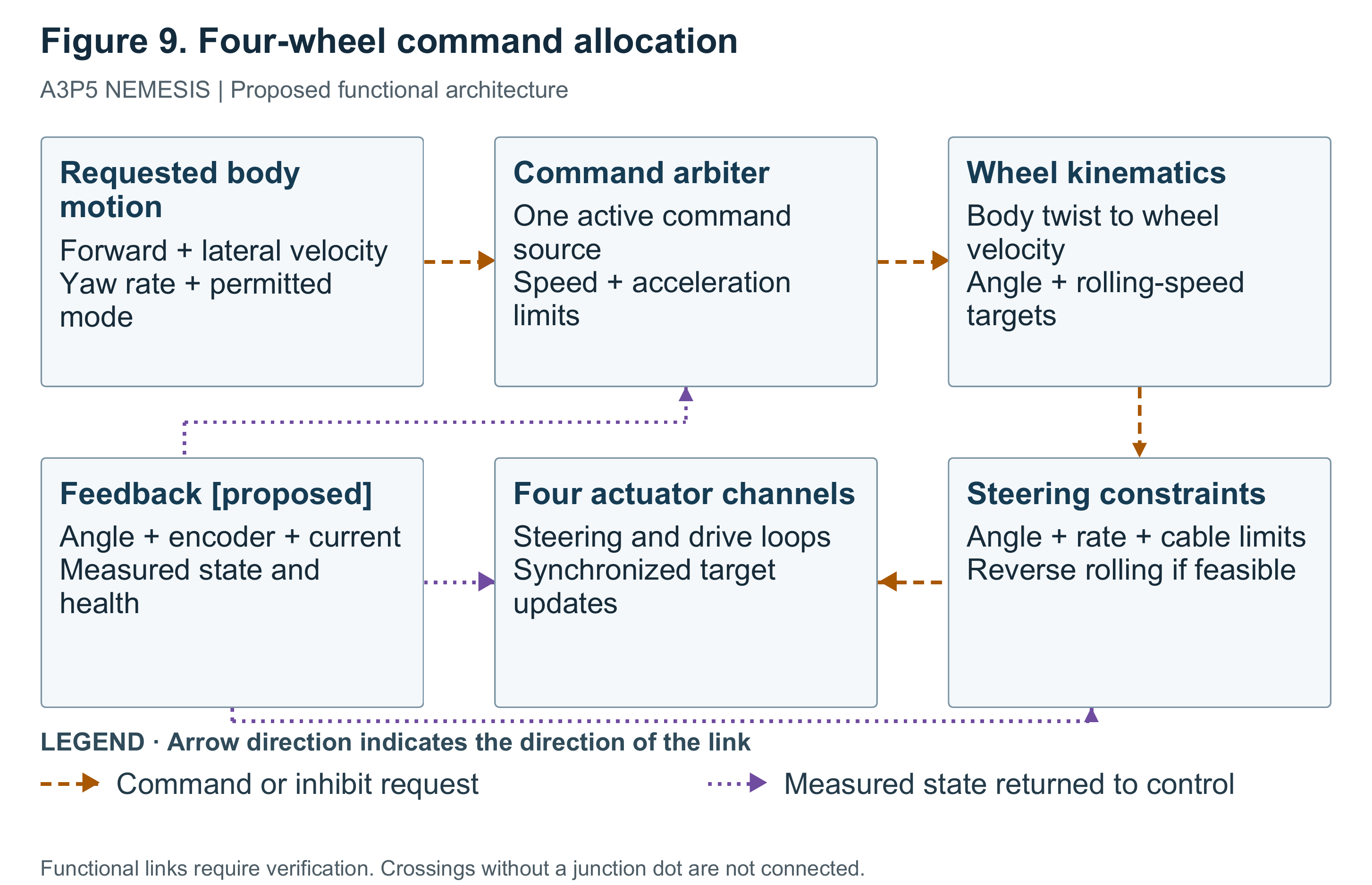}
\caption{Four-wheel command allocation. A common body-motion request is converted into wheel-angle and speed targets subject to steering travel, actuator limits and proposed feedback.}
\label{fig:9}
\end{figure}

\begin{figure}[!htbp]
\centering
\includegraphics[width=6.662in,height=3.250in,keepaspectratio]{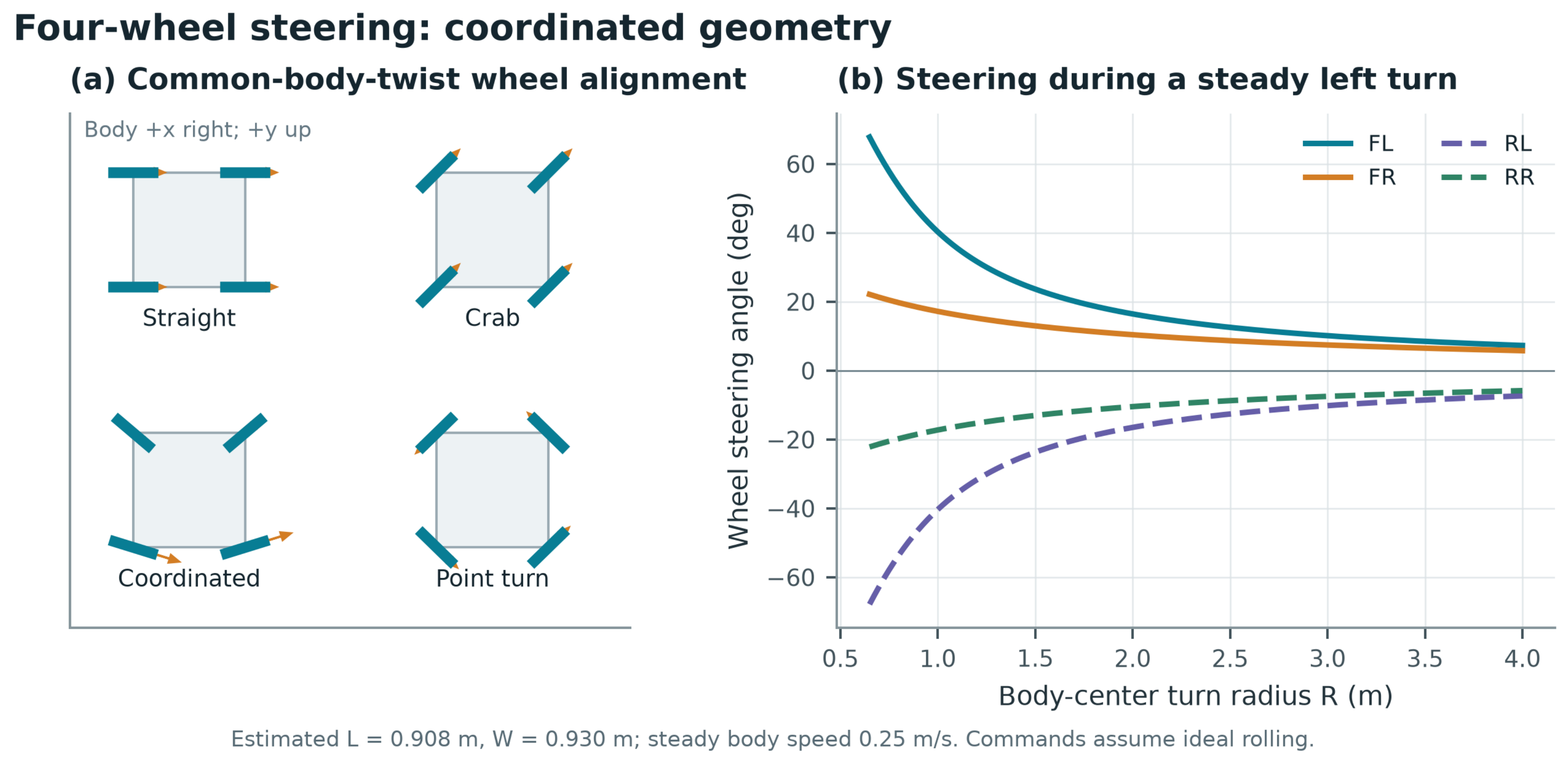}
\caption{Calculated wheel coordination for the estimated wheelbase and track. The plotted angle convention permits an equivalent reversed rolling direction. This is ideal geometry and does not establish unrestricted steering, zero-radius motion or measured path accuracy.}
\label{fig:10}
\end{figure}

\FloatBarrier

\subsection{Feedback, slip and implementation limits}\label{sec:5.2}

The recommended implementation includes independent steering-angle feedback and wheel encoders, neither established by photographs alone. A speed controller can use proportional-integral action with saturation and anti-windup, while the steering loop tracks an angle reference within hard and software limits. Driver current and motor temperature provide additional fault information. Each wheel channel should report its target, measured state and saturation status so that a high-level controller does not mistake commanded motion for achieved motion. The proposed feedback and traction-monitoring chain is summarized in Figure 11.

\begin{figure}[!htbp]
\centering
\includegraphics[page=4,width=6.045in,height=3.900in,keepaspectratio]{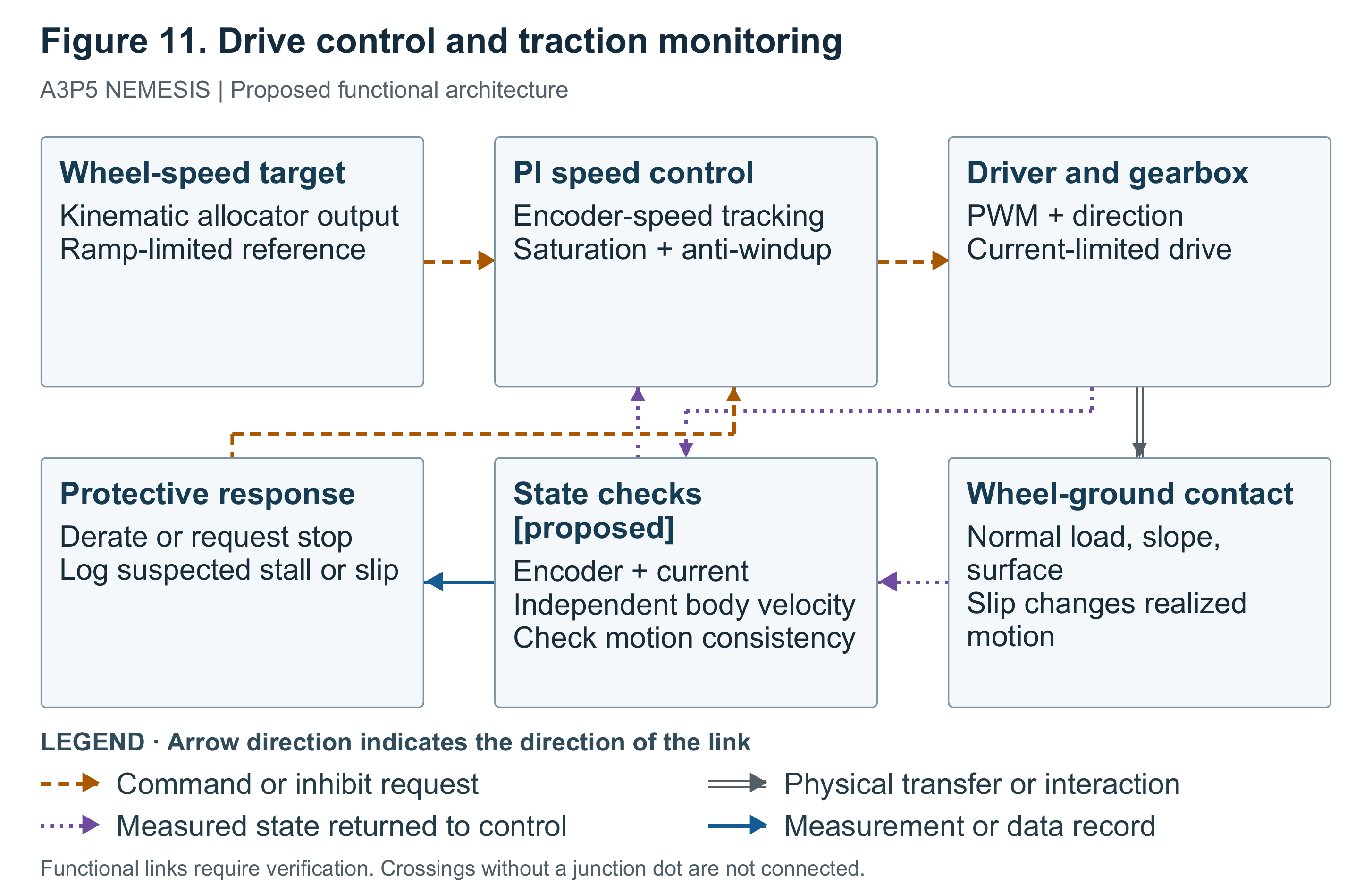}
\caption{Proposed drive-feedback architecture. Encoder and independently estimated body-motion feedback support tracking and traction monitoring; wheel--ground interaction remains a source of uncertainty.}
\label{fig:11}
\end{figure}

A useful kinematic diagnostic is the velocity component perpendicular to a wheel heading. Under ideal rolling it should be zero. In practice it can reveal inconsistency between steering, wheel commands and an independently estimated body velocity. It is not a direct slip sensor when body velocity is computed only from the same wheel encoders. Longitudinal and lateral slip require separate interpretation; Burghi et al. \citep{burghi2024_cccce5} examine adaptive slip compensation for a differential-drive robot through numerical simulation. Their results provide methodological context for controller development; this four-wheel-steering platform requires its own identified model and validation.

\begin{equation}
\label{eq:3}
e_{\perp,i}=-\sin\delta_i\,(v_x-\omega y_i)+\cos\delta_i\,(v_y+\omega x_i)
\end{equation}

The study does not assume a particular suspension law. The photographs show mechanical links and wheel supports, but do not reveal spring stiffness, damping, articulation limits or load equalization. Applying an existing rocker-bogie or omni-wheel model without identifying the actual mechanism would introduce false precision. First measure wheel-contact locations and their motion relative to the chassis during controlled articulation, then derive constraints that match the assembled vehicle.

\FloatBarrier

\subsection{Grade force and torque demand}\label{sec:5.3}

A first longitudinal sizing model balances acceleration, gravity and rolling resistance. For mass m, slope \ensuremath{\alpha}, forward acceleration a and rolling-resistance coefficient \(C_{rr}\), required force is written below. An optional external force represents a tether or contact load. The model assumes continuing wheel contact and omits sinkage, bulldozing losses and steering scrub. These limitations matter on loose terrain, where the contact mechanics discussed by Iagnemma and Dubowsky \citep{iagnemma2004_9f7418} require richer treatment.

\begin{equation}
\label{eq:4}
F_{\mathrm{req}}=ma+mg\sin\alpha+C_{rr}mg\cos\alpha+F_{\mathrm{ext}}
\end{equation}

\begin{equation}
\label{eq:5}
\tau_{w,\mathrm{req}}=\frac{rF_{\mathrm{req}}}{4},\qquad F_{\mathrm{req}}\leq\mu mg\cos\alpha
\end{equation}

The torque expression assumes equal force sharing by four driven wheels and gives gearbox-output torque at the wheel. If a datasheet already specifies geared output torque, do not apply the gear ratio again. For a motor-shaft comparison, divide output torque by gear ratio and gear efficiency once. The traction inequality is a necessary screen under uniform friction and loading; individual limits are |\(F_i\)| \ensuremath{\leq} \(\mu_i N_i\). A lightly loaded wheel can saturate before the total vehicle inequality is reached. Figure 12 plots the resulting grade-dependent wheel torque and minimum friction requirements.

\begin{figure}[H]
\centering
\includegraphics[width=6.239in,height=3.000in,keepaspectratio]{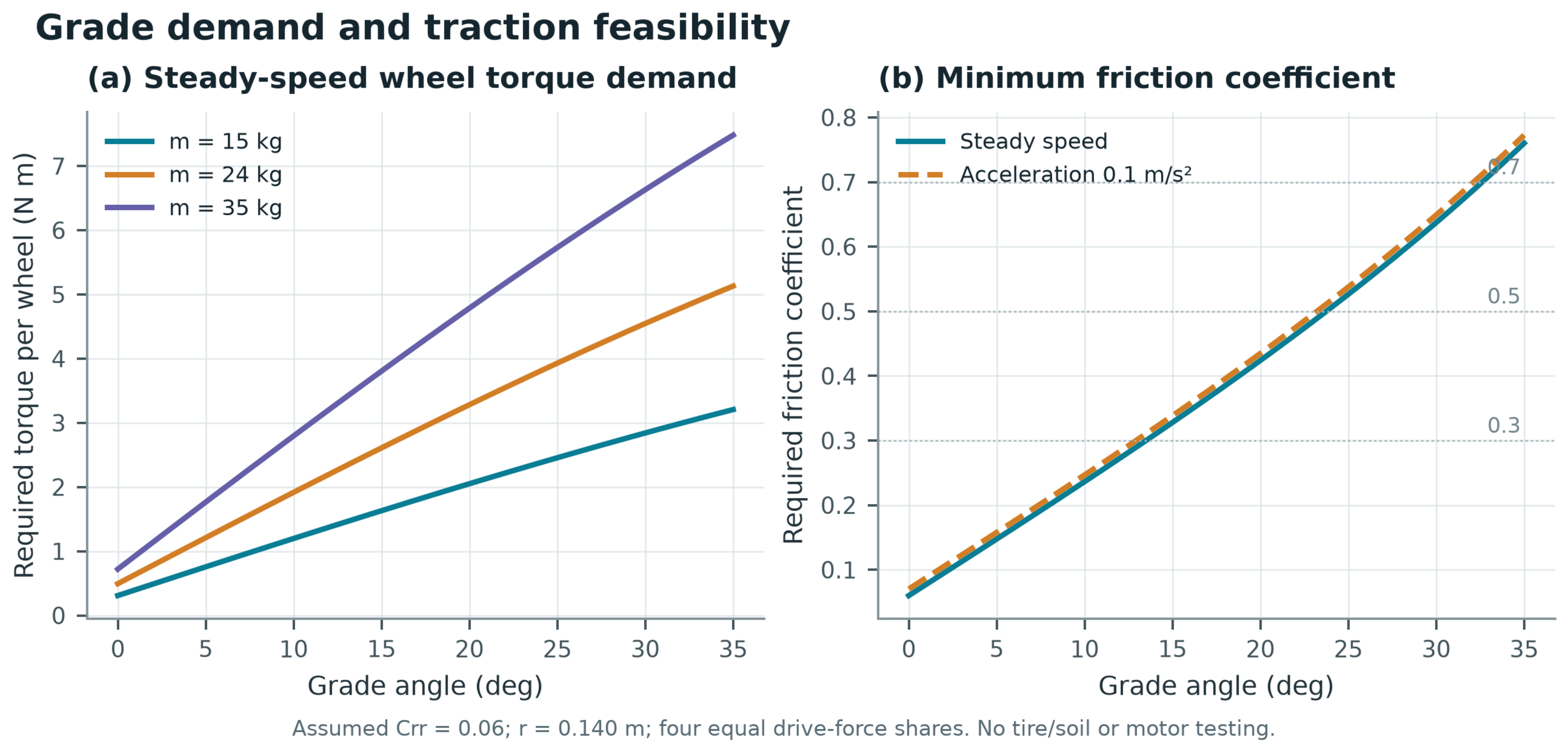}
\caption{Analytical grade demand for declared mass and rolling-resistance scenarios. Torque is required gearbox-output torque per wheel under equal sharing. The friction boundary is a simplified necessary condition, not evidence of loose-soil mobility or safe climbing.}
\label{fig:12}
\end{figure}

For the declared 24 kg, \(C_{rr}\) = 0.06, zero-acceleration case, a 20\textdegree{} grade requires 93.77 N total force and 3.282 N\ensuremath{\cdot}m per wheel at a 0.140 m radius. The corresponding uniform friction coefficient is at least 0.424. On level ground, the same model requires 0.494 N\ensuremath{\cdot}m per wheel. This difference shows why a successful indoor drive is insufficient evidence for a sloped outdoor mission. Continuous torque, motor heating, battery sag and actual friction must be measured before selecting an operating limit.

\FloatBarrier

\section{Manipulation, sampling and stability}\label{sec:6}

\FloatBarrier

\subsection{Arm function and integration}\label{sec:6.1}

The manipulator retains its original deck location and folded configuration. Its visible structure includes a yaw base, shoulder and elbow assemblies, laterally mounted motors, a wrist enclosure and a two-finger gripper. The model exposes named pivots for editing, but it is not a calibrated kinematic chain. Forward kinematics require measured joint axes, link transforms, zero positions and limits. Payload capacity additionally requires link masses, gearbox ratings, bearing loads and mounting stiffness. Figures 13--15 connect the manipulation workflow with the full-rover handling arrangement and the gripper contact detail.

For a measured mechanism, tool pose can be expressed as a product of joint transforms. The required tool wrench \(W_{\mathrm{req}}\) is the three-component force and three-component moment that the manipulator must exert, expressed in the same frame as the 6 \ensuremath{\times} n geometric Jacobian. It maps to actuator torque through the Jacobian transpose. An environment-applied wrench uses the opposite sign in the actuator compensation equation. For n joints, q denotes the joint-coordinate vector and \(T_j\) the successive link transforms from base frame B to tool frame E. In the torque expression, the gravity term accounts for link weight, while the remaining dynamic term represents inertial and frictional contributions. A gripper that holds an object while stationary does not establish safe reach or manipulation during driving. The proposed sampling mode inhibits chassis motion before the arm approaches a sample and releases that inhibition only after the arm and probes reach a verified travel configuration.

\begin{equation}
\label{eq:6}
\begin{gathered}
{}^{B}T_E(\mathbf{q})=\prod_{j=1}^{n}T_j(q_j), \\ \symbfit{\tau}=J(\mathbf{q})^T\mathbf{W}_{\mathrm{req}}+\symbfit{\tau}_g+\symbfit{\tau}_{\mathrm{dyn}}
\end{gathered}
\end{equation}

\begin{figure}[H]
\centering
\includegraphics[page=9,width=6.045in,height=3.900in,keepaspectratio]{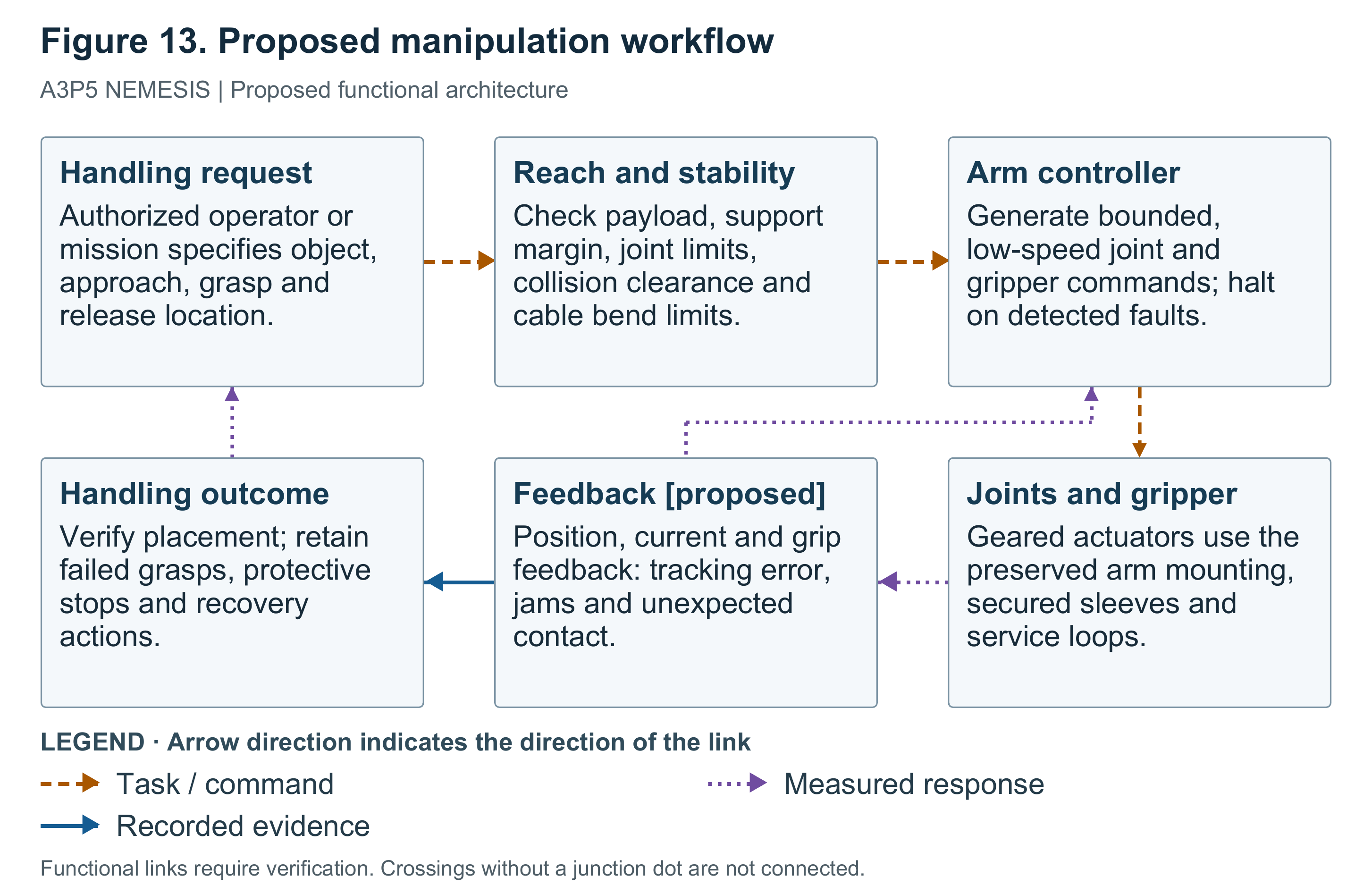}
\caption{Proposed manipulation workflow. Reach, stability, collision and cable-clearance checks precede motion; joint, current and grip feedback support execution and recovery.}
\label{fig:13}
\end{figure}

\begin{figure}[H]
\centering
\includegraphics[width=3.744in,height=2.600in,keepaspectratio]{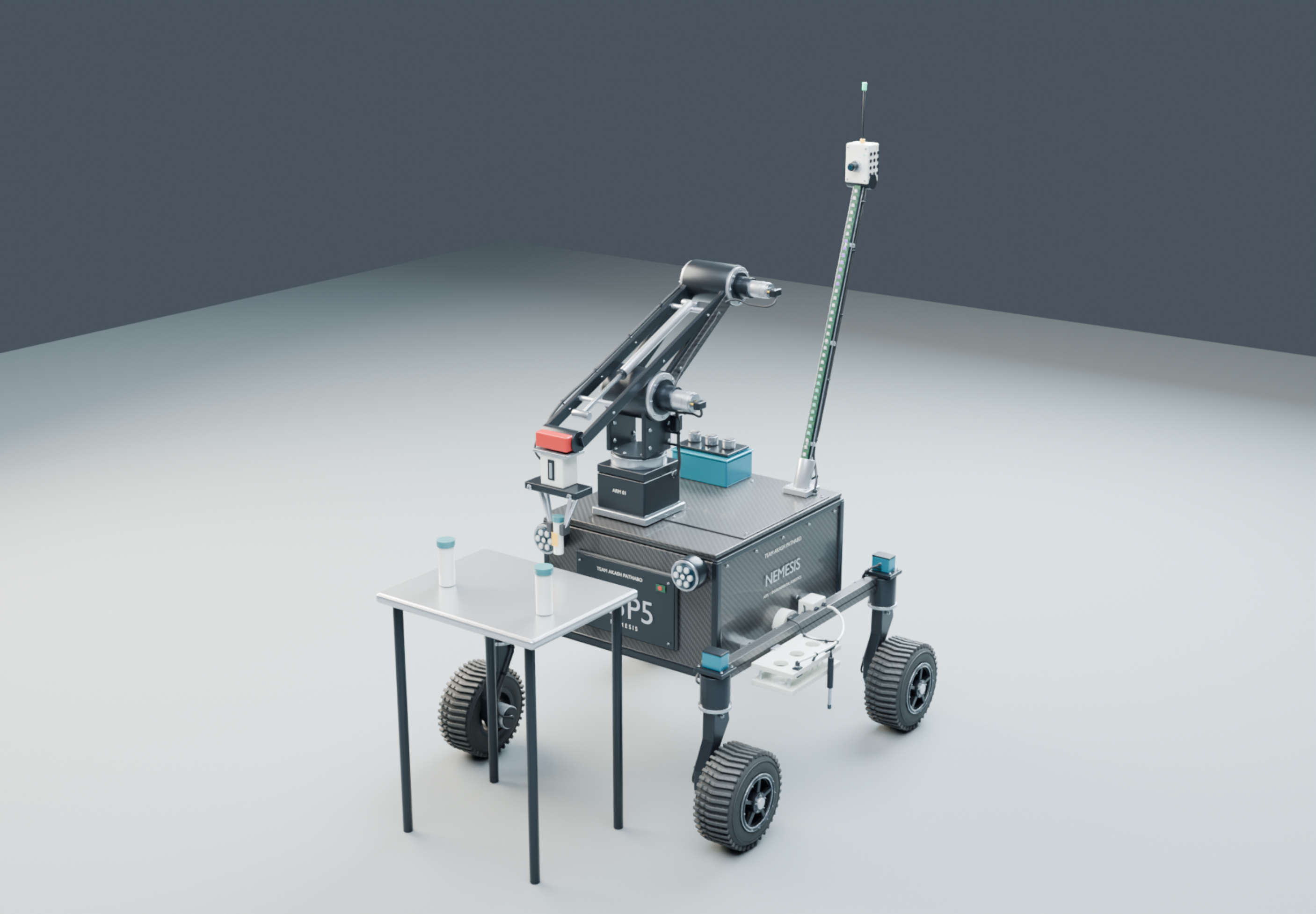}
\caption{Blender lightweight-handling concept with a sample vial between the gripper tips and a nearby preparation bench. The principal arm position is preserved; payload and grasp performance remain unmeasured.}
\label{fig:14}
\end{figure}

\begin{figure}[!htbp]
\centering
\includegraphics[width=3.182in,height=2.400in,keepaspectratio]{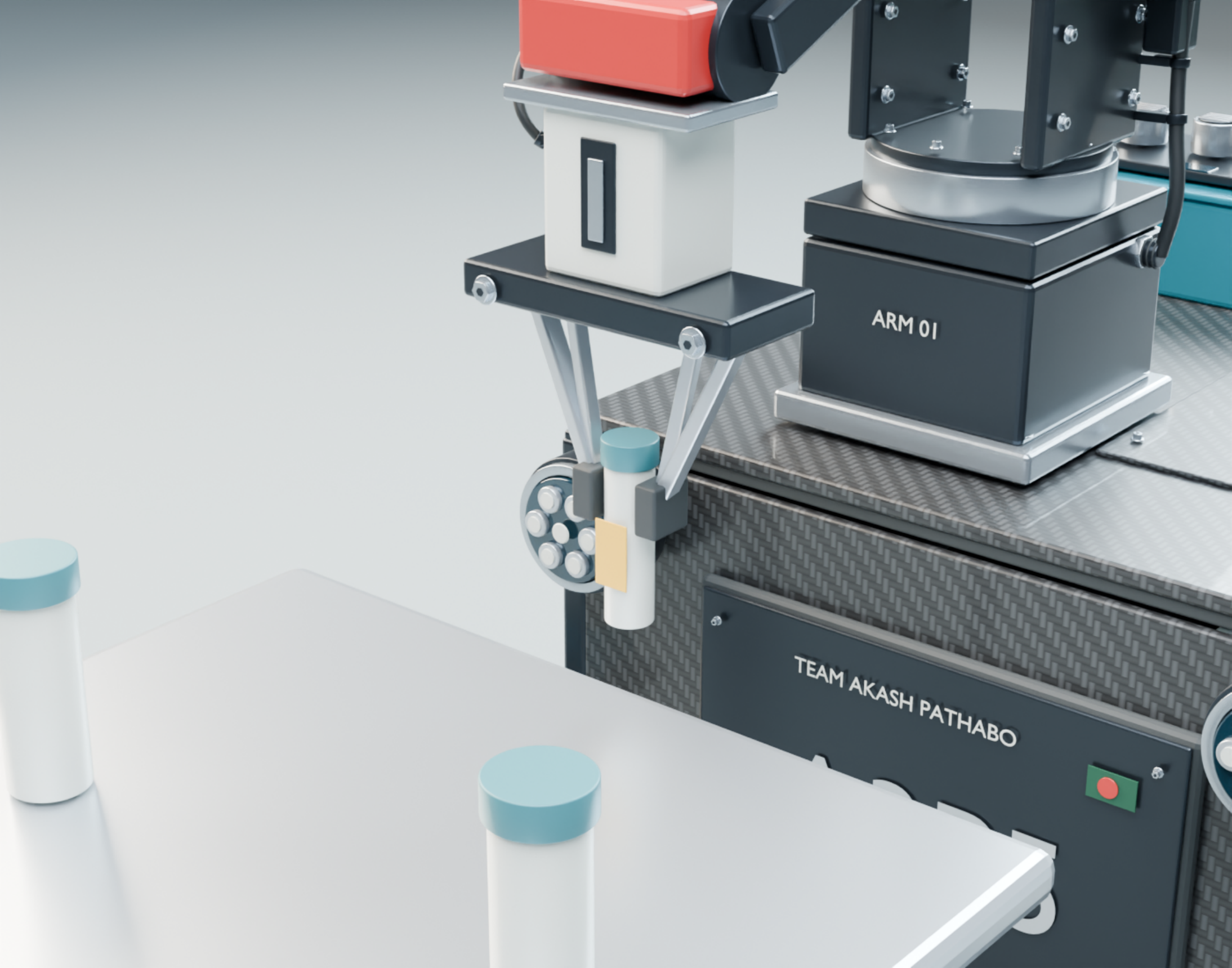}
\caption{Detail of the sample vial and gripper-tip contact arrangement in the preserved arm pose.}
\label{fig:15}
\end{figure}

\FloatBarrier

\subsection{Payload-dependent support margin}\label{sec:6.2}

A mobile manipulator must be evaluated as one mass distribution. Centre of mass depends on arm pose, payload location and every substantial component. The simplified calculation combines base mass \(m_b\) at position \(\mathbf{p}_b\) with payload mass \(m_p\) at position \(\mathbf{p}_p\). It holds the unladen base centre of mass fixed and adds the payload at a declared forward location, isolating one sensitivity while omitting movement of the arm links themselves. Force-angle and edge-moment methods, including the stability measure of Papadopoulos and Rey \citep{papadopoulos1996_3a8aa9}, provide ways to extend the assessment to loading and contact geometry beyond this simplified coplanar model.

\begin{equation}
\label{eq:7}
\mathbf{p}_{CG}=\frac{m_b\mathbf{p}_b+m_p\mathbf{p}_p}{m_b+m_p}
\end{equation}

\begin{equation}
\label{eq:8}
\begin{gathered}
M_{\mathrm{front}}=\frac{L}{2}-x_{CG}-h_{CG}\tan\beta, \\ \beta_{\mathrm{tip}}=\tan^{-1}\!\left(\frac{L/2-x_{CG}}{h_{CG}}\right)
\end{gathered}
\end{equation}

\(M_{\mathrm{front}}\) is the projected distance to the front support edge on a slope descending toward the front. Zero defines an ideal geometric tipping boundary, not an allowable operating slope. Compliance, wheel lift, acceleration, arm contact, uneven ground and centre-of-mass uncertainty reduce the useful margin. Even a positive static value can be inadequate when a wheel unloads or arm motion produces a transient reaction moment. Figure 16 shows how payload and centre-of-mass assumptions affect the geometric support margin.

The assumed unloaded front-tipping bound is 38.05\textdegree{}. Adding 1 kg at the stated position reduces it to 35.26\textdegree{}, while 2 kg reduces it to 32.66\textdegree{}. At a 20\textdegree{} downhill slope, corresponding front margins are approximately 0.243, 0.205 and 0.170 m. These values show the direction and magnitude of model sensitivity. They do not justify a 2 kg payload because arm torque, gripper retention and actual mass distribution remain unmeasured.

\begin{figure}[!htbp]
\centering
\includegraphics[width=5.449in,height=2.620in,keepaspectratio]{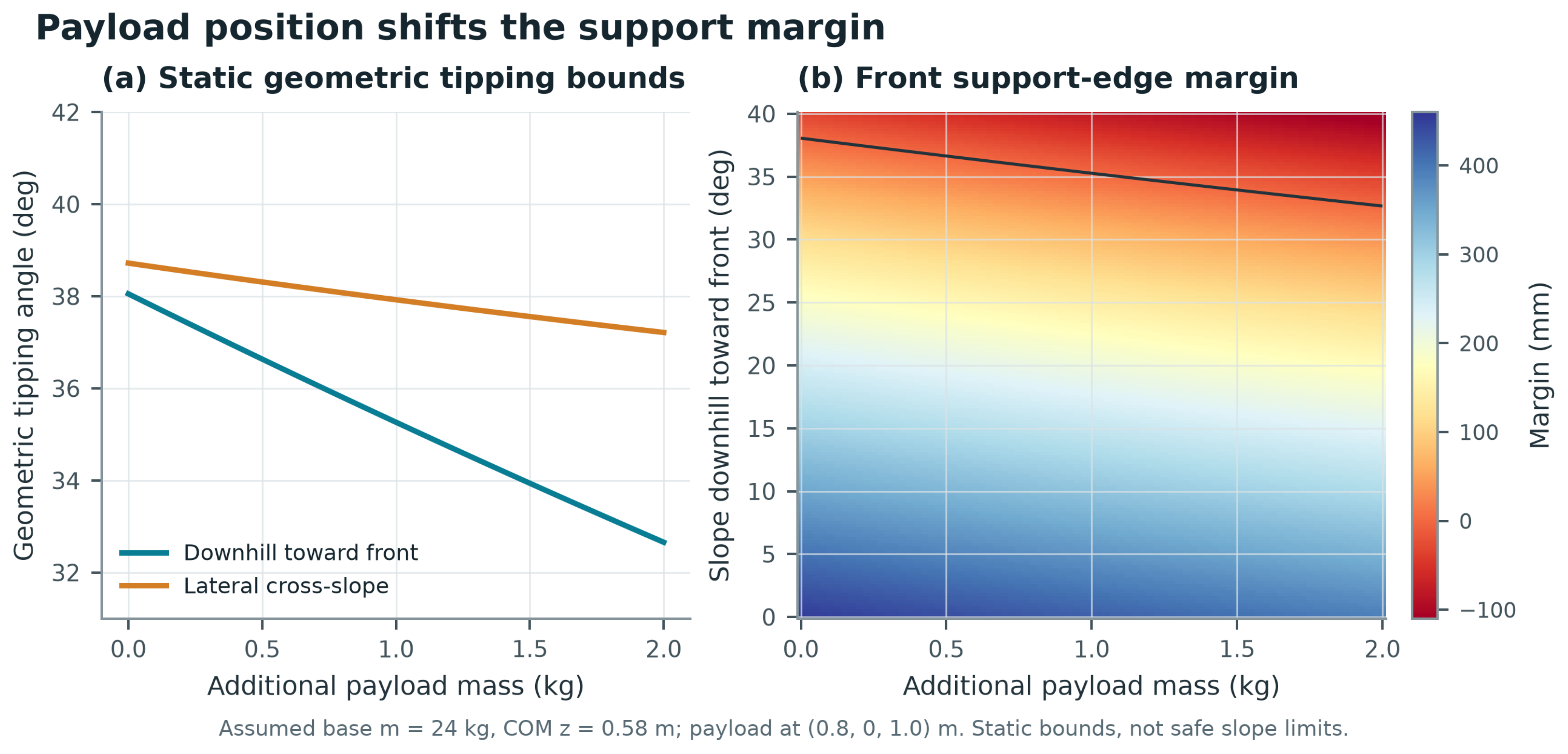}
\caption{Calculated support-margin sensitivity for an assumed 24 kg base with centre-of-mass height 0.58 m and a payload at (0.8, 0, 1.0) m. The curves are ideal static geometric bounds and must not be used as approved operating slopes.}
\label{fig:16}
\end{figure}

A conservative implementation should enforce a pose-dependent envelope after physical identification. It can reject a requested arm pose when the estimated support margin is insufficient, accounting for uncertainty rather than a single nominal value. Low-confidence terrain contact or an unknown payload should trigger a more restricted mode. Cable routing must be checked within the same envelope: a reachable pose is unusable if it pulls a connector or traps a harness against a wheel support.

\FloatBarrier

\subsection{Water access and sample logistics}\label{sec:6.3}

The side rack provides organized sample handling, but access to water is a separate mechanical problem. In the mission illustration, a staged vessel sits beneath the existing pH probe so that the electrode enters the liquid while the upper housing remains clear. This shows spatial compatibility in one configuration. It does not demonstrate automatic deployment into a river, a sealed wet enclosure or a mechanism for raising and lowering the probe. The complete sampling arrangement and the electrode-immersion detail are shown in Figures 17 and 18.

\begin{figure}[H]
\centering
\includegraphics[width=4.954in,height=3.440in,keepaspectratio]{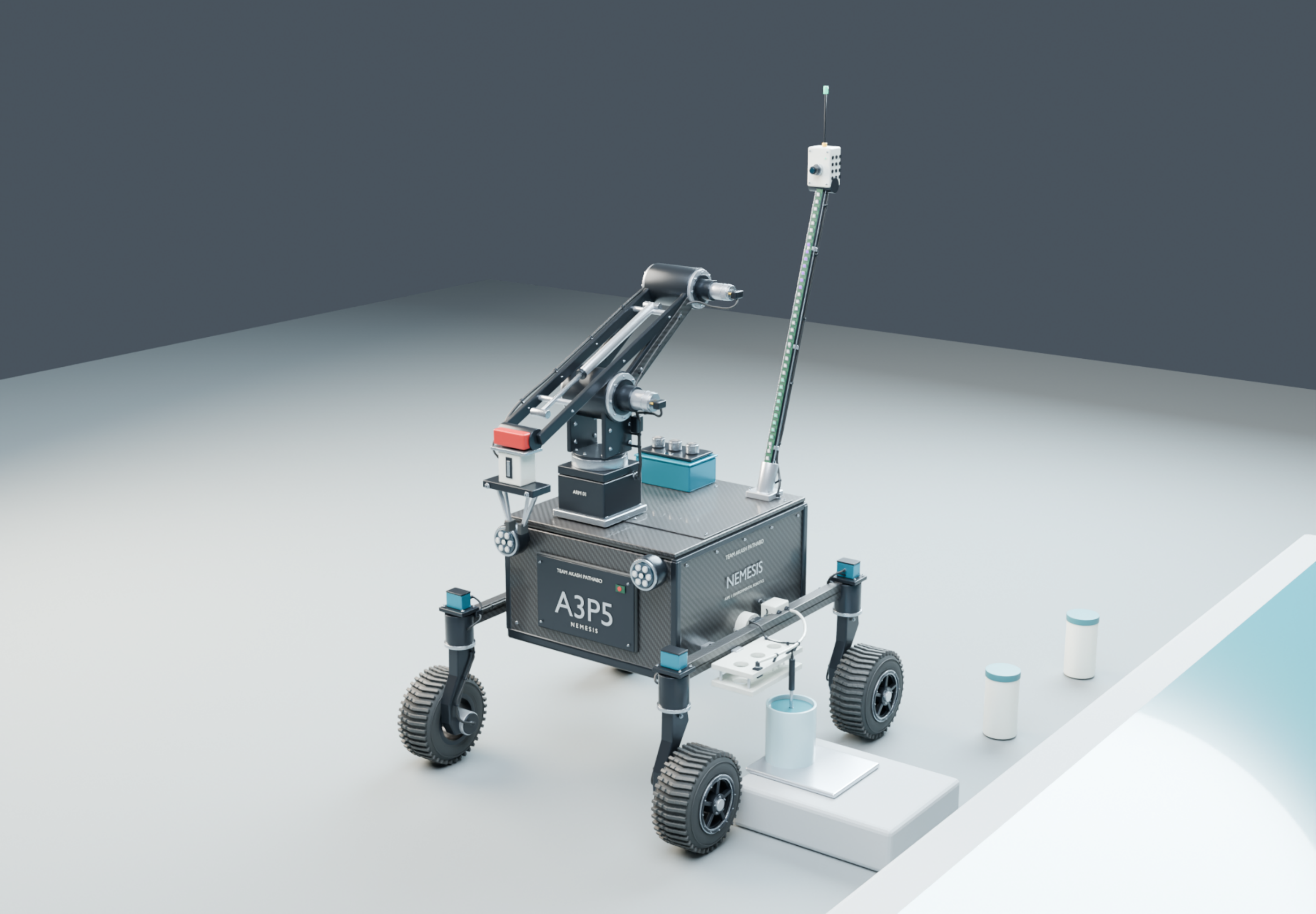}
\caption{Blender water-assessment concept with a staged vessel beneath the existing side-mounted pH probe and the chassis on dry support. Autonomous deployment and water-ingress protection require physical validation.}
\label{fig:17}
\end{figure}

Sampling records should connect a physical container to time, position estimate, operator or mission ID, probe calibration and acquisition state. Rinsing, stabilization and contamination control affect interpretation and belong in the workflow. Turbidity can be sensitive to bubbles, container geometry and residue. Samples requiring laboratory confirmation should retain their identity through collection, transport and analysis; the rover reading can then be compared with the laboratory result rather than used as an unsupported replacement.

\begin{figure}[!htbp]
\centering
\includegraphics[width=4.200in,height=3.300in,keepaspectratio]{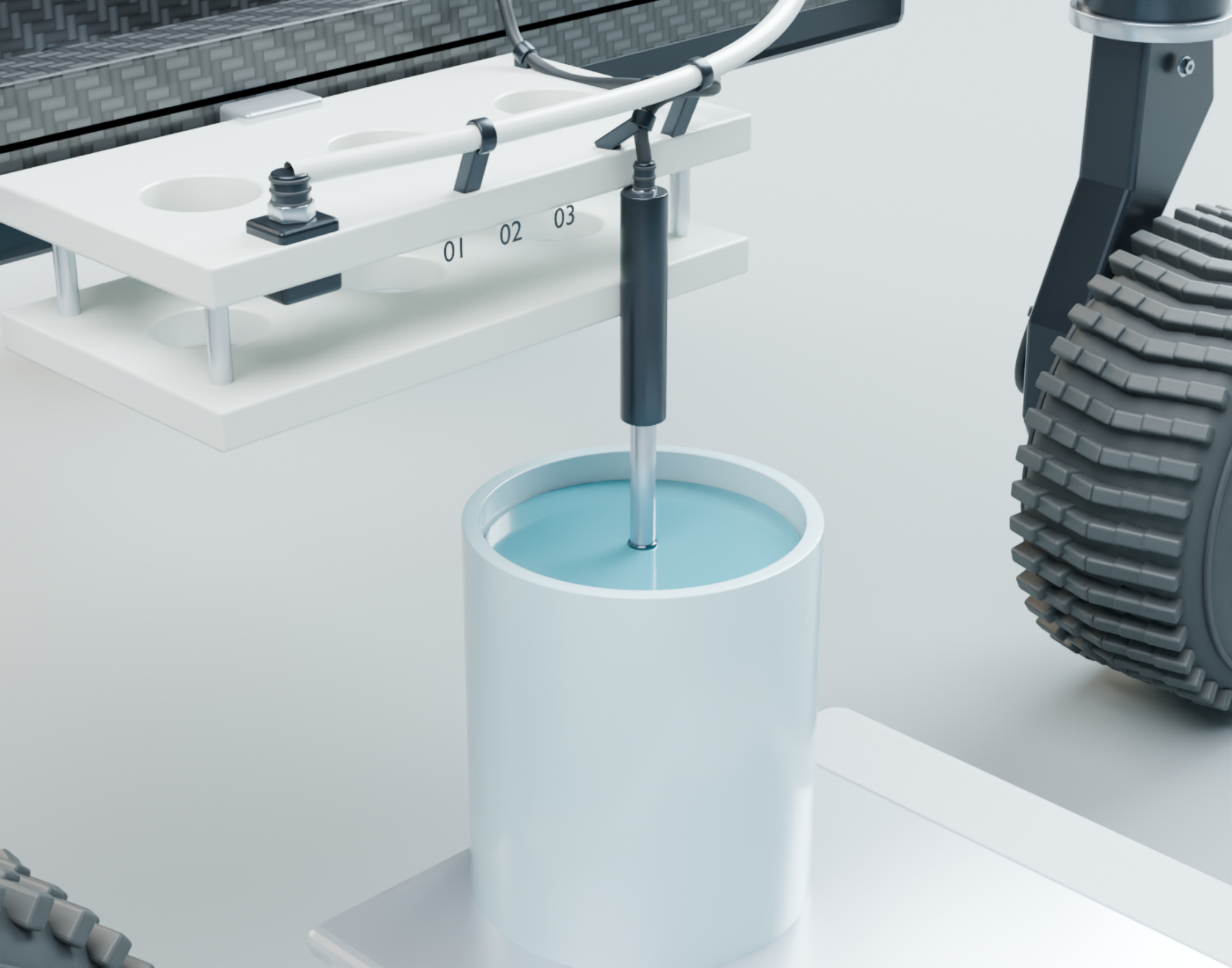}
\caption{Detail of the illustrative sampling interface: the electrode tip enters the liquid while its upper housing remains clear.}
\label{fig:18}
\end{figure}

\FloatBarrier

\section{Environmental sensing and measurement assurance}\label{sec:7}

\FloatBarrier

\subsection{Measurement scope and traceability}\label{sec:7.1}

The supplied A3P5 description identifies MQ135, MQ7, MQ4, PMS7003, GUVA and BMP180 devices, together with pH and turbidity sensing. These identifiers define the reported sensor inventory; they do not constitute a verified as-built bill of materials. Calibration, electrical interface revisions and analytical performance require separate identification. Accordingly, the environmental subsystem is specified as a collection of measurement chains, each retaining its raw response, conversion version, timestamp and quality flags. Concentration estimates require comparison with an independent reference. A numerical display alone does not establish analyte selectivity, traceability or suitability for regulatory monitoring. The following design provisions therefore define work required to convert the prototype inventory into defensible environmental observations.

\FloatBarrier

\subsection{Gas sensing, heater scheduling and electrical integration}\label{sec:7.2}

The three MQ channels should be interpreted jointly with operating conditions. MQ135 is a broadly responsive semiconductor sensor rather than a selective carbon-dioxide instrument; its response can arise from several gases and vapours. For a conventional divider, the sensing resistance is reconstructed from the measured load voltage:

\begin{equation}
\label{eq:9}
R_s=R_L\left(\frac{V_c}{V_L}-1\right).
\end{equation}

Here, \(R_{s}\) and \(R_{L}\) are sensing and load resistances in ohms, \(V_{c}\) is the divider supply in volts and \(V_{L}\) is the voltage across \(R_{L}\). This expression requires the actual module topology and resistor value to be verified. The ratio \(R_{s}\)/\(R_{0}\) is useful only when the reference condition defining \(R_{0}\) is recorded. A manufacturer sensitivity curve supplies a starting model, not a substitute for unit-specific calibration with interfering gases \citep{winsen2021_36e9d6}. Figure 19 summarizes the proposed gas-sensing sequence from sensor identification and heater control to quality-checked interpretation.

Heater operation is part of the measurement, not merely a power requirement. The MQ-7B manual specifies a 60 s high-temperature phase at 5 V followed by a 90 s low-temperature phase at 1.5 V. Because the rover description says MQ7, the exact installed variant must be checked before applying that schedule. A compatible implementation would timestamp the phase, reject cleaning and transition readings, and log heater voltage alongside accepted samples. Initial conditioning and recovery after prolonged storage require separate procedures \citep{winsen2021_174511}. For MQ4, the manufacturer's nominal methane range does not quantify accuracy or analyte selectivity \citep{winsen2021_dd85bc}. Experiments on another low-cost methane device demonstrate why temperature, humidity and cross-interference deserve explicit calibration terms, while providing no transferable MQ4 coefficients \citep{lin2023_f90829}.

\begin{figure}[!htbp]
\centering
\includegraphics[page=5,width=6.800in,height=4.387in,keepaspectratio]{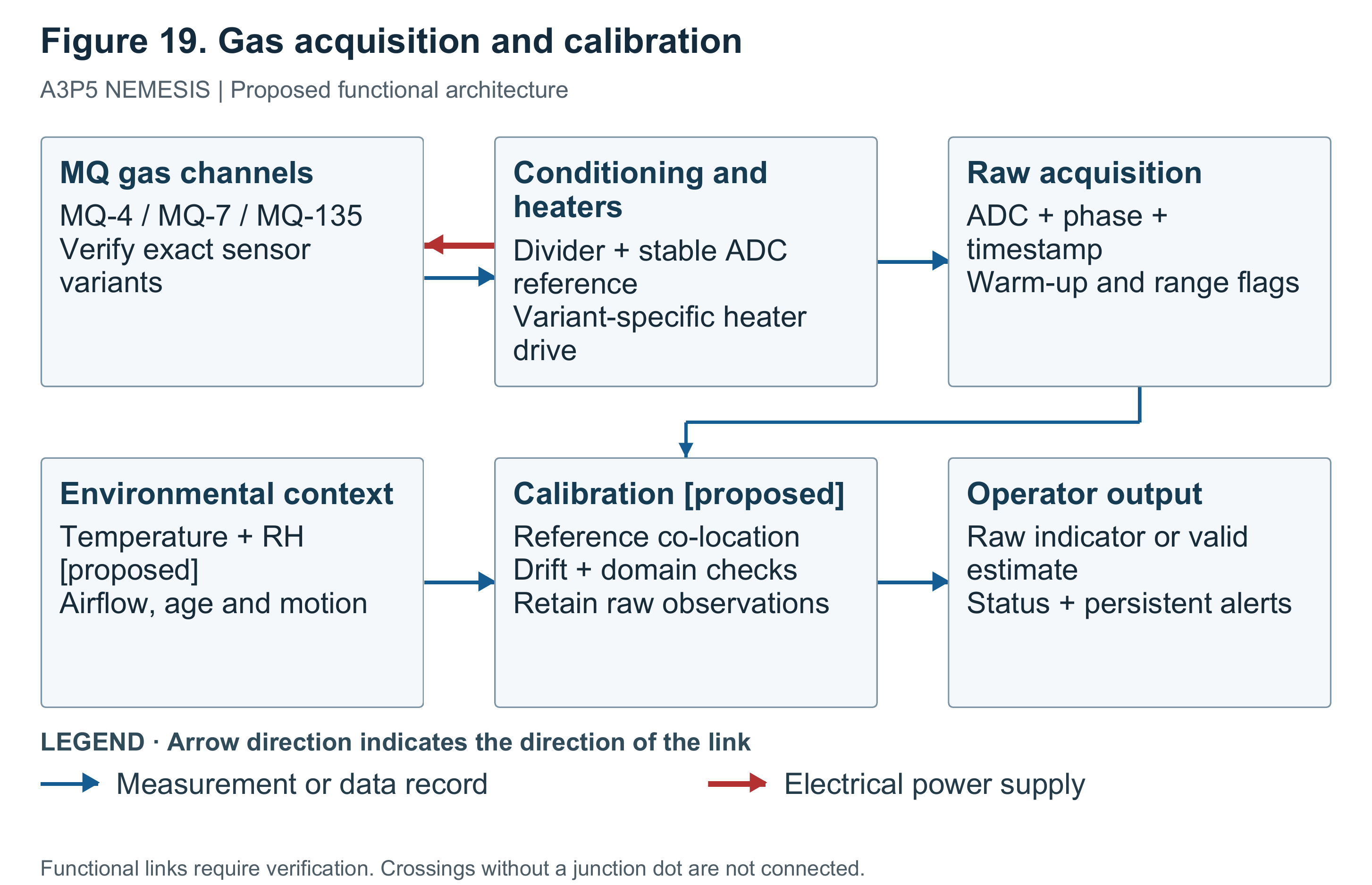}
\caption{Gas-measurement chain incorporating sensor identification, heater scheduling, raw-response acquisition, reference calibration and validity checks.}
\label{fig:19}
\end{figure}

The proposed harness separates motor and heater returns from low-level measurement paths, with a defined connection at the power-distribution reference. Regulated sensor rails, protected inputs and local decoupling reduce the chance that acceleration or servo operation is interpreted as a chemical event. Every connector should be keyed, labelled and mechanically retained; flexible loops must accommodate steering and arm travel without loading terminals. Motor-state and supply-voltage logs should accompany commissioning data so that correlated electrical artefacts can be identified. Cable sleeves improve protection and appearance, but their contribution to measurement integrity depends on routing, termination and verified continuity.

\FloatBarrier

\subsection{Optical particles, humidity and sensor identity}\label{sec:7.3}

Optical particle readings depend on the sampled aerosol and its interaction with moisture. Jayaratne et al. \citep{jayaratne2018_d2f9d3} observed substantial humidity and fog effects with a PMS1003; those measurements motivate humidity-aware quality control but are not PMS7003 calibration data. A collocated temperature--humidity instrument is therefore a proposed addition. The listed BMP180 measures pressure and temperature and cannot supply relative humidity. Its temperature reading may also reflect board heating rather than undisturbed ambient air \citep{boschsensortec2013_0d84c1}. The sampling inlet should remain exposed to representative airflow while shielding direct rain and avoiding wheel-generated dust or enclosure exhaust. Figures 20--22 place the proposed particulate-acquisition chain alongside published humidity-response and calibration evidence.

A useful family of humidity corrections represents hygroscopic enhancement explicitly:

\begin{equation}
\label{eq:10}
\widehat{c}_{\mathrm{dry}}=c_{\mathrm{raw}}\frac{m}{1+\kappa/(100/H-1)},\qquad 0<H<100.
\end{equation}

Here, H is relative humidity in percent, \(c_{\mathrm{raw}}\) and estimated dry concentration have matching mass-concentration units, and m and \ensuremath{\kappa} are fitted scale and growth parameters. Patel et al. \citep{patel2024_4264a8} investigated this structure with PMS5003 observations. It is presented as a candidate for testing, with numerical coefficients deliberately unspecified for A3P5. Behaviour near saturation and changes in aerosol composition require validation and quality flags rather than unrestricted extrapolation. Published comparisons between uncorrected, national and locally adjusted estimates illustrate the importance of evaluating the correction under the intended exposure conditions.

\begin{figure}[!htbp]
\centering
\includegraphics[page=6,width=5.890in,height=3.650in,keepaspectratio]{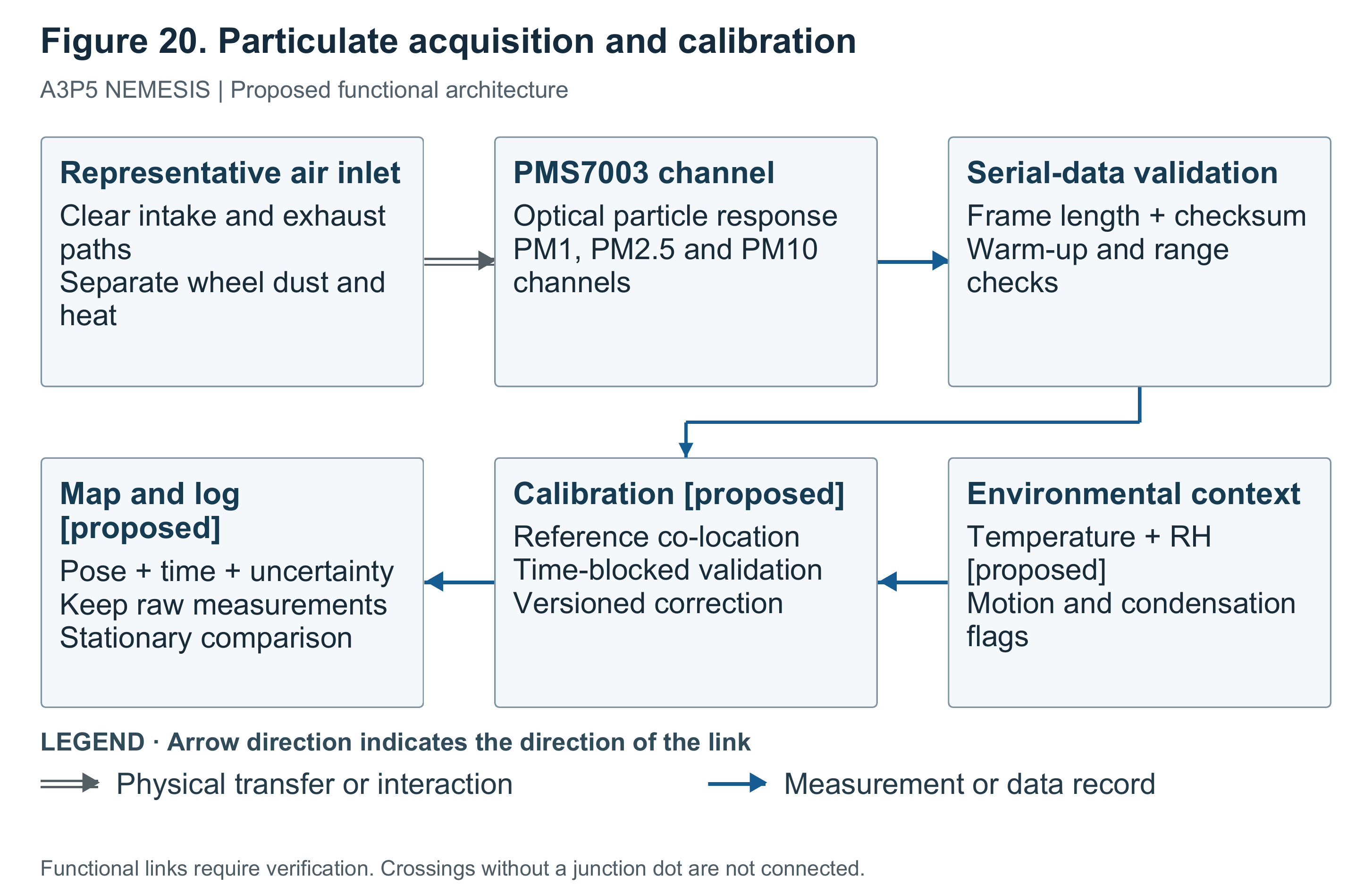}
\caption{Particulate-measurement chain incorporating representative airflow, device checks, environmental covariates and reference-based correction.}
\label{fig:20}
\end{figure}

\begin{figure}[!htbp]
\centering
\includegraphics[width=3.094in,height=2.360in,keepaspectratio]{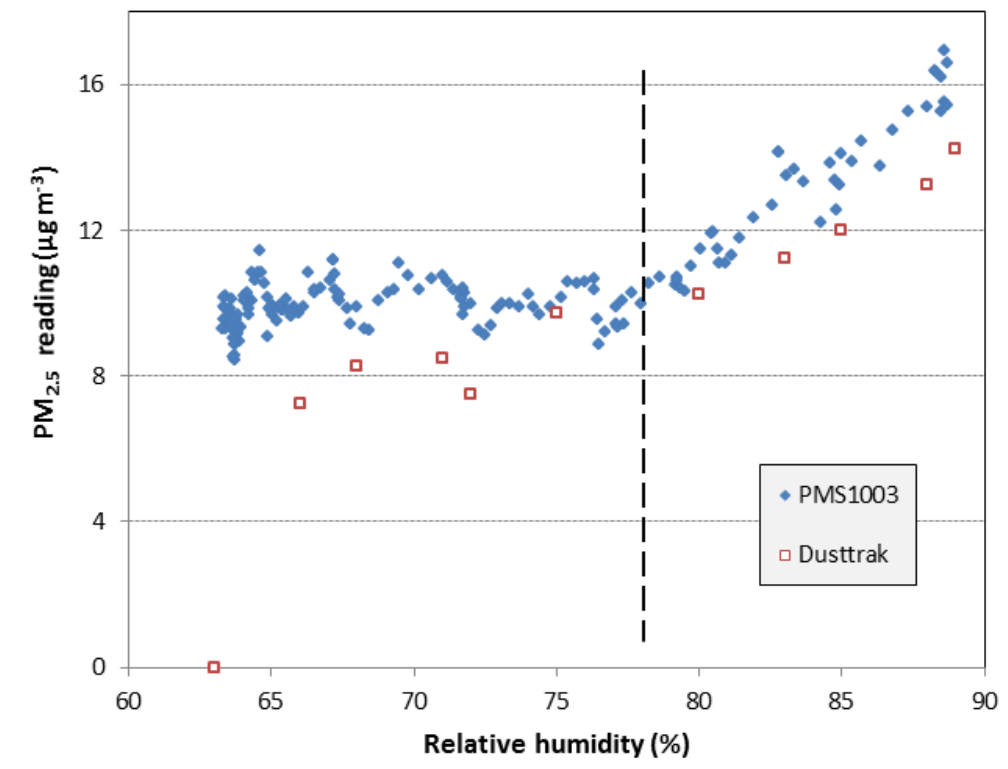}
\caption{External laboratory measurements of PMS1003 and DustTrak PM2.5 response as chamber relative humidity increased. Reproduced unchanged from Jayaratne et al. \citep{jayaratne2018_d2f9d3}, Fig. 2, CC BY 4.0. Source: \url{https://doi.org/10.5194/amt-11-4883-2018}. Licence: \url{https://creativecommons.org/licenses/by/4.0/}.}
\label{fig:21}
\end{figure}

\begin{figure}[!htbp]
\centering
\includegraphics[width=6.004in,height=2.100in,keepaspectratio]{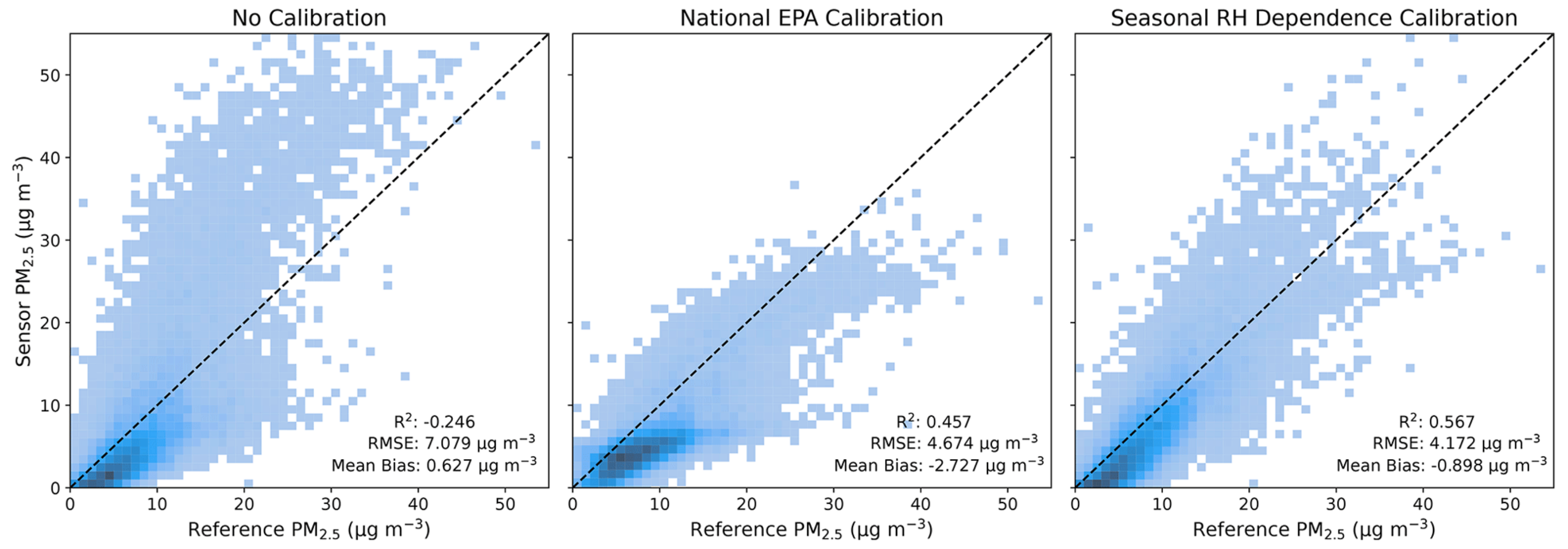}
\caption{External comparison of uncorrected and calibrated PMS5003-system PM2.5 estimates with reference measurements at the Laney site, 2021--2022. Reproduced unchanged from Patel et al. \citep{patel2024_4264a8}, Fig. 4, CC BY 4.0. Source: \url{https://doi.org/10.5194/amt-17-1051-2024}. Licence: \url{https://creativecommons.org/licenses/by/4.0/}.}
\label{fig:22}
\end{figure}

\FloatBarrier

\subsection{Temporal response and environmental mapping}\label{sec:7.4}

Mobility introduces a mismatch between the location of acquisition and the air parcel influencing the sensor. A first-order approximation makes that limitation explicit:

\begin{equation}
\label{eq:11}
\tau\frac{dz(t)}{dt}+z(t)=c(t),\qquad d_{\mathrm{lag}}\approx v\tau.
\end{equation}

\begin{figure}[!htbp]
\centering
\includegraphics[width=5.407in,height=2.600in,keepaspectratio]{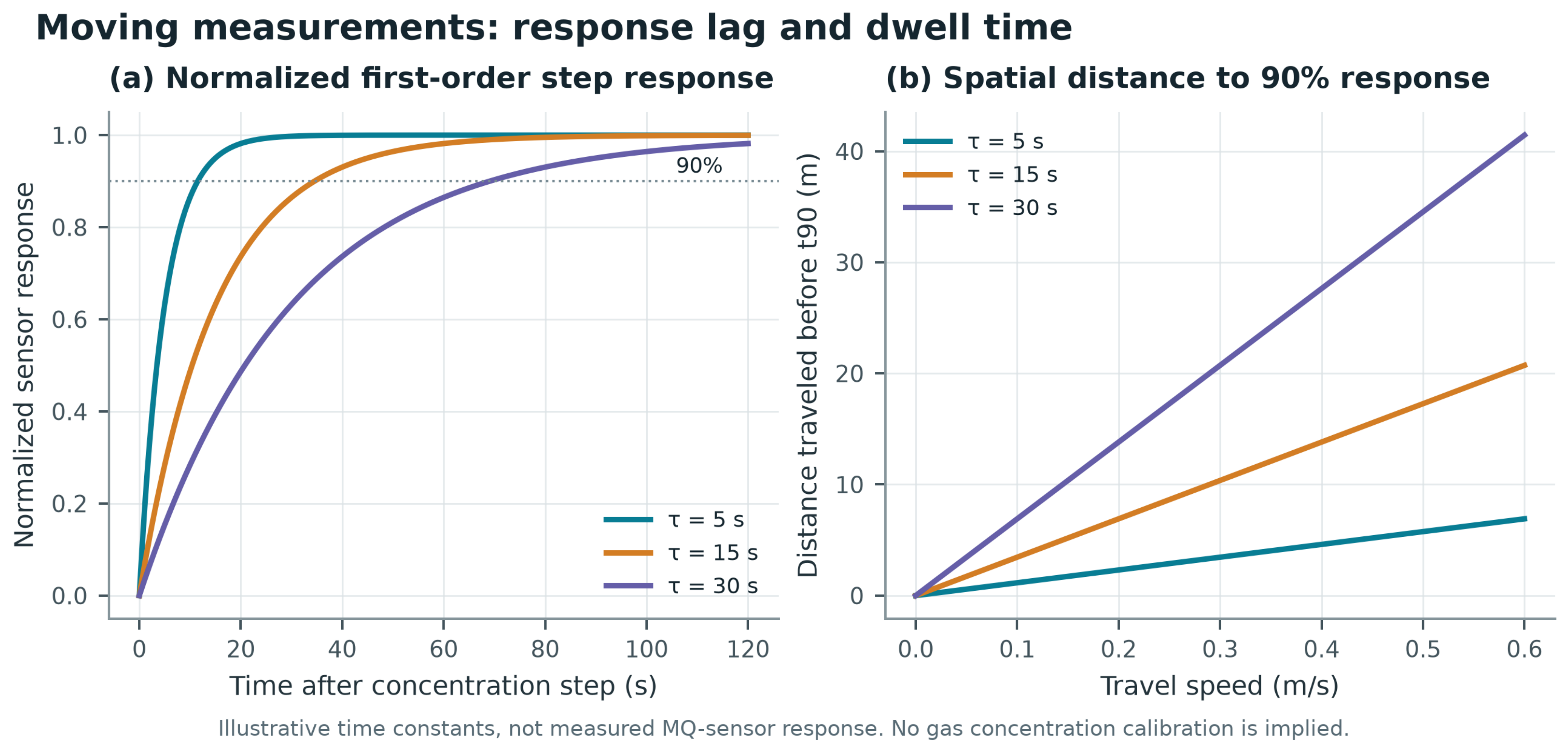}
\caption{Analytical first-order sensor response and motion-lag sensitivity. Time constants and rover speeds are declared scenarios. The curves illustrate why acquisition frequency alone cannot remove physical sensor lag; they are not fitted response tests.}
\label{fig:23}
\end{figure}

\begin{figure}[!htbp]
\centering
\includegraphics[width=4.919in,height=4.750in,keepaspectratio]{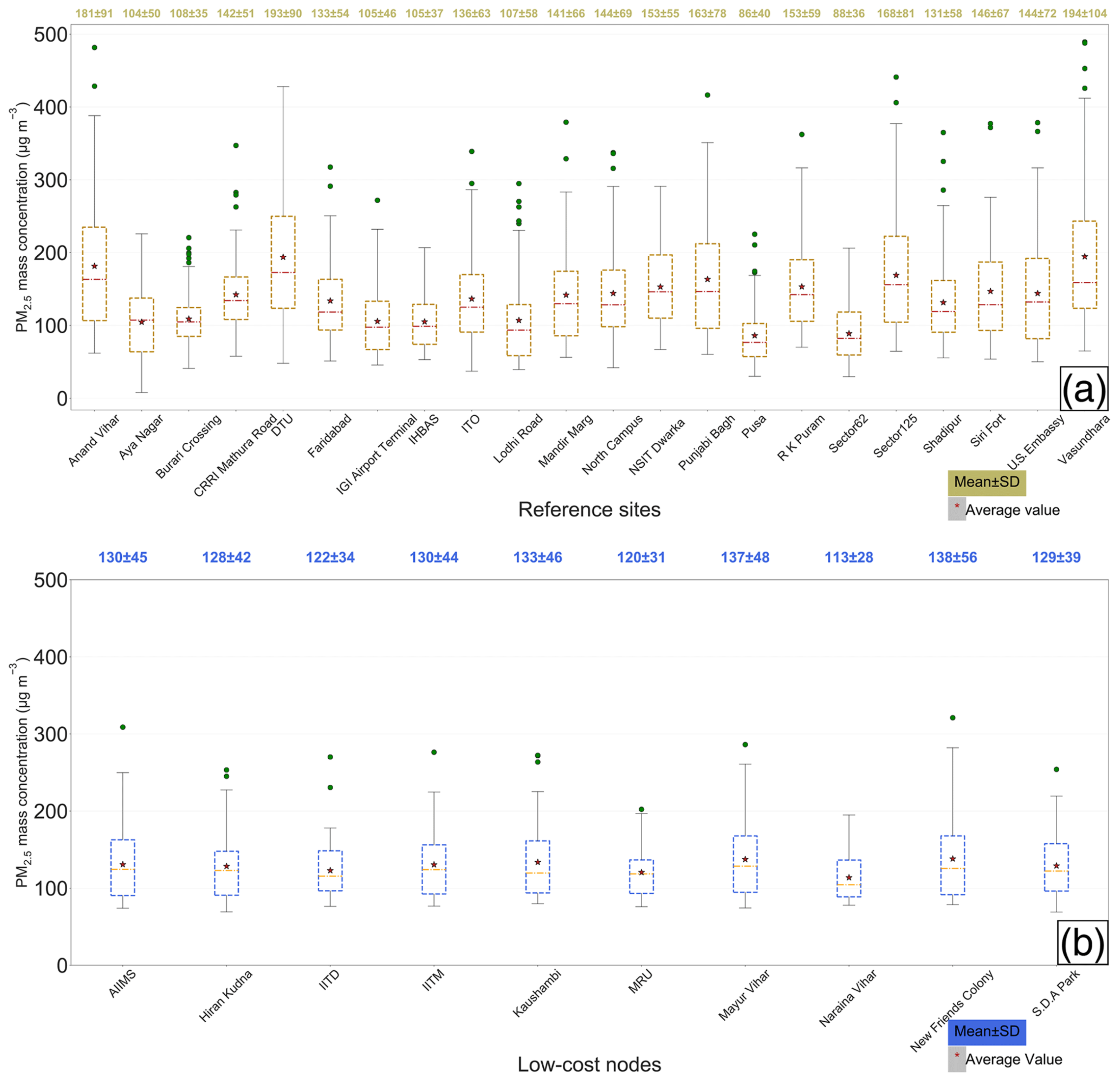}
\caption{External Delhi-network distributions of 24 h PM2.5 reference measurements and Gaussian-process-calibrated low-cost-node observations, including site means and standard deviations. The network included PMS7003 sensors. Reproduced unchanged from Zheng et al. \citep{zheng2019_9520c4}, Fig. 4, CC BY 4.0. Source: \url{https://doi.org/10.5194/amt-12-5161-2019}. Licence: \url{https://creativecommons.org/licenses/by/4.0/}.}
\label{fig:24}
\end{figure}

In this local approximation, z is a concentration-equivalent response, c is the input concentration, \(\tau\) is the response time constant in seconds, v is rover speed in metres per second and \(d_{\mathrm{lag}}\) is an approximate along-track displacement in metres. For a step input, the time to reach 90\% of the final response is \(t_{90}\) = \(\tau\) ln(10), and the corresponding travel distance is \(d_{90}\) = v\(\tau\) ln(10). These threshold quantities differ from the characteristic lag length v\(\tau\): for \(\tau\) = 15 s and v = 0.25 m/s, \(t_{90}\) = 34.54 s and \(d_{90}\) = 8.63 m. Actual semiconductor adsorption and recovery can be asymmetric, so \(\tau\) requires experimental identification. Sampling faster cannot remove this physical lag. Initial mapping trials should combine stationary dwell measurements with travel intervals, record inlet position and pose uncertainty, and distinguish recovery transients from spatial gradients. The analytical response-lag scenarios and an external spatial monitoring example are presented in Figures 23 and 24.

Zheng et al. \citep{zheng2019_9520c4} demonstrated dynamic Gaussian-process calibration in a Delhi network that included PMS7003 devices. Their spatial distributions illustrate achievable analysis after reference collocation, not an accuracy specification for the rover. Information-driven gas mapping additionally depends on localization, environmental transport and an observation model \citep{gongora2023_ca6a1d}. Those inputs must be established before uncertainty-guided sampling is claimed. A mission visualization alone does not establish a valid dispersion field; transport simulation requires separate assumptions and boundary conditions, as exemplified by GADEN \citep{monroy2017_e93884}.

\FloatBarrier

\subsection{Water probes, ultraviolet exposure and calibration uncertainty}\label{sec:7.5}

The pH channel requires a verified electrode interface because glass electrodes have high source impedance. Buffer input bias, surface leakage and moisture can introduce offsets even when digital communication appears reliable. A short, shielded probe connection, clean high-impedance input region and mechanically supported connector are proposed. The ideal temperature-dependent electrode slope is expressed by

\begin{equation}
\label{eq:12}
E=E_7-S(T)(\mathrm{pH}-7),\qquad S(T)=\frac{\ln(10)RT}{F}.
\end{equation}

Here, E is the electrode potential and \(E_{7}\) is its value at pH 7, both in volts, T is solution temperature in kelvin, R is the molar gas constant and F is the Faraday constant. The ideal magnitude is approximately 59.16 mV per pH unit at 25 \textdegree{}C; actual offset and slope require buffer calibration. The conditioning calculation assumes gain G = 3, a 5 V reference and 10-bit ADC with a 2.5 V midpoint; the proposed comparison uses a 3.3 V reference and 12-bit ADC with a 1.65 V midpoint. For an ideal N-bit ADC, one least-significant bit (LSB) represents \(V_{\mathrm{ref}}\)/2\textsuperscript{N} \citep{microchip_adc_an2537}, giving a pH increment of \(V_{\mathrm{ref}}\)/[2\textsuperscript{N} G S(T)]. At 25 \textdegree{}C, the respective increments are 0.0275122 and 0.00453951 pH/LSB. These quantization intervals describe ideal resolution; electrode, amplifier and reference errors determine measurement accuracy. A suitable reference circuit demonstrates the importance of a low-bias buffer and guarded input, but is not asserted to be installed on A3P5 \citep{analogdevices2013_f3acfa}. Field procedures should document buffer temperature, stabilization, rinsing, electrode condition and an independent check solution \citep{usgs2021_e859f3}. Figure 25 presents the calculated pH-conditioning response, Figure 26 the water-assessment workflow, and Figure 27 the ultraviolet, pressure and temperature interfaces.

\begin{figure}[!htbp]
\centering
\includegraphics[width=5.344in,height=2.350in,keepaspectratio]{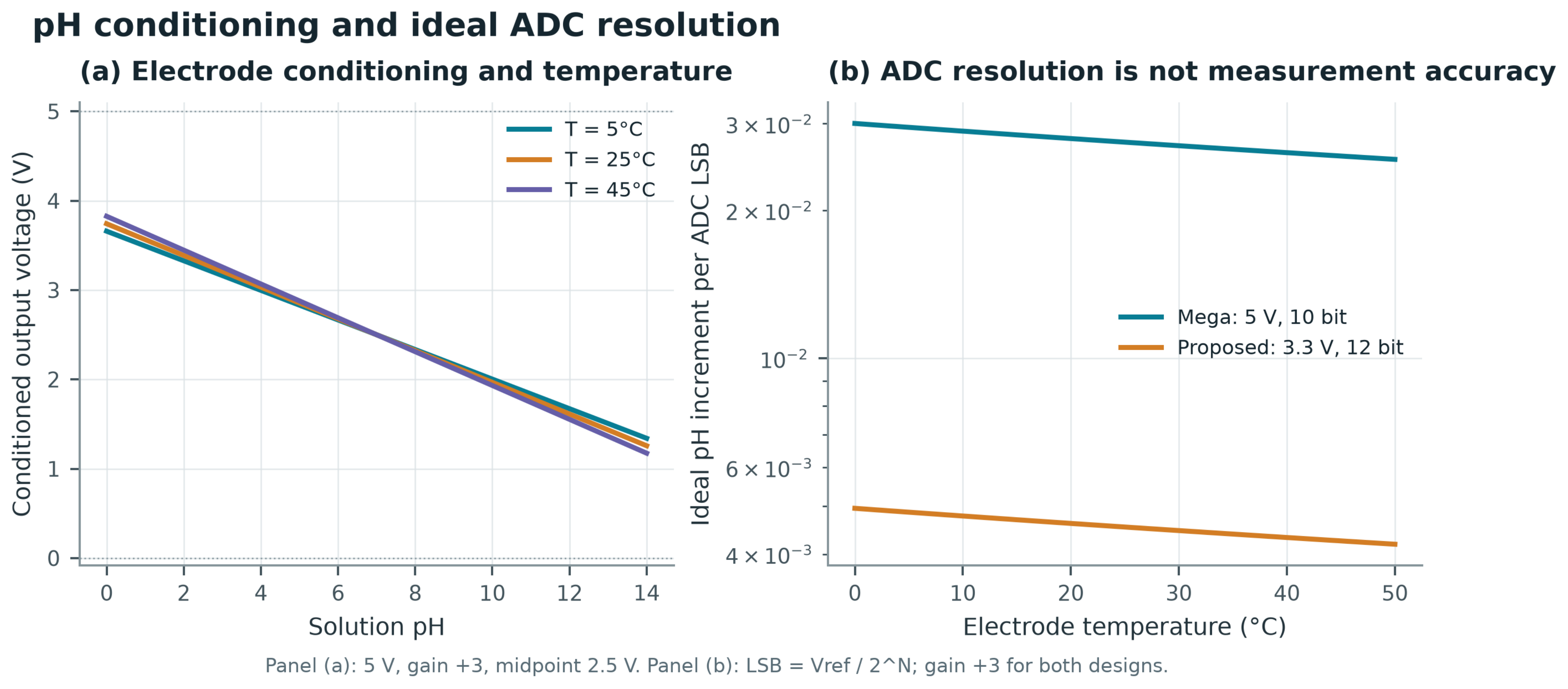}
\caption{Ideal pH-electrode slope and ADC quantization under the declared conditioning assumptions. ADC LSB width is \(V_{\mathrm{ref}}\)/2\textsuperscript{N}; the corresponding pH increment is \(V_{\mathrm{ref}}\)/[2\textsuperscript{N} G S(T)]. Quantization resolution is distinct from calibrated measurement accuracy.}
\label{fig:25}
\end{figure}

Turbidity measurements require local calibration against appropriate standards, control of bubbles and a documented optical path. Fouling and changes in particle properties can alter response between deployments; published low-cost instruments consequently assess calibration and maintenance together \citep{trevathan2020_3d3a03}, \citep{wang2024_112661}. The unidentified rover module cannot inherit their coefficients. Likewise, the GUVA identifier alone does not define the complete amplifier gain, spectral correction or angular response needed for a defensible ultraviolet index. Its board identity and calibration must be established before converting voltage into an exposure metric \citep{dfrobotnd_97cee5}.

\begin{figure}[!htbp]
\centering
\includegraphics[page=8,width=5.890in,height=3.800in,keepaspectratio]{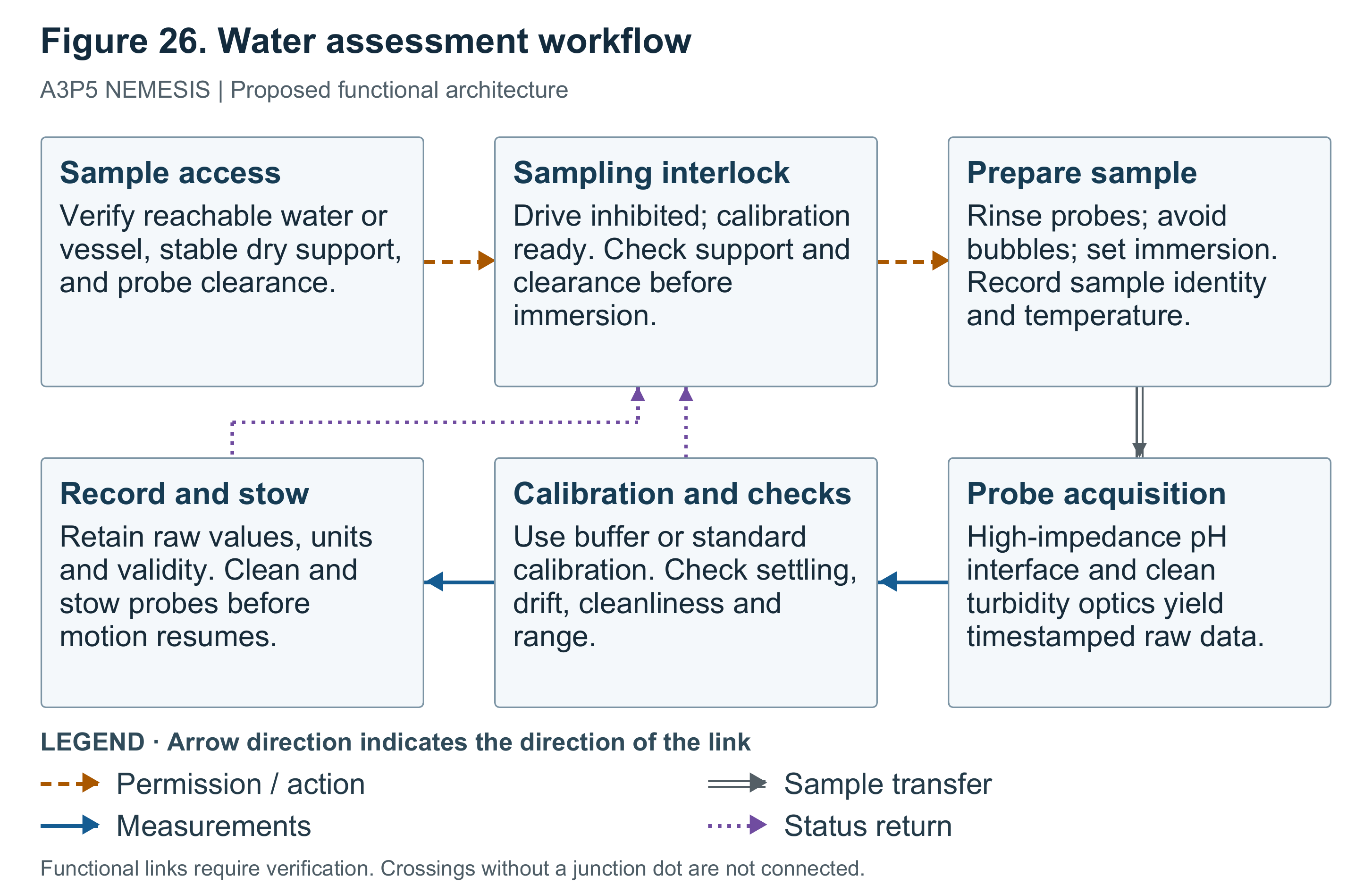}
\caption{Water-assessment workflow linking probe conditioning, reference checks, stabilization, pH/turbidity acquisition and sample identity.}
\label{fig:26}
\end{figure}

\begin{figure}[H]
\centering
\includegraphics[page=7,width=6.200in,height=4.000in,keepaspectratio]{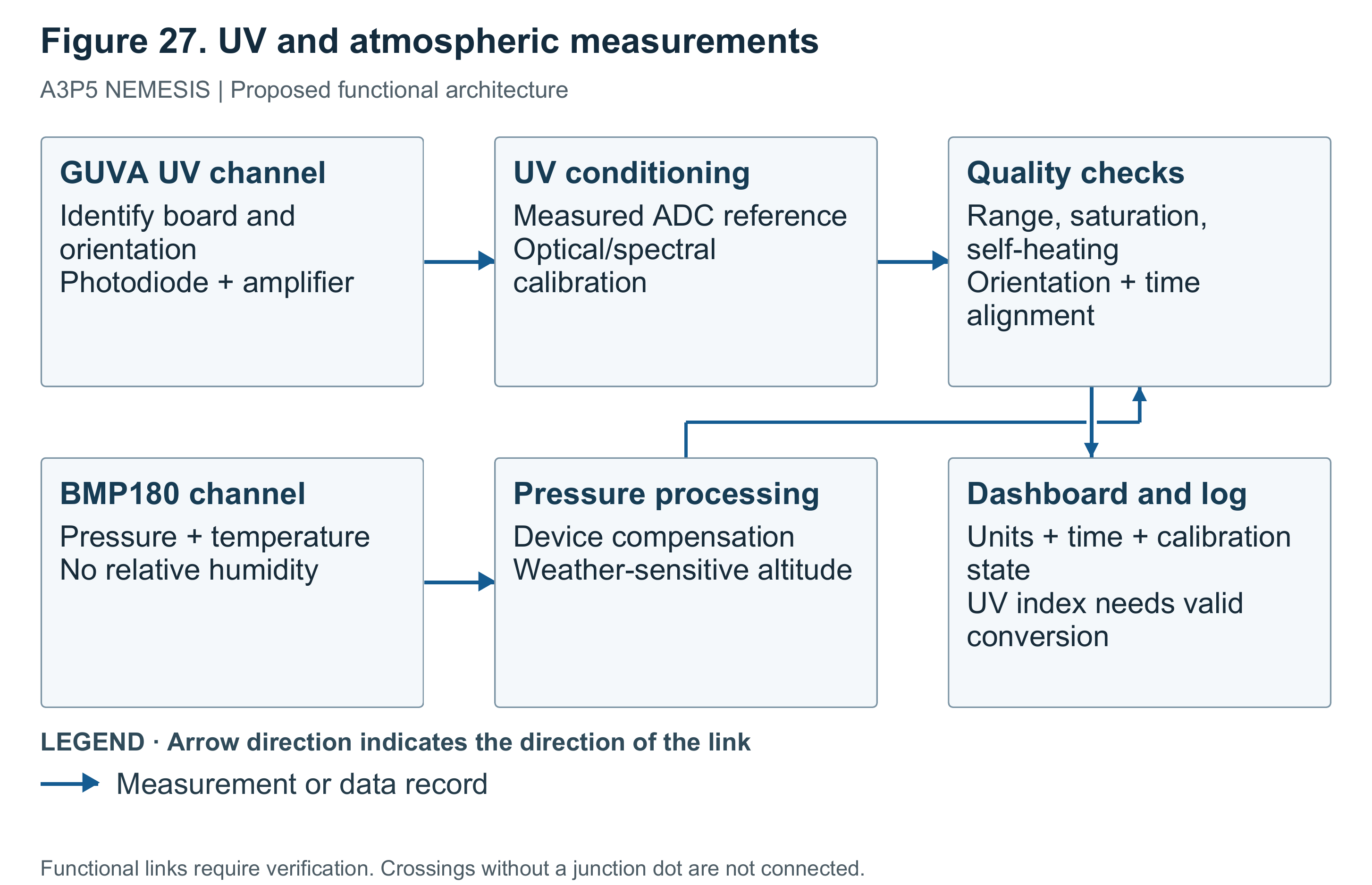}
\caption{Ultraviolet, pressure and temperature measurement interfaces. BMP180 provides no relative-humidity channel; ultraviolet conversion depends on the identified GUVA board and its calibration.}
\label{fig:27}
\end{figure}

For a calibrated measurand y=f(x), a first-order uncertainty budget can be written

\begin{equation}
\label{eq:13}
u_c^2(y)\approx J\Sigma_xJ^{\mathsf T}+u_{\mathrm{model}}^2.
\end{equation}

Here, J contains sensitivities of the measurement function, \(\Sigma_x\) is the covariance matrix of input uncertainties, and \(u_{\mathrm{model}}\) represents additional calibration-model uncertainty when treated as independent. The budget should distinguish reference uncertainty, repeatability, environmental correction and drift; correlated terms must not be counted twice. This is an assessment framework rather than a numerical uncertainty claim \citep{jcgm2008_0d3206}. A3P5 presently requires the underlying collocation and stability measurements before calibrated environmental performance can be stated.

\FloatBarrier

\section{Power, embedded control and communications}\label{sec:8}

\FloatBarrier

\subsection{Electrical power architecture}\label{sec:8.1}

\begin{figure}[!htbp]
\centering
\includegraphics[page=2,width=6.045in,height=3.900in,keepaspectratio]{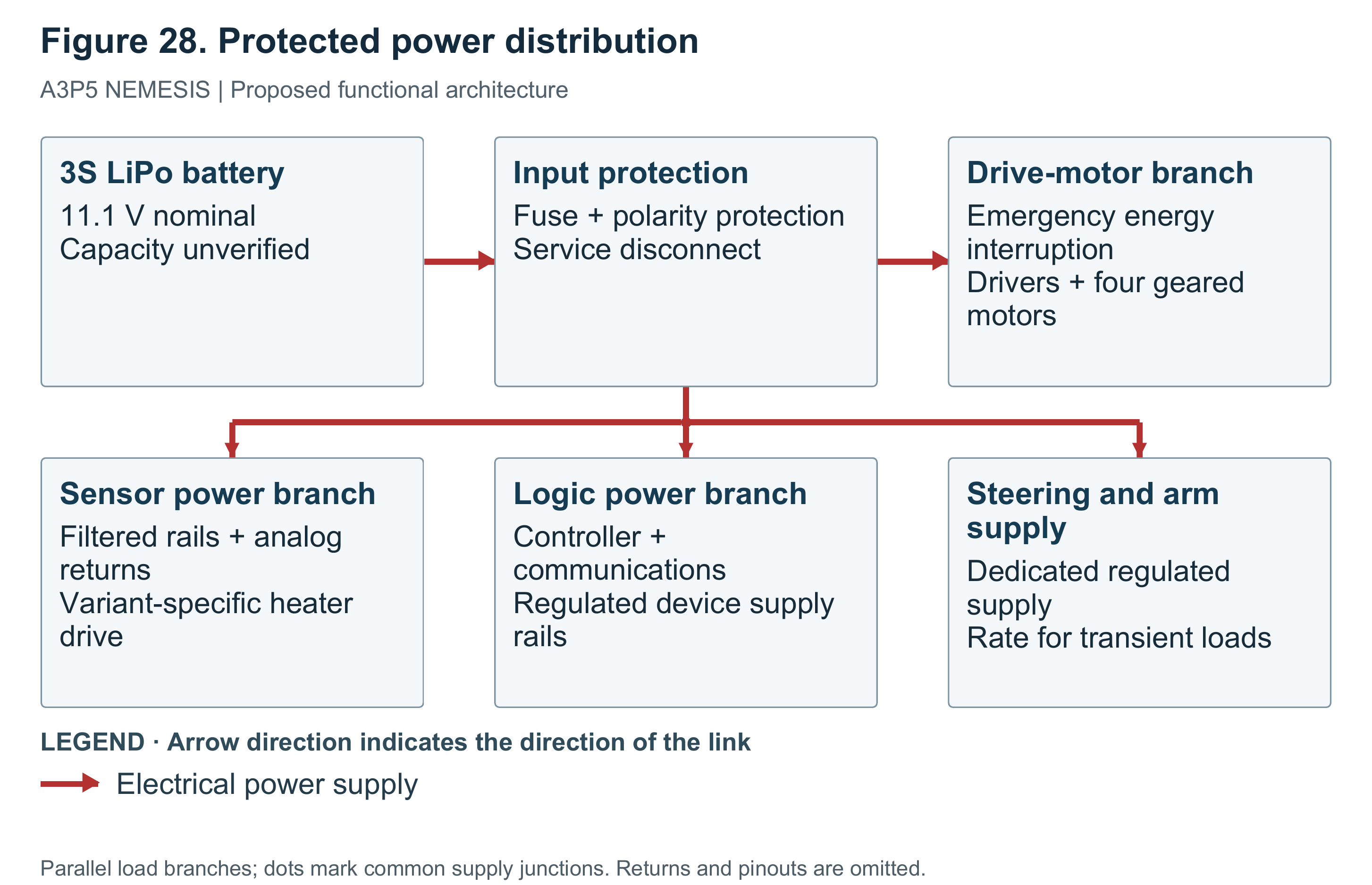}
\caption{Proposed power architecture with separate motor, servo, logic and sensor branches. Regulation, conductor sizing and protection require measured loads and component specifications.}
\label{fig:28}
\end{figure}

The project description identifies a typical 3S LiPo configuration at 11.1 V nominal. Battery capacity, discharge capability, protection thresholds and usable energy at the intended duty cycle are not documented. The design therefore separates battery selection from an illustrative energy calculation. A fused input and service disconnect feed distinct motor, servo, logic and sensor branches. High-current switching paths should be routed so that their voltage drops and transients do not become the reference for sensitive analog measurements. Figures 28 and 29 connect the proposed power-distribution architecture with the assumed energy-budget scenarios.

In the declared scenario, a 10 Ah battery, 0.85 depth-of-discharge factor and 0.90 available-energy derating yield 84.915 Wh. These factors are assumptions, not measured battery characterization. Mechanical traction power is divided by a separate battery-to-ground efficiency of 0.80 and added to an assumed battery-side auxiliary demand of 20 W. The energy derating and conversion efficiency act on different quantities. With battery capacity expressed in ampere-hours, usable energy is in watt-hours; dividing by battery power in watts gives endurance in hours.

\begin{equation}
\label{eq:14}
E_{\mathrm{use}}=V_{\mathrm{nom}}C_{Ah}f_{DoD}f_E
\end{equation}

\begin{equation}
\label{eq:15}
P_{\mathrm{bat}}=\frac{F_{\mathrm{req}}v}{\eta_d}+P_{\mathrm{aux}},\qquad t=\frac{E_{\mathrm{use}}}{P_{\mathrm{bat}}}
\end{equation}

The servo branch requires a regulator selected for simultaneous transient loads rather than average current alone. The gas-sensor heater supply should preserve the required waveform independently of ADC sampling. The logic branch needs a defined brownout response, and inter-board signals need voltage-compatible levels. A common reference may be necessary between boards, but high-current returns should not impose uncontrolled offsets on analog measurement ground.

At 0.25 m/s the calculated level-ground demand is 24.41 W, giving 3.48 h. A continuous 20\textdegree{} grade raises demand to 49.30 W and reduces the estimate to 1.72 h. A real route combines acceleration, stops, steering, slopes and sampling dwell, so neither value predicts a route. Current and voltage logging during representative missions should replace the constant-load approximation, and the battery should be characterized at the relevant temperature and discharge threshold.

\begin{figure}[!htbp]
\centering
\includegraphics[width=6.239in,height=2.600in,keepaspectratio]{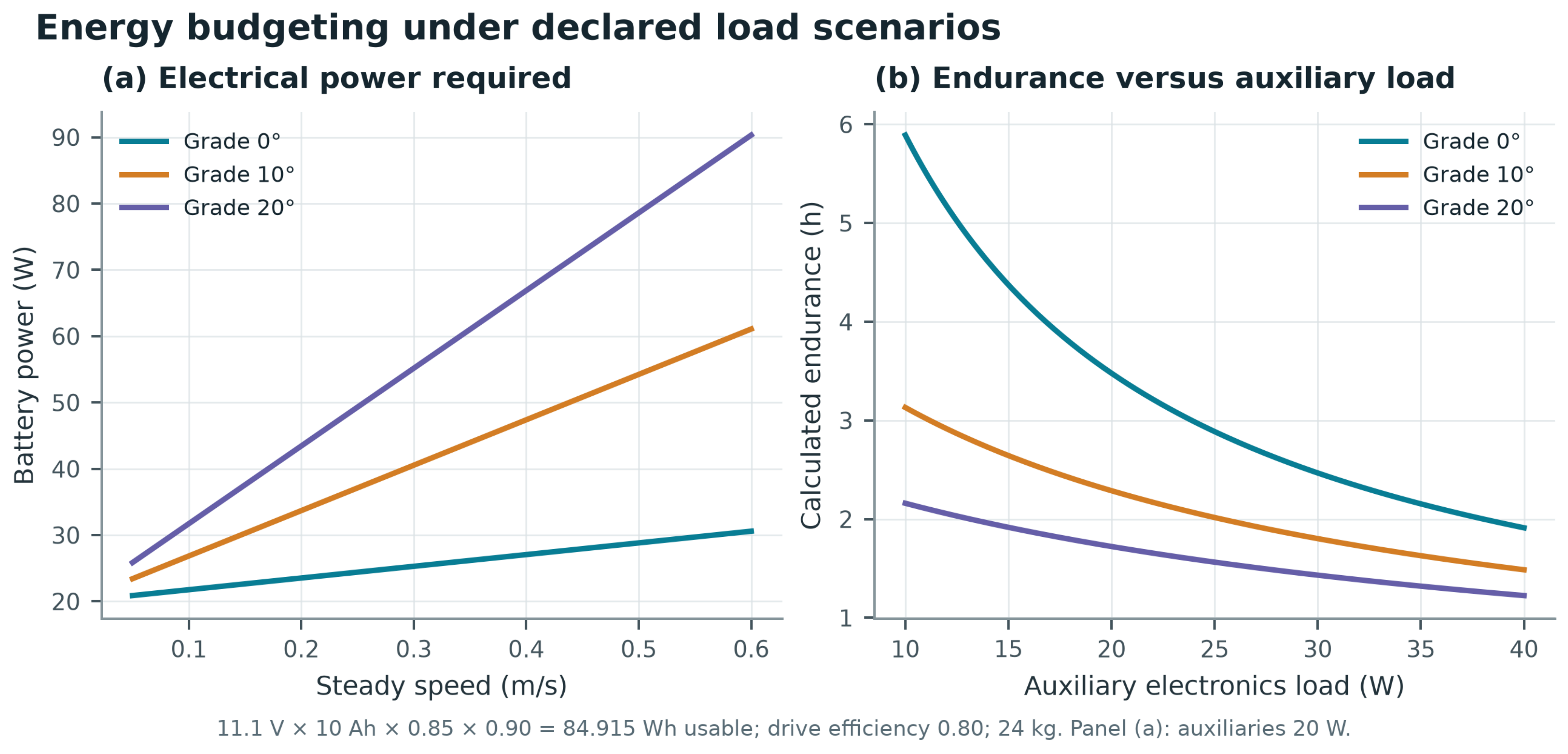}
\caption{Calculated endurance under declared battery, load and constant-grade assumptions. The model omits voltage sag, temperature-dependent capacity, intermittent arm motion and detailed motor efficiency maps. It is an energy budget, not a measured runtime.}
\label{fig:29}
\end{figure}

\FloatBarrier

\subsection{Scheduling and data integrity}\label{sec:8.2}

Low-level software should separate actuator timing from slower environmental acquisition and network traffic. A bounded control update should not wait for a web request or a slow sensor response. Each acquisition task should expose its latest valid value, acquisition timestamp and age rather than block the scheduler until a sensor is ready. Heater-cycle sensors require a state variable identifying when a reading is chemically and thermally interpretable. Delayed and invalid samples should remain visible in the log.

A useful record schema includes mission ID, monotonic sequence number, acquisition time, sensor ID, raw value, engineering value, unit, calibration version, validity flags, battery state and estimated pose with uncertainty. Monotonic time supports ordering within a run; wall-clock time supports comparison with reference instruments. Synchronization uncertainty should be recorded because assigning a delayed gas response to a precise coordinate can create a visually sharp but physically misleading map.

\FloatBarrier

\begingroup

\fontsize{9.5}{11.2}\selectfont

\renewcommand{\arraystretch}{1.16}

\setlength{\tabcolsep}{5pt}

\begin{longtable}{>{\raggedright\arraybackslash}p{\dimexpr 0.25\linewidth-2\tabcolsep\relax}>{\raggedright\arraybackslash}p{\dimexpr 0.29\linewidth-2\tabcolsep\relax}>{\raggedright\arraybackslash}p{\dimexpr 0.46\linewidth-2\tabcolsep\relax}}
\caption{Subsystem interfaces and verification requirements.}\label{tab:3}\\
\toprule
\textbf{Interface} & \textbf{Primary responsibility} & \textbf{Required checks} \\
\midrule
\endfirsthead
\multicolumn{3}{l}{\small Table \thetable\ continued}\\
\toprule
\textbf{Interface} & \textbf{Primary responsibility} & \textbf{Required checks} \\
\midrule
\endhead
\midrule
\multicolumn{3}{r}{\footnotesize Continued on the next page}\\
\endfoot
\bottomrule
\endlastfoot
Sensor to low-level controller & Acquire and timestamp raw data & Range, checksum, heater phase, warm-up, age \\
Low-level to communications & Publish actual state and faults & Sequence continuity, units, calibration version \\
Operator to arbiter & Request mode and bounded motion & Authority, freshness, limits, deliberate enable \\
Companion to low-level controller & Propose velocity or arm action & Expiration, feasibility, health conditions \\
Hardware protection to motor branch & Remove or inhibit drive energy & Independent wiring, latch state, reset \\
Logger to analysis pipeline & Retain reproducible evidence & Raw data, configuration and provenance \\
\end{longtable}

\endgroup

\FloatBarrier

\subsection{Local networking and operator authority}\label{sec:8.3}

A local access point allows a nearby operator to communicate without external Wi-Fi infrastructure. This is a connectivity arrangement, not a guarantee of range or resilience in a damaged building. Antenna placement, multipath, obstruction, competing traffic and power noise affect link quality. Video needs a separate bandwidth budget from command and health messages so that a large image stream cannot indefinitely delay a stop request or conceal stale state. The communications, dashboard and command-authority interfaces are summarized in Figure 30.

\begin{figure}[!htbp]
\centering
\includegraphics[page=11,width=6.355in,height=4.100in,keepaspectratio]{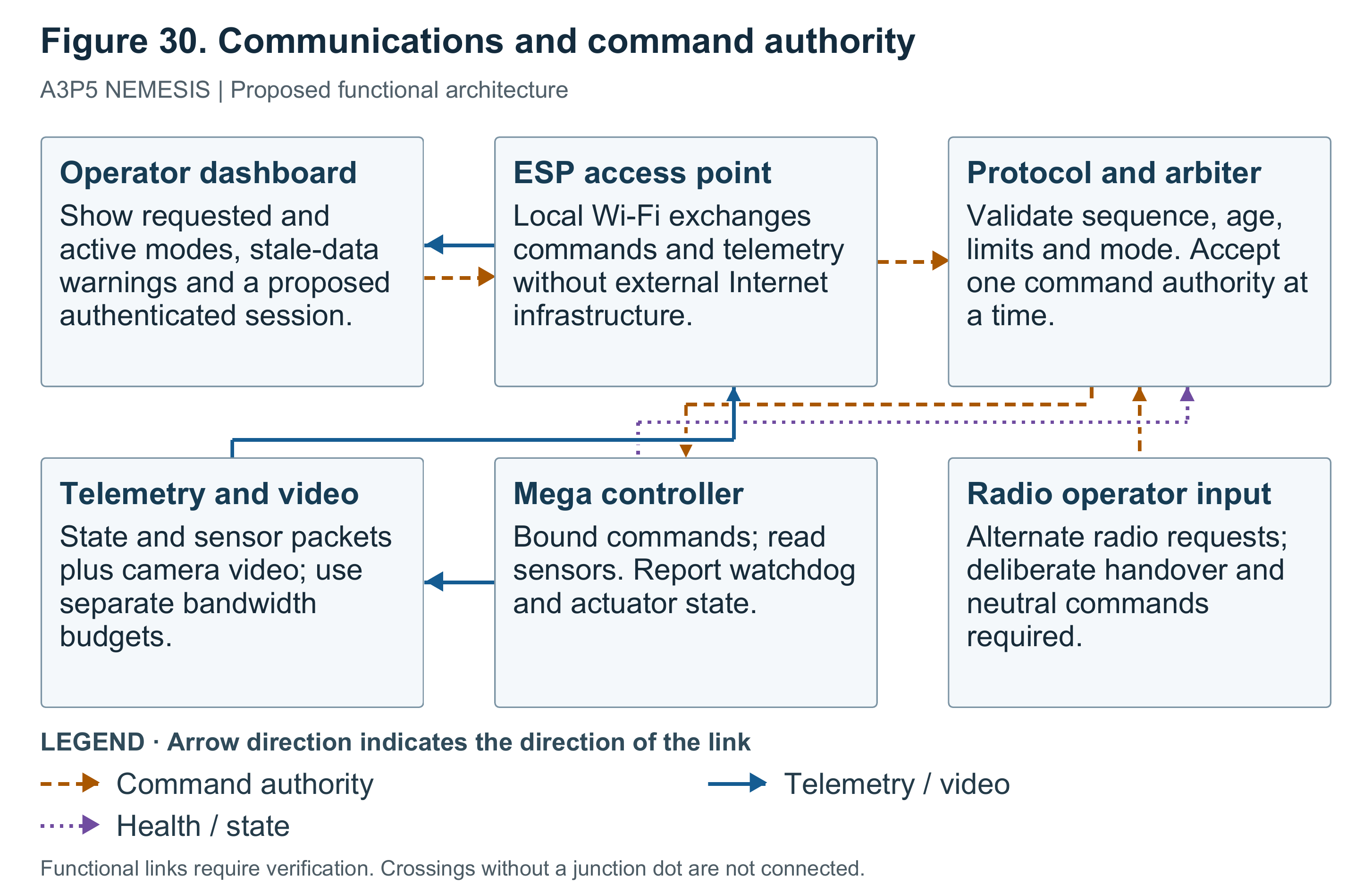}
\caption{Communications and operator-control architecture. Local Wi-Fi supports the dashboard, radio control supplies an alternate command source, and an arbiter rejects stale or conflicting requests.}
\label{fig:30}
\end{figure}

Manual, radio and autonomous requests should enter one arbiter with an explicit priority policy. A mode change should require neutral commands and a known actuator state. Commands must expire: an old packet received after an outage must not restart motion. The dashboard should distinguish requested mode from active mode and show the age of the last accepted command. Sensor validity and calibration status belong beside the value, preventing an uncalibrated indicator from appearing as an authoritative concentration.

\FloatBarrier

\section{Navigation, mission concepts and protection}\label{sec:9}

\FloatBarrier

\subsection{Proposed autonomy architecture}\label{sec:9.1}

Autonomous navigation requires pose estimation, obstacle representation, a planner respecting vehicle limits and a controller able to realize the motion. None follows automatically from mounting a camera. The proposed architecture adds wheel/steering feedback, an IMU and a range-sensing modality, with a companion processor where required. LiDAR, depth sensing and visual-inertial approaches should be compared under the intended lighting, texture, dust and compute constraints. Figures 31 and 32 relate the proposed navigation pipeline to the Blender patrol configuration.

ORB-SLAM3 and LIO-SAM illustrate different estimation architectures and sensor dependencies \citep{campos2021_973bd6}, \citep{shan2020_c48065}. Their performance does not transfer to NEMESIS without compatible sensors, calibration, timing, computing resources and evaluation. The dynamic-window approach connects local velocity choice with obstacle and braking constraints \citep{fox1997_7c8b94}, but a conventional implementation must be adapted to steering-rate limits and the rover footprint. The arm and mast expand the swept volume beyond the chassis rectangle.

Obstacles should be inflated by clearance incorporating body geometry, localization uncertainty and stopping distance. On degraded localization, a controlled stop may be preferable to continuing along a plausible route. Recovery logic should explicitly request assistance, return to a verified location if feasible, or remain stopped. A successful route in a simple virtual environment cannot establish robustness to reflective surfaces, missing returns, wheel slip or moving people.

\begin{figure}[!htbp]
\centering
\includegraphics[page=10,width=5.890in,height=3.800in,keepaspectratio]{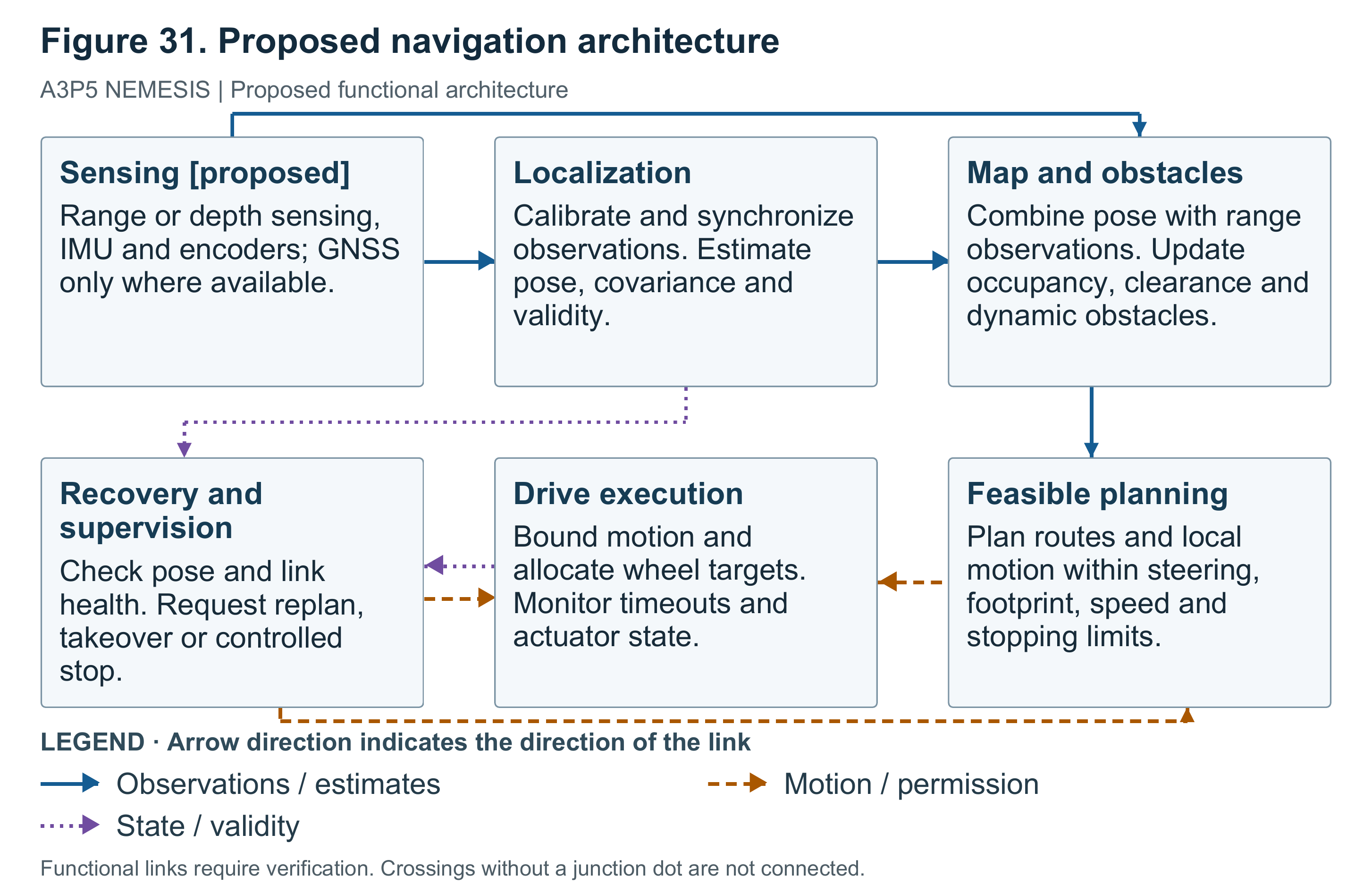}
\caption{Proposed navigation pipeline. Calibrated, synchronized observations support localization and mapping; planning accounts for clearance, braking and steering constraints.}
\label{fig:31}
\end{figure}

\begin{figure}[!htbp]
\centering
\includegraphics[width=4.608in,height=3.200in,keepaspectratio]{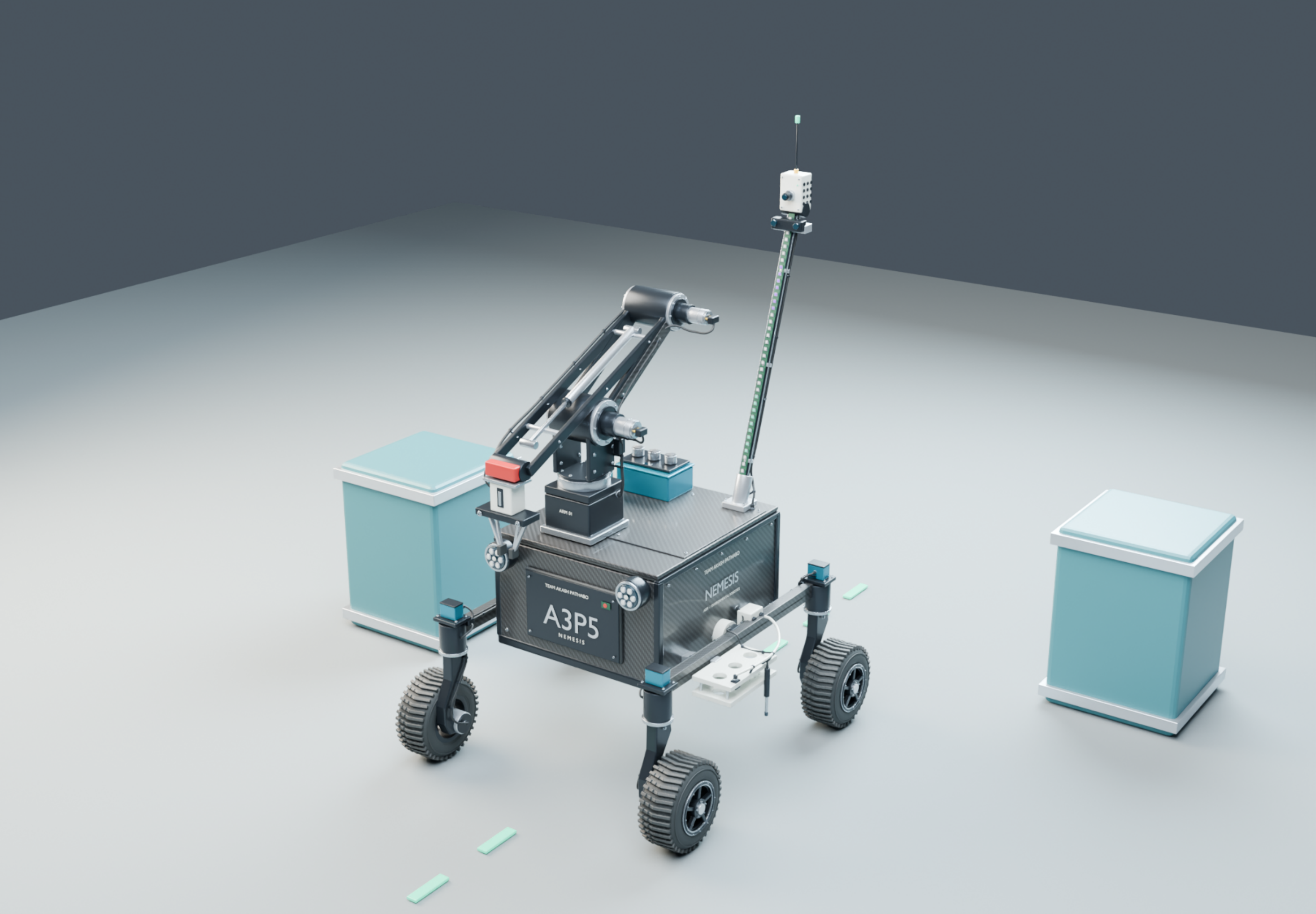}
\caption{Blender patrol concept incorporating proposed deck LiDAR, a humidity pod and a depth camera below the existing mast head. Route markings illustrate mission intent.}
\label{fig:32}
\end{figure}

\FloatBarrier

\subsection{Mission-level behaviour}\label{sec:9.2}

The proposed environmental-patrol sequence combines travel and observation: approach a station, reduce speed, confirm readiness, dwell as required, record the observation and resume travel. Industrial inspection prioritizes camera access and interpretable environmental indicators while preserving inlet clearance. Water assessment adds sample access and wet/dry boundaries. Disaster reconnaissance emphasizes visual information and cautious movement in uncertain terrain; it does not imply casualty transport or entry into explosive atmospheres. Figures 33--35 combine the mission-data workflow with industrial-observation and visual-reconnaissance illustrations.

\begin{figure}[!htbp]
\centering
\includegraphics[page=13,width=6.800in,height=4.387in,keepaspectratio]{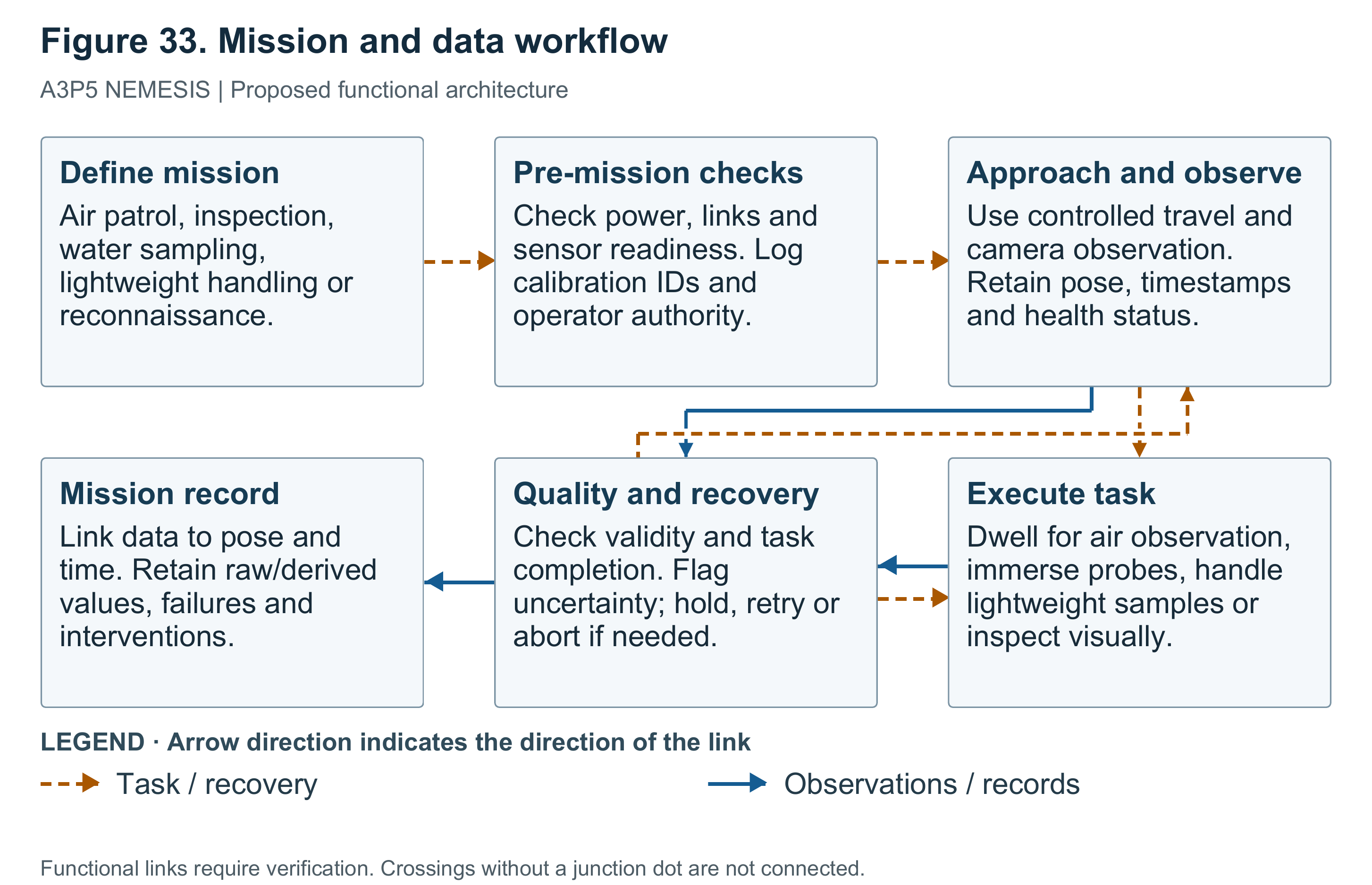}
\caption{Mission and data workflow for environmental patrol, industrial inspection, water assessment, lightweight handling and reconnaissance, including quality checks and recovery conditions.}
\label{fig:33}
\end{figure}

\begin{figure}[!htbp]
\centering
\includegraphics[width=4.968in,height=3.450in,keepaspectratio]{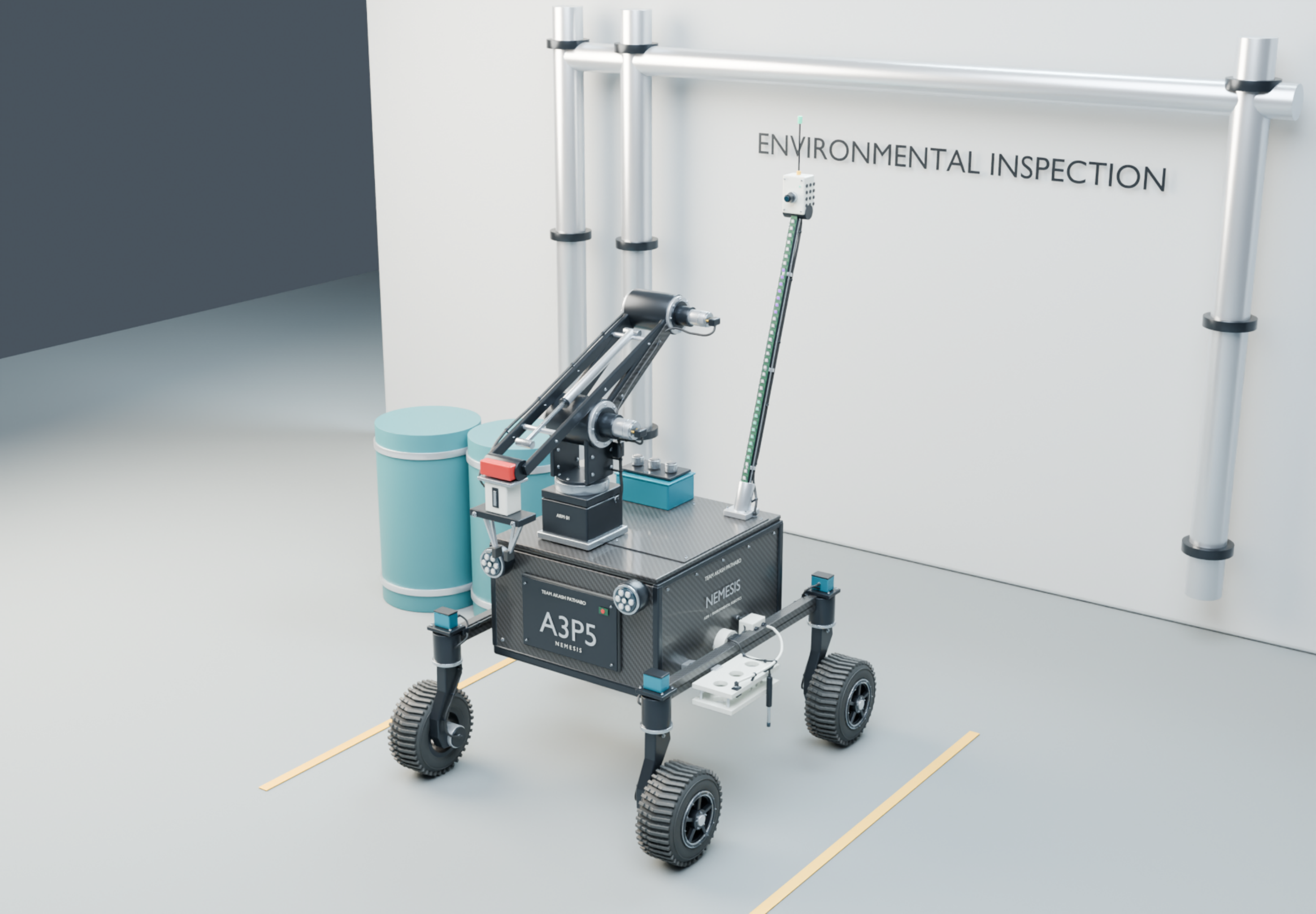}
\caption{Blender concept of stationary environmental observation near industrial piping. The scene illustrates component placement and inspection context.}
\label{fig:34}
\end{figure}

Each task needs a measurable completion condition. An air observation is complete only when a valid sample with adequate timing/context is recorded; reaching a waypoint is insufficient. Manipulation needs confirmation that the object was grasped and placed, not merely that motor commands were issued. Inspection should record view coverage and unresolved occlusions. Abort and recovery events should remain in the dataset, because suppressing them makes the system appear more reliable than it is.

\begin{figure}[!htbp]
\centering
\includegraphics[width=5.832in,height=4.050in,keepaspectratio]{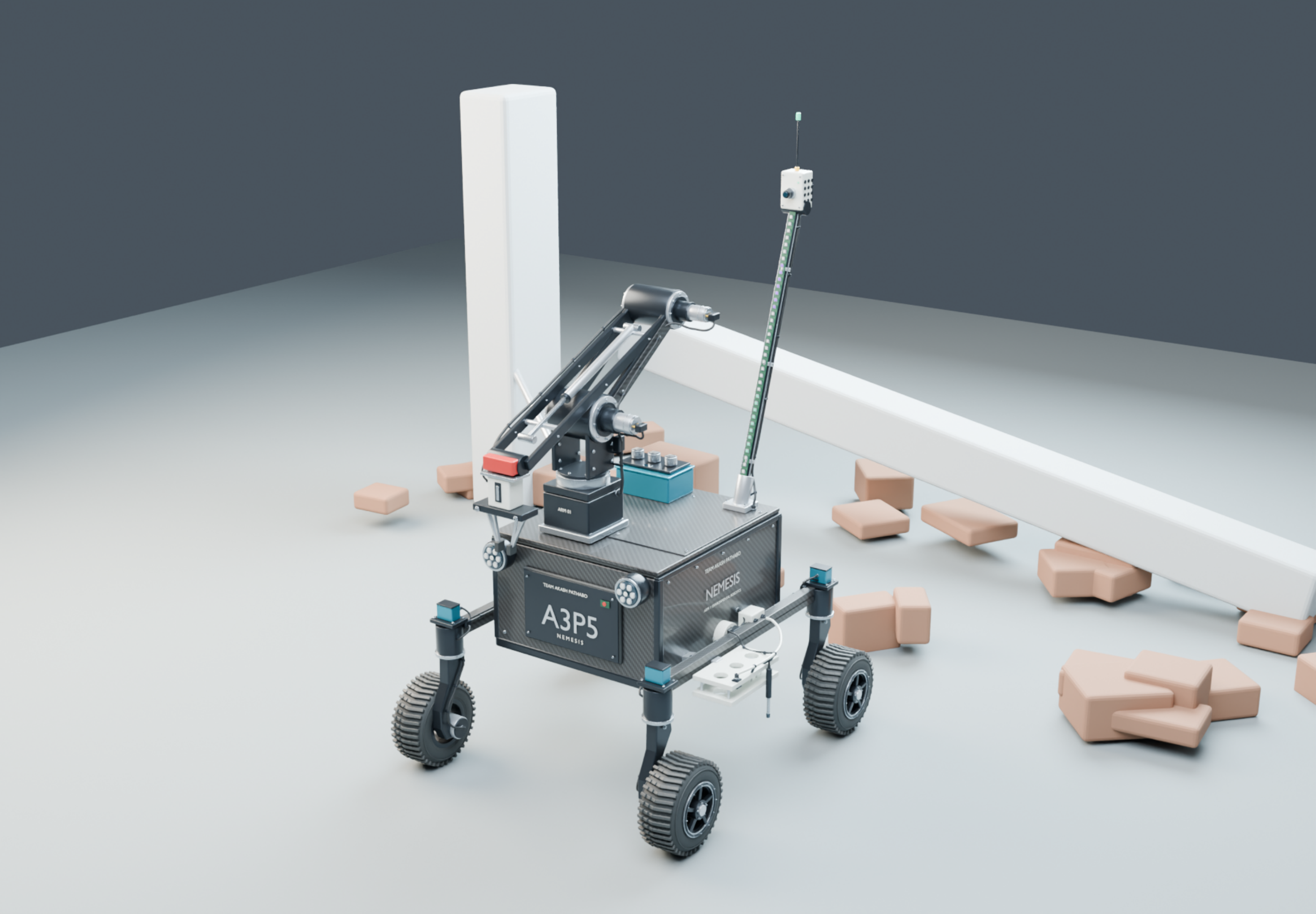}
\caption{Blender visual-reconnaissance concept near debris, with all wheels on a clear support surface. The arrangement is not a rubble-traversal test.}
\label{fig:35}
\end{figure}

\FloatBarrier

\subsection{Stopping and fault handling}\label{sec:9.3}

The simplified stopping-clearance model combines command latency, braking distance and a geometric allowance. Available deceleration is a ground-level quantity to measure on relevant surfaces and slopes, not an unloaded motor specification. Obstacle motion, slip and the run-down of a hardware power cut may require additional clearance.

\begin{equation}
\label{eq:16}
d_{\mathrm{clear}}=vT_{\mathrm{latency}}+\frac{v^2}{2a_{\mathrm{brake}}}+d_{\mathrm{margin}}
\end{equation}

\begin{figure}[!htbp]
\centering
\includegraphics[width=5.407in,height=2.600in,keepaspectratio]{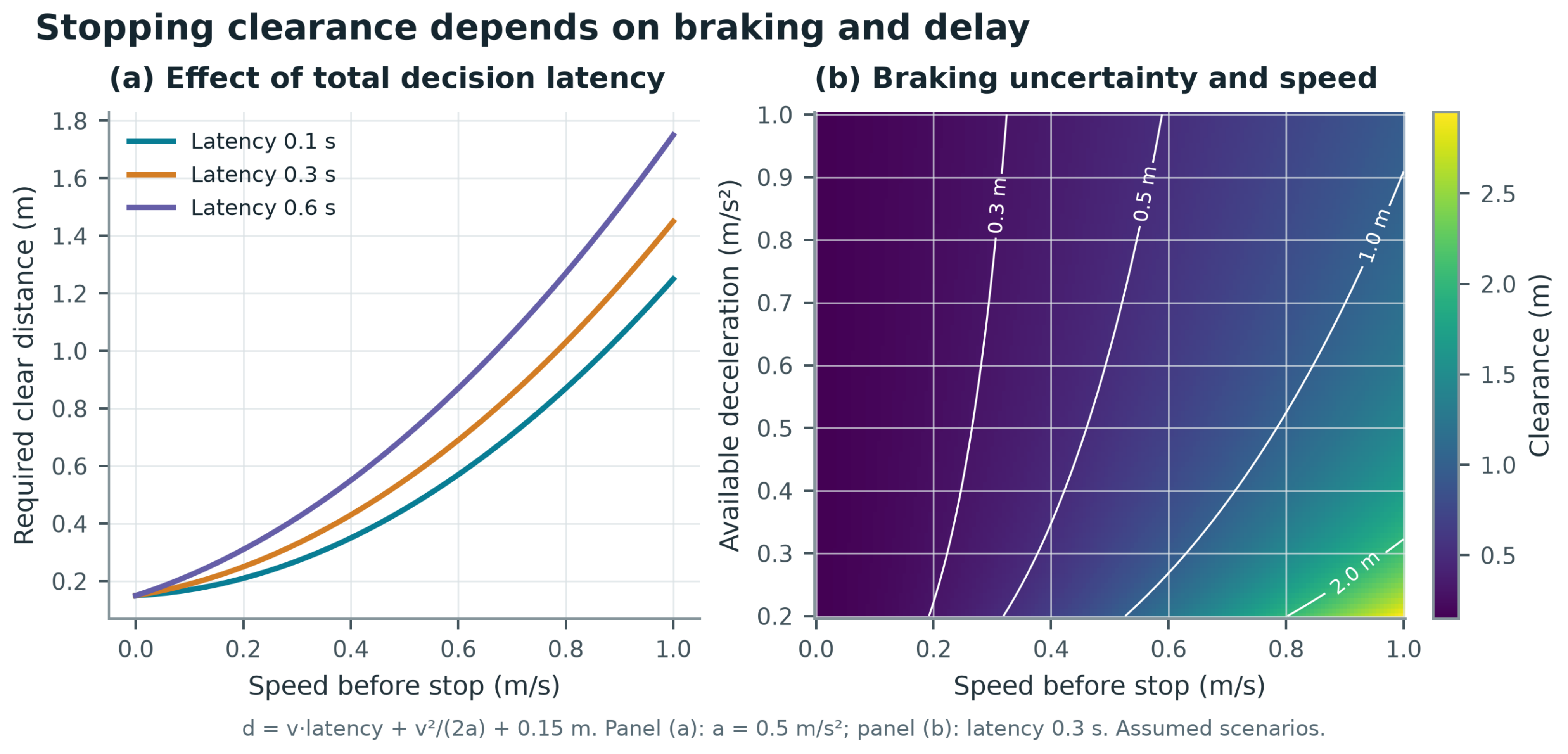}
\caption{Calculated stopping clearance across assumed speed, delay and deceleration scenarios, including a 0.15 m allowance. The curves define quantities for braking trials and do not establish a certified protective distance.}
\label{fig:36}
\end{figure}

A software stop and emergency energy interruption serve different purposes. A controlled stop can regulate deceleration and preserve a safe arm state. A hardware emergency-stop circuit needs an independently defined effect on actuator energy and a deliberate reset. Removing all power indiscriminately may let a gravity-loaded joint move, so the mechanism needs a holding or parking strategy. These functions are specified at architecture level; no safety-standard compliance is claimed. Figures 36 and 37 connect the stopping-distance scenarios with the proposed operating and protective states.

At 0.4 m/s, 0.3 s latency, 0.5 m/s\textsuperscript{2} deceleration and 0.15 m margin, calculated clearance is 0.430 m. Raising speed to 1.0 m/s increases it to 1.450 m. The quadratic term explains why speed restriction can matter more than a small improvement in communication delay. Fault injection should test stale commands, receiver loss, low voltage, sensor dropout, encoder inconsistency and controller reset, recording both detection time and actual stopping motion.

\begin{figure}[!htbp]
\centering
\includegraphics[page=12,width=6.045in,height=3.900in,keepaspectratio]{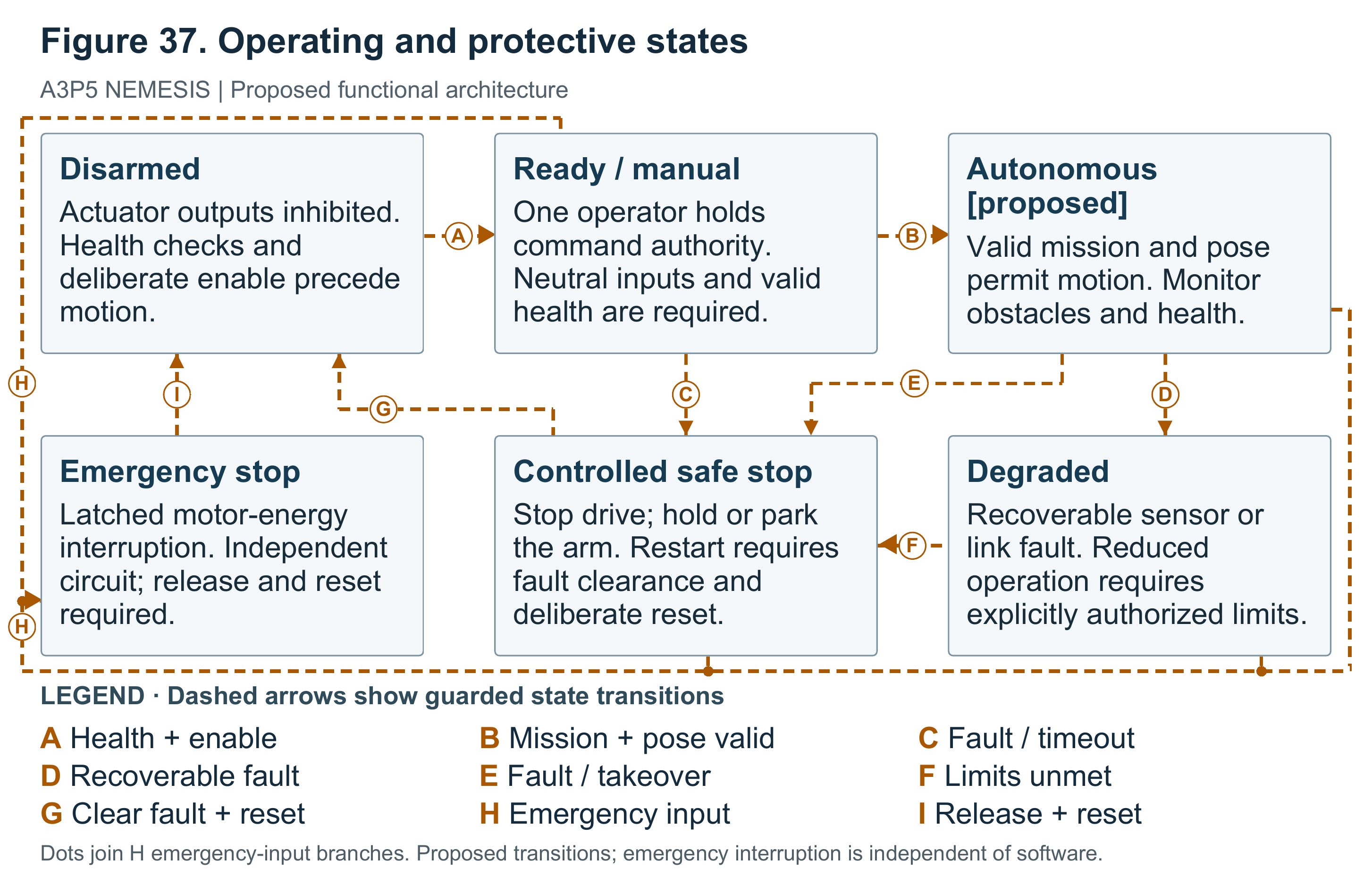}
\caption{Operating and protective states. Transitions require explicit health, authority and reset conditions; hardware energy interruption remains distinct from a controlled software stop.}
\label{fig:37}
\end{figure}

\FloatBarrier

\section{External-data machine-learning calibration benchmark}\label{sec:10}

\FloatBarrier

\subsection{Dataset, task definition and eligibility}\label{sec:10.1}

A reproducible external-data experiment was conducted to examine calibration choices before rover-specific training data become available. The UCI Air Quality archive contains fixed-station observations from a multisensor electronic nose and reference analysers \citep{vito2008_026a72}. The associated publication concerns benzene calibration \citep{devito2008_1548b8}; the present experiment defines a separate carbon-monoxide regression task. Its results support methodological development, not a claim that the rover's MQ devices have been calibrated or deployed successfully. Figure 38 documents the time coverage, observation eligibility and distribution shifts of the external dataset.

The target was CO(GT), in mg m\textsuperscript{-3}. Exactly eight predictors were permitted: PT08.S1(CO), PT08.S2(NMHC), PT08.S3(NOx), PT08.S4(NO2), PT08.S5(O3), temperature T, relative humidity RH and absolute humidity AH. Channel labels identify supplied sensor responses, not perfectly selective analyte measurements. All other reference-gas columns were excluded. Timestamps served partitioning and visualization only; neither target lags nor time-derived predictors were introduced. Thus, prediction uses the contemporaneous sensor array and recorded environment without access to additional reference-analyser outputs.

The parsed file contained 9,471 rows. Removing 114 rows without valid timestamps left 9,357 unique timestamped observations. The sentinel \ensuremath{-}200 was converted to missing. Excluding 1,683 unavailable CO targets and another 330 observations with all predictors missing yielded 7,344 eligible observations. The retained predictors contained no missing cells, so the prespecified training-median imputer was inert. Targets were never imputed, and observations were neither interpolated nor concentration-filtered. Actual timestamps span 10 March 2004, 18:00, to 4 April 2005, 14:00; this auditable file range is retained despite differences from the repository summary. Dataset-local timestamps are used without inferring a timezone.

\begin{figure}[!htbp]
\centering
\includegraphics[width=6.800in,height=4.795in,keepaspectratio]{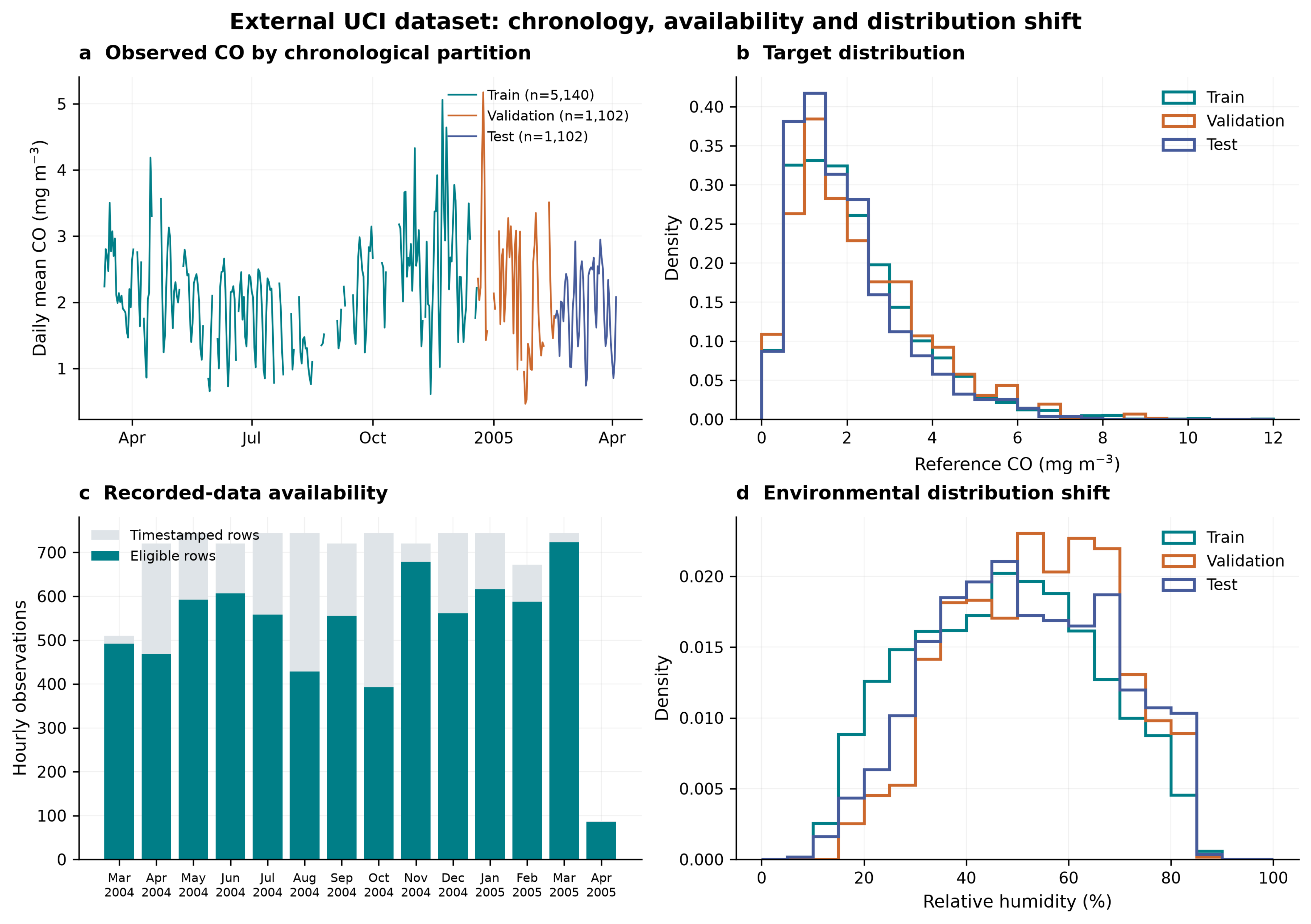}
\caption{External-data coverage and distributions: (a) daily CO means by chronological partition, requiring at least 12 observed hours; (b) separately normalized CO distributions; (c) monthly timestamped and eligible counts; (d) separately normalized RH distributions. Missing periods remain unfilled. Original analysis of Vito \citep{vito2008_026a72}, UCI Air Quality, \url{https://doi.org/10.24432/C59K5F}, CC BY 4.0, \url{https://creativecommons.org/licenses/by/4.0/}.}
\label{fig:38}
\end{figure}

\FloatBarrier

\subsection{Chronological evaluation and model selection}\label{sec:10.2}

Chronologically ordered eligible rows were partitioned at floor(0.70n) and floor(0.85n). Training comprised 5,140 observations from 10 March 2004, 18:00, through 19 December 2004, 17:00. Validation comprised 1,102 observations from 19 December 2004, 18:00, through 16 February 2005, 13:00. Testing comprised the subsequent 1,102 observations through 4 April 2005, 14:00. Separating later periods limits leakage from randomly mixing temporally dependent observations, although it cannot establish geographic transfer \citep{roberts2017_8ac43f}, \citep{kapoor2023_8dd715}. The ordered calibration-analysis stages are summarized in Figure 39.

\Needspace{9\baselineskip}

Four model families were compared: a training-mean baseline, standardized ridge regression, random forest and histogram gradient boosting. Ridge solves

\begin{equation}
\label{eq:17}
\begin{aligned}(\widehat{\beta}_0,\widehat{\symbfit{\beta}})=\arg\min_{\beta_0,\symbfit{\beta}}\Bigl[&\sum_{i\in\mathcal{T}}(y_i-\beta_0-\mathbf{z}_i^{\mathsf{T}}\symbfit{\beta})^2\\&+\alpha\lVert\symbfit{\beta}\rVert_2^2\Bigr].\end{aligned}
\end{equation}

Here, \ensuremath{\mathcal{T}} denotes training observations, \(z_{i}\) denotes predictors standardized using training means and standard deviations, and \ensuremath{\alpha} controls coefficient shrinkage. The intercept is unpenalized. This provides a compact linear comparator to nonlinear tree ensembles; random forests aggregate randomized trees to address more complex response surfaces \citep{breiman2001_75dc04}.

The baseline was the mean CO of the training partition, rather than a mean recomputed from the later test period. Similarly, standardization statistics were estimated before validation or test transformation. Retaining these operations inside each saved pipeline ensures that evaluation applies the stored training statistics. The candidate grid was intentionally finite and common to the recorded protocol. Its outcome compares these specified configurations; it does not establish the best possible algorithm or exhaust every nonlinear calibration strategy.

\begin{figure}[!htbp]
\centering
\includegraphics[page=14,width=6.045in,height=3.900in,keepaspectratio]{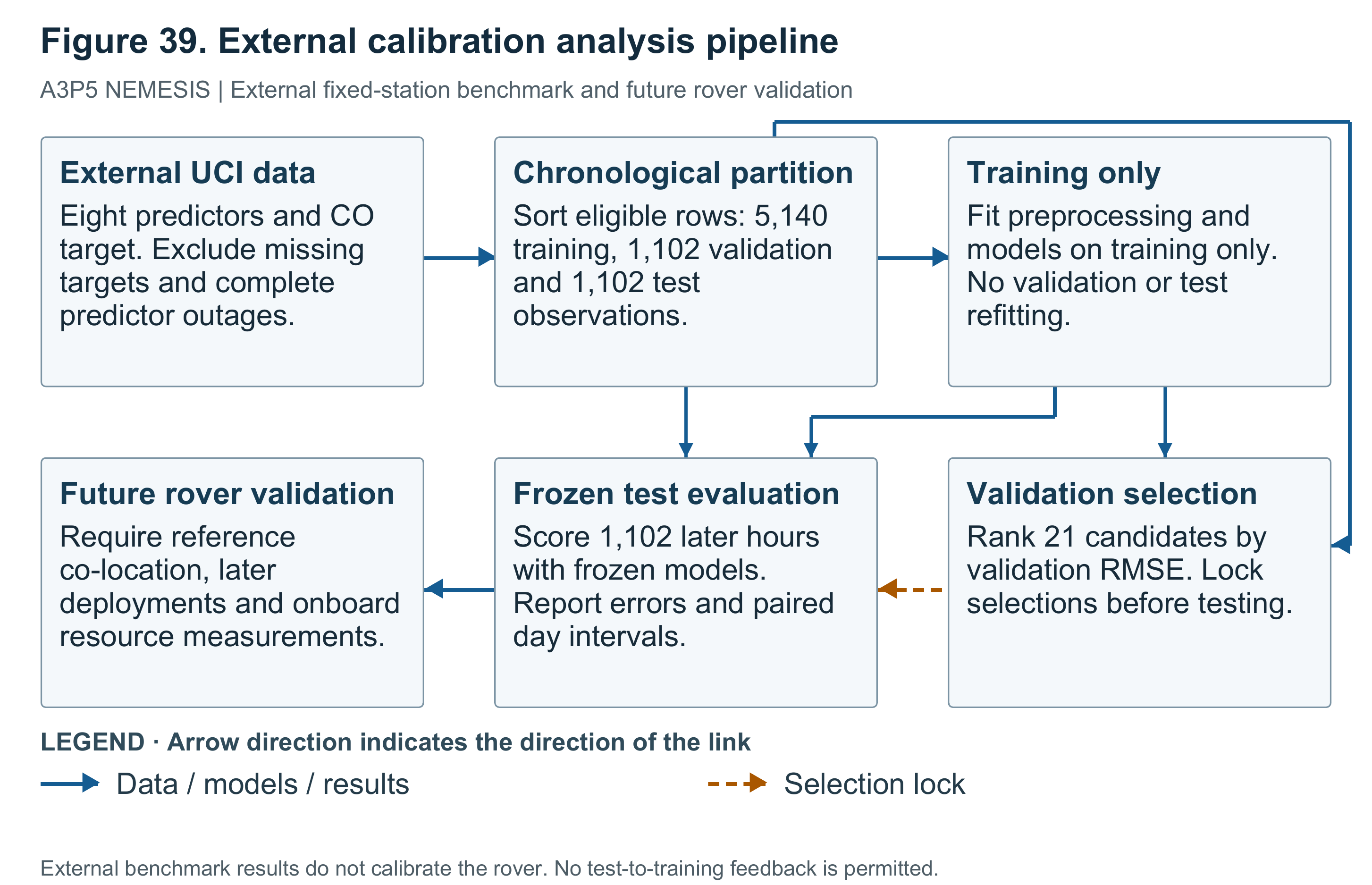}
\caption{Calibration-analysis pipeline separating cleaning, chronological partitioning, training-only preprocessing, validation selection and frozen testing. External-data results require subsequent rover-specific validation.}
\label{fig:39}
\end{figure}

\par\noindent\begin{minipage}{\linewidth}

\begin{lstlisting}[caption={Chronological model-selection and frozen-test evaluation core. The runnable companion supplies the NumPy/scikit-learn imports, original-data cleaning and fixed candidate grid. No training-plus-validation refit is performed.},label={lst:evaluation}]

def temporal_benchmark(frame, features, target, candidates):
    data = frame.sort_values("timestamp", kind="stable")
    n = len(data)
    a, b = int(0.70 * n), int(0.85 * n)
    train, valid, test = data.iloc[:a], data.iloc[a:b], data.iloc[b:]
    Xtr, ytr = train[features], train[target].to_numpy()
    Xva, yva = valid[features], valid[target].to_numpy()
    selected, search = {}, []
    for family, prototype in candidates:
        steps = [("imputer", SimpleImputer(strategy="median"))]
        if family == "Ridge":
            steps.append(("scaler", StandardScaler()))
        model = Pipeline(steps + [("regressor", clone(prototype))])
        model.fit(Xtr, ytr)
        score = np.sqrt(mean_squared_error(yva, model.predict(Xva)))
        search.append((family, float(score)))
        previous = selected.get(family, (np.inf, None))[0]
        if score < previous:  # Retain the first candidate on exact ties.
            selected[family] = (float(score), model)
    winner = min(selected, key=lambda name: selected[name][0])
    Xte, yte = test[features], test[target].to_numpy()
    predictions = {name: fit.predict(Xte)
                   for name, (_, fit) in selected.items()}
    return winner, selected, search, predictions, yte

\end{lstlisting}

\end{minipage}\par

Twenty-one candidate configurations were evaluated. Ridge used \ensuremath{\alpha}\ensuremath{\in}\{0.01,0.1,1,10,100,1000\}. Forests used 200 trees, all predictors available at each split, depths of 10 or unrestricted, and minimum leaf sizes of 1, 5 or 15. Boosting used 250 iterations, minimum leaf size 20, learning rates 0.05 or 0.1, 15 or 31 leaf nodes, and L2 penalties of 0 or 1; internal early stopping was disabled. All preprocessing and fitting used training data only. Minimum validation RMSE determined each family's configuration and the overall choice, with the first candidate resolving exact ties. Selections were saved before test evaluation; no training-plus-validation refit or test-based retuning occurred. The random seed was 20260915.

Listing 1 exposes the evaluation core used for the external-data benchmark. Its inputs are the eligible observations defined in Section 10.1, the eight permitted predictors and the ordered sequence of 21 prespecified candidate estimators. Each candidate is cloned before fitting. The pipeline estimates the imputer statistics and, for Ridge, the scaling statistics from the training partition alone; validation observations are transformed by those stored operations. The strict less-than comparison preserves the first candidate when validation scores tie. All family choices and the overall winner are frozen before any test predictions are produced. Returning predictions for every selected family preserves the complete comparison without using the test ranking to revise the winner.

The executable companion reparses the archived CSV, applies the documented eligibility rule and constructs the exact candidate sequence. Re-execution reproduced all 21 saved validation RMSE values and every family’s four test metrics to an absolute tolerance of 10\textsuperscript{-12}. The selected family remained Ridge. Calendar-day resampling is applied afterward to these frozen hourly predictions, as specified in Section 10.3; it does not refit the pipelines or modify the selection. This implementation provides an inspectable reference for the reported protocol, while deployment on the rover still requires new reference measurements and validation on the intended hardware.

\FloatBarrier

\Needspace{9\baselineskip}

\subsection{Metrics and uncertainty procedure}\label{sec:10.3}

\begin{figure}[H]
\centering
\includegraphics[width=5.021in,height=5.000in,keepaspectratio]{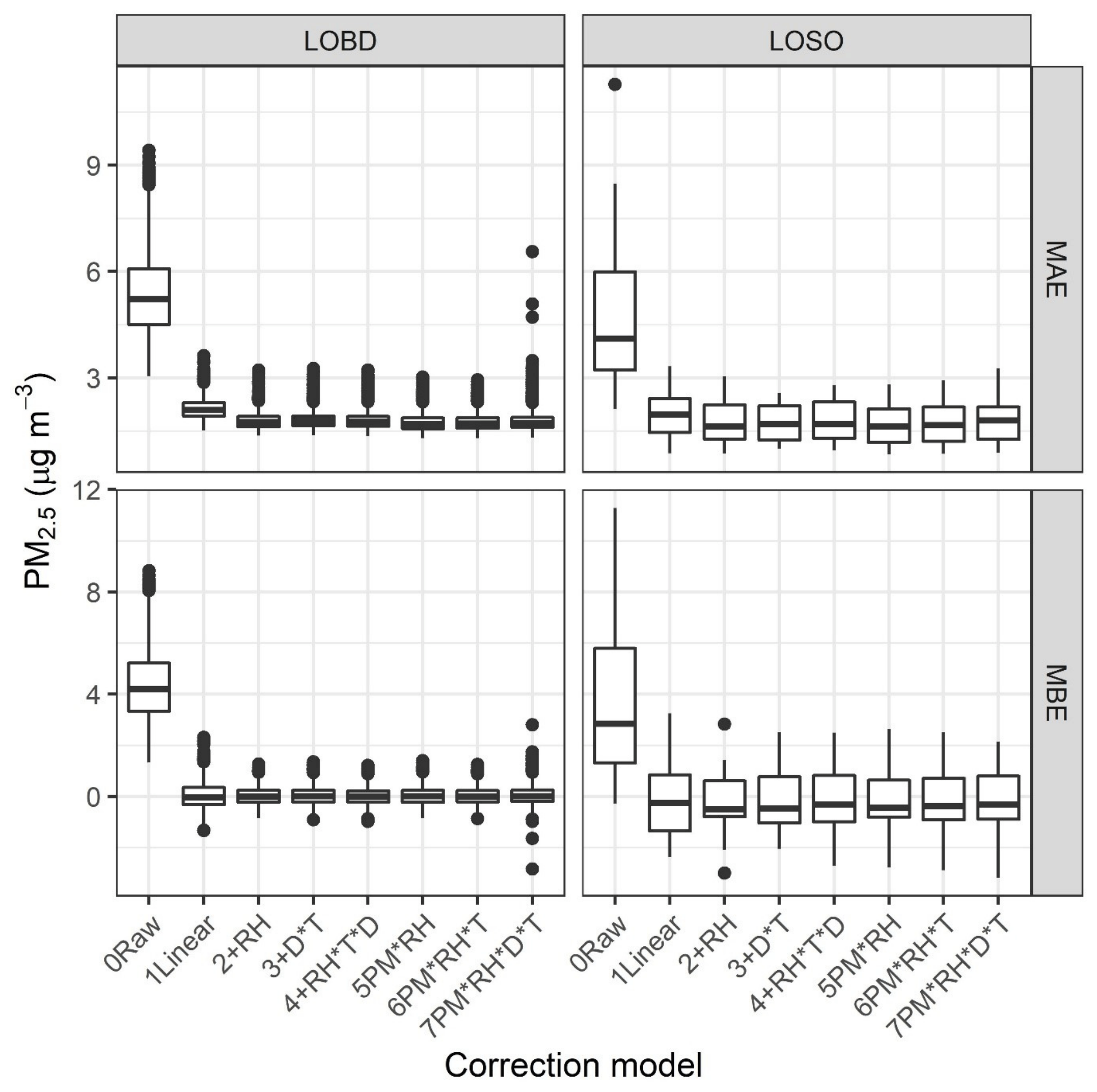}
\caption{External PurpleAir calibration errors under leave-out-by-date (LOBD) and leave-one-state-out (LOSO) evaluation. Panels summarize mean absolute and mean bias errors for raw and corrected PM2.5. Reproduced unchanged from Barkjohn et al. \citep{barkjohn2021_7ba5aa}, Fig. 4, CC BY 4.0. Source: \url{https://doi.org/10.5194/amt-14-4617-2021}. Licence: \url{https://creativecommons.org/licenses/by/4.0/}.}
\label{fig:40}
\end{figure}

\Needspace{12\baselineskip}

For n held-out observations, errors were evaluated using

\begin{equation}
\label{eq:18}
\begin{gathered}
\mathrm{MAE}=\frac1n\sum_i|\widehat y_i-y_i|, \\ \mathrm{RMSE}=\sqrt{\frac1n\sum_i(\widehat y_i-y_i)^2}, \\ R^2=1-\frac{\sum_i(\widehat y_i-y_i)^2}{\sum_i(y_i-\overline y)^2}.
\end{gathered}
\end{equation}

Here, \(y_{i}\) denotes reference CO and its hatted counterpart denotes the prediction; the reference mean is calculated over the scored observations. MAE and RMSE retain mg m\textsuperscript{-3} units; R\textsuperscript{2} is dimensionless. Mean prediction-minus-reference residual reports directional bias. Predictions were not clipped. The published withheld-date and withheld-state comparison in Figure 40 illustrates why evaluation partitions must match the intended deployment claim.

The test block spans 48 observed calendar days, with 10--24 eligible hours per day. Two thousand bootstrap replicates sampled calendar-day blocks with replacement, retaining every eligible observation within each selected day and using the same sampled indices across model families. Variable-length days retained their available hours without interpolation. The 2.5th and 97.5th percentiles form conditional score intervals. Within-day dependence is preserved, but dependence across days, fitted-model uncertainty and missing-data uncertainty are not incorporated. These intervals therefore describe variability of frozen-model scores under this resampling scheme, not prediction intervals for future readings. Published calibration comparisons using withheld dates and locations provide complementary evidence that the validation design affects the question answered \citep{barkjohn2021_7ba5aa}.

\FloatBarrier

\subsection{Held-out results and diagnostic interpretation}\label{sec:10.4}

Ridge with \ensuremath{\alpha}=10 achieved the lowest validation RMSE, 0.6853 mg m\textsuperscript{-3}, compared with 0.6900 for random forest, 0.8202 for boosting and 1.5136 for the mean baseline. Its test RMSE was 0.5020 mg m\textsuperscript{-3}, with a 95\% daily-block interval of 0.4353--0.5688; MAE was 0.3574 (0.3135--0.4062), and R\textsuperscript{2} was 0.8480 (0.7971--0.8868). Mean bias was \ensuremath{-}0.1571 mg m\textsuperscript{-3}, indicating underprediction on average. The paired RMSE reduction against the mean baseline was 0.8039 mg m\textsuperscript{-3} (0.7014--0.9030). Figures 41--43 report the frozen-model scores, selected-model prediction agreement and humidity-dependent residual patterns.

\FloatBarrier

\begingroup

\fontsize{9.5}{11.2}\selectfont

\renewcommand{\arraystretch}{1.16}

\setlength{\tabcolsep}{5pt}

\begin{longtable}{>{\raggedright\arraybackslash}p{\dimexpr 0.27\linewidth-2\tabcolsep\relax}>{\raggedright\arraybackslash}p{\dimexpr 0.2\linewidth-2\tabcolsep\relax}>{\raggedright\arraybackslash}p{\dimexpr 0.18\linewidth-2\tabcolsep\relax}>{\raggedright\arraybackslash}p{\dimexpr 0.18\linewidth-2\tabcolsep\relax}>{\raggedright\arraybackslash}p{\dimexpr 0.17\linewidth-2\tabcolsep\relax}}
\caption{Validation and frozen-test comparison on external UCI data.}\label{tab:4}\\
\toprule
\textbf{Model} & \textbf{Validation RMSE} & \textbf{Test RMSE} & \textbf{Test MAE} & \textbf{Test R\textsuperscript{2}} \\
\midrule
\endfirsthead
\multicolumn{5}{l}{\small Table \thetable\ continued}\\
\toprule
\textbf{Model} & \textbf{Validation RMSE} & \textbf{Test RMSE} & \textbf{Test MAE} & \textbf{Test R\textsuperscript{2}} \\
\midrule
\endhead
\midrule
\multicolumn{5}{r}{\footnotesize Continued on the next page}\\
\endfoot
\bottomrule
\endlastfoot
Training mean & 1.5136 & 1.3059 & 1.0463 & -0.0283 \\
Ridge* & 0.6853 & 0.5020 & 0.3574 & 0.8480 \\
Random forest & 0.6900 & 0.4871 & 0.3194 & 0.8569 \\
Gradient boosting & 0.8202 & 0.5651 & 0.3910 & 0.8075 \\
\end{longtable}

\endgroup

*Ridge was selected by validation RMSE before test evaluation. MAE and RMSE are in mg/m\textsuperscript{3}; R\textsuperscript{2} is dimensionless. All test scores use the same 1,102 held-out observations.

The selected forest used unrestricted depth and a minimum leaf size of 15. Its test RMSE was 0.4871 mg m\textsuperscript{-3}, slightly below Ridge, but this later ranking does not alter the locked choice. Boosting selected learning rate 0.05, 31 leaves and zero L2 penalty; its test RMSE was 0.5651 mg m\textsuperscript{-3}. The mean baseline returned RMSE 1.3059 mg m\textsuperscript{-3} and R\textsuperscript{2}=\ensuremath{-}0.0283. All families remain reported. Overlapping bootstrap intervals are not interpreted as a formal equivalence test, and aggregate error alone does not establish adequate performance during uncommon high-concentration events.

The plots cover the full test date range; the daily trace includes days with at least 12 eligible observations, while the hourly agreement plot and reported scores use all 1,102 eligible test observations. Humidity-conditioned residual plots use fixed 10-percentage-point bins with at least 20 observations; displayed ranges are interquartile spreads, not confidence intervals. Validation and test mean CO were 2.2337 and 1.9329 mg m\textsuperscript{-3}, respectively; corresponding mean RH values were 53.79\% and 51.49\%. These distribution differences may contribute to the lower test errors, although the present analysis does not isolate their effects. R\textsuperscript{2} also depends on target variance and should accompany absolute error.

\begin{figure}[!htbp]
\centering
\includegraphics[width=6.800in,height=2.211in,keepaspectratio]{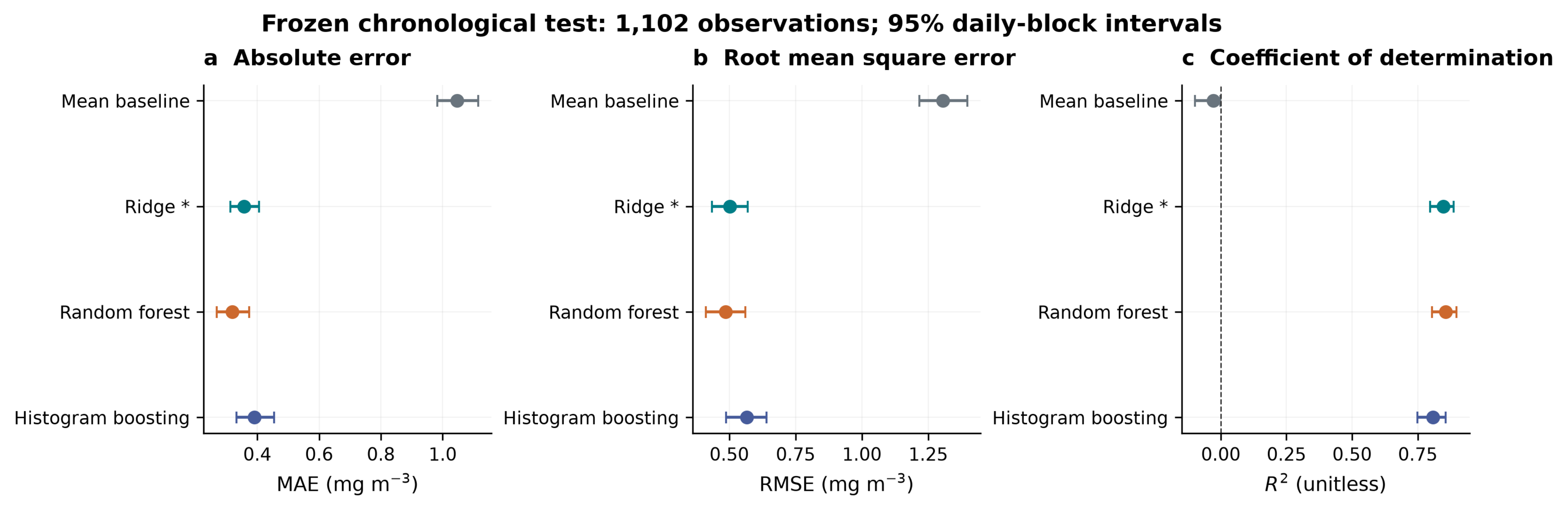}
\caption{Frozen-model MAE, RMSE and R\textsuperscript{2} on 1,102 external-data test observations. Error bars show conditional 95\% percentile intervals from 2,000 paired calendar-day bootstrap resamples. The asterisk identifies validation-selected Ridge. Original analysis of Vito \citep{vito2008_026a72}, CC BY 4.0.}
\label{fig:41}
\end{figure}

\begin{figure}[!htbp]
\centering
\includegraphics[width=6.210in,height=4.320in,keepaspectratio]{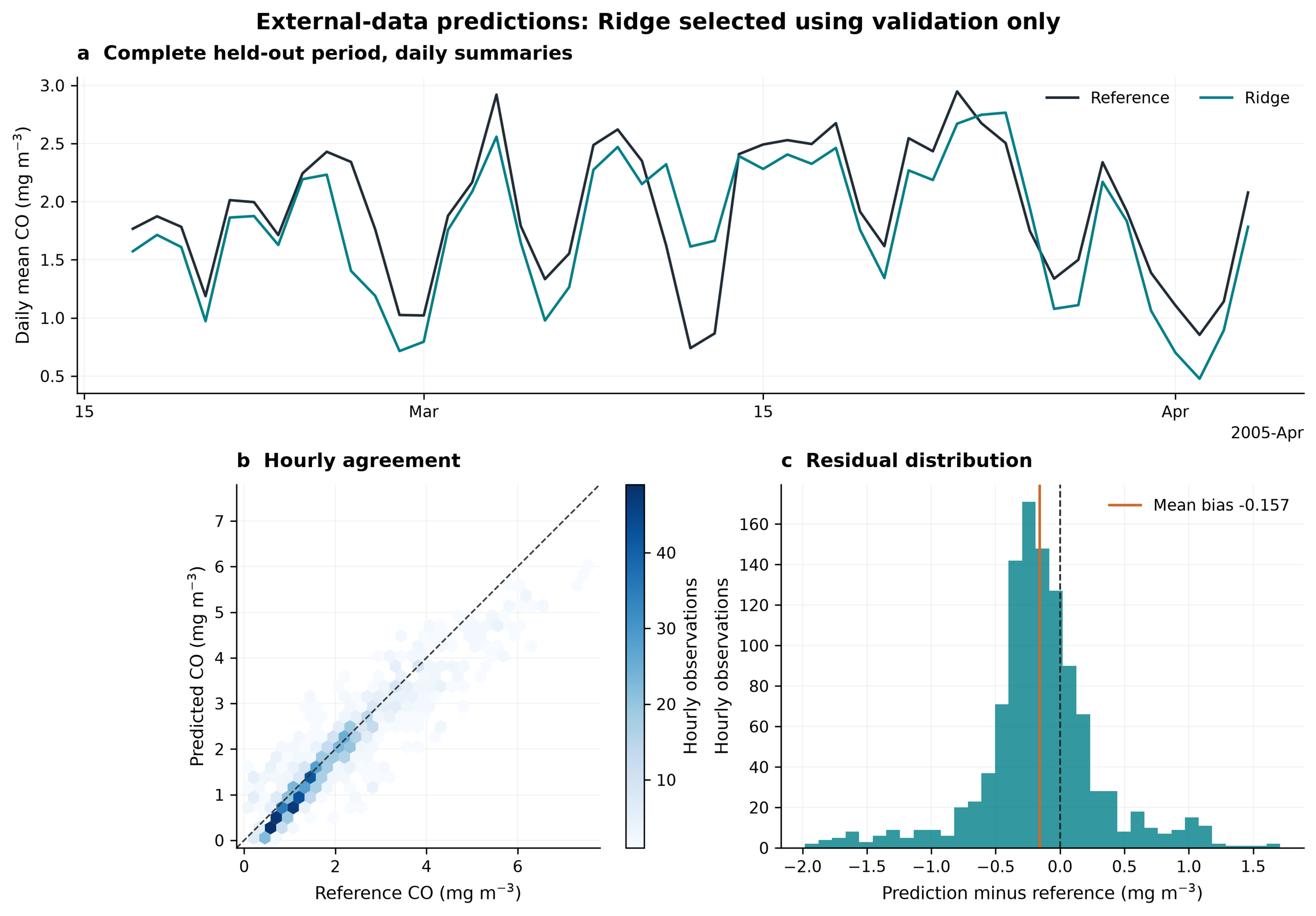}
\caption{Validation-selected Ridge predictions on external test data: (a) daily reference and predicted CO means, requiring at least 12 observed hours; (b) hourly agreement counts in hexagonal bins and the identity line; (c) prediction-minus-reference residuals, with mean bias \ensuremath{-}0.157 mg/m\textsuperscript{3}. Scores use hourly observations. Original analysis of Vito \citep{vito2008_026a72}, CC BY 4.0.}
\label{fig:42}
\end{figure}

\begin{figure}[!htbp]
\centering
\includegraphics[width=6.800in,height=4.472in,keepaspectratio]{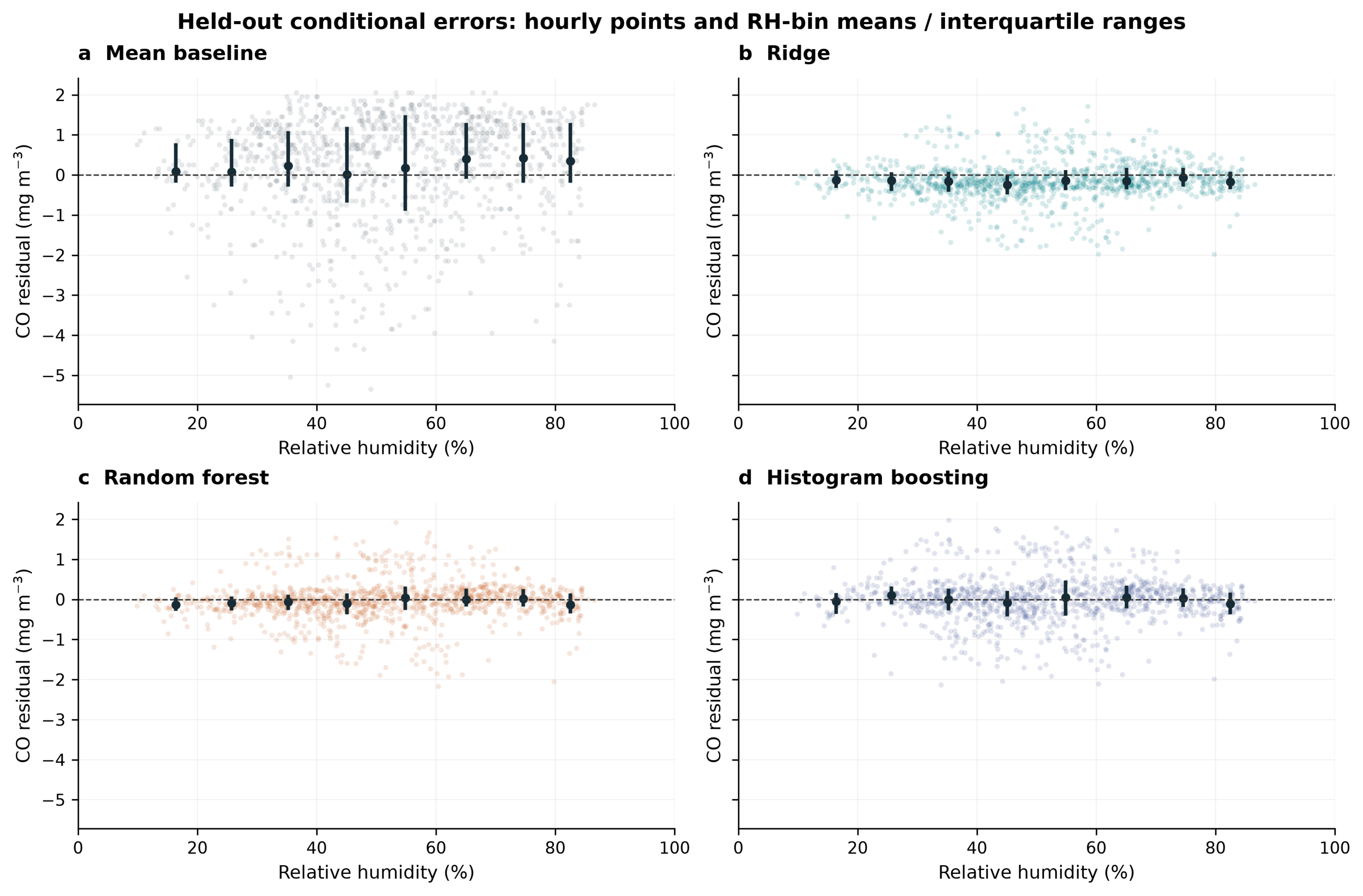}
\caption{Hourly external-test CO residuals against relative humidity. Markers show means within fixed 10-percentage-point RH bins containing at least 20 observations, positioned at median bin RH; bars show interquartile ranges. Axes use common limits. The associations do not establish humidity causality. Original analysis of Vito \citep{vito2008_026a72}, CC BY 4.0.}
\label{fig:43}
\end{figure}

Eligible observations covered 75.41\% of hours within the training span, 78.05\% within validation and 97.61\% within testing. This change matters because the scored populations differ in both season and data availability. Hourly scores weight each available observation equally, whereas a displayed daily average summarizes a varying number of measurements. The day bootstrap resamples observed clusters without manufacturing unrecorded hours; it cannot reconstruct pollution episodes lost during sensor or reference outages. Consequently, the manuscript reports the cleaning audit and coverage alongside predictive performance. A later rover study should document acquisition failures prospectively and evaluate whether missingness is associated with humidity, concentration or actuator operation before adopting the same eligibility rule.

\FloatBarrier

\subsection{Computational cost, reproducibility and transfer limits}\label{sec:10.5}

Desktop measurements included preprocessing and used one-worker settings with one-thread BLAS/OpenMP execution. The recorded environment comprised Python 3.12.14 and scikit-learn 1.9.1 on Windows 11, with processor identifier Intel64 Family 6 Model 183 Stepping 1. Each model received three full-batch warm-up predictions before 21 repetitions, each timing a batch prediction and a single-observation prediction. Median single-observation prediction times across 21 warmed repetitions were 0.583 ms for Ridge, 5.819 ms for the forest and 1.707 ms for boosting. Their uncompressed serialized pipelines occupied 1,937, 4,899,034 and 920,233 bytes respectively. These measurements describe the current desktop implementation, not embedded RAM, flash, energy or real-time feasibility. For the validation-selected Ridge model, permutation sensitivity evaluated on validation data over ten repeats was largest for PT08.S2(NMHC): shuffling it increased RMSE by 1.3255 \ensuremath{\pm} 0.0201 mg m\textsuperscript{-3} (mean \ensuremath{\pm} standard deviation over ten permutations). Correlated predictors and distribution disruption preclude interpreting this ranking as chemical specificity or causation. Figure 44 compares serialized model size, desktop prediction cost and validation-only predictor sensitivity.

The archive hash, software versions, source-row assignments, candidate scores, frozen models, predictions and bootstrap distributions are retained. Reloaded artifacts reproduce predictions within an absolute tolerance of 10\textsuperscript{-12}. Nevertheless, the experiment uses historical fixed-station sensors with different materials and electronics from the proposed MQ array. Excluding unavailable targets and sensor outages can also introduce selection bias. Transfer to A3P5 therefore requires rover-specific reference collocation, heater-aware acquisition, independent later deployments and deployment-hardware measurements. The present contribution is an auditable calibration benchmark and a testable development route, not demonstrated autonomous environmental accuracy.

\begin{figure}[!htbp]
\centering
\includegraphics[width=6.800in,height=2.815in,keepaspectratio]{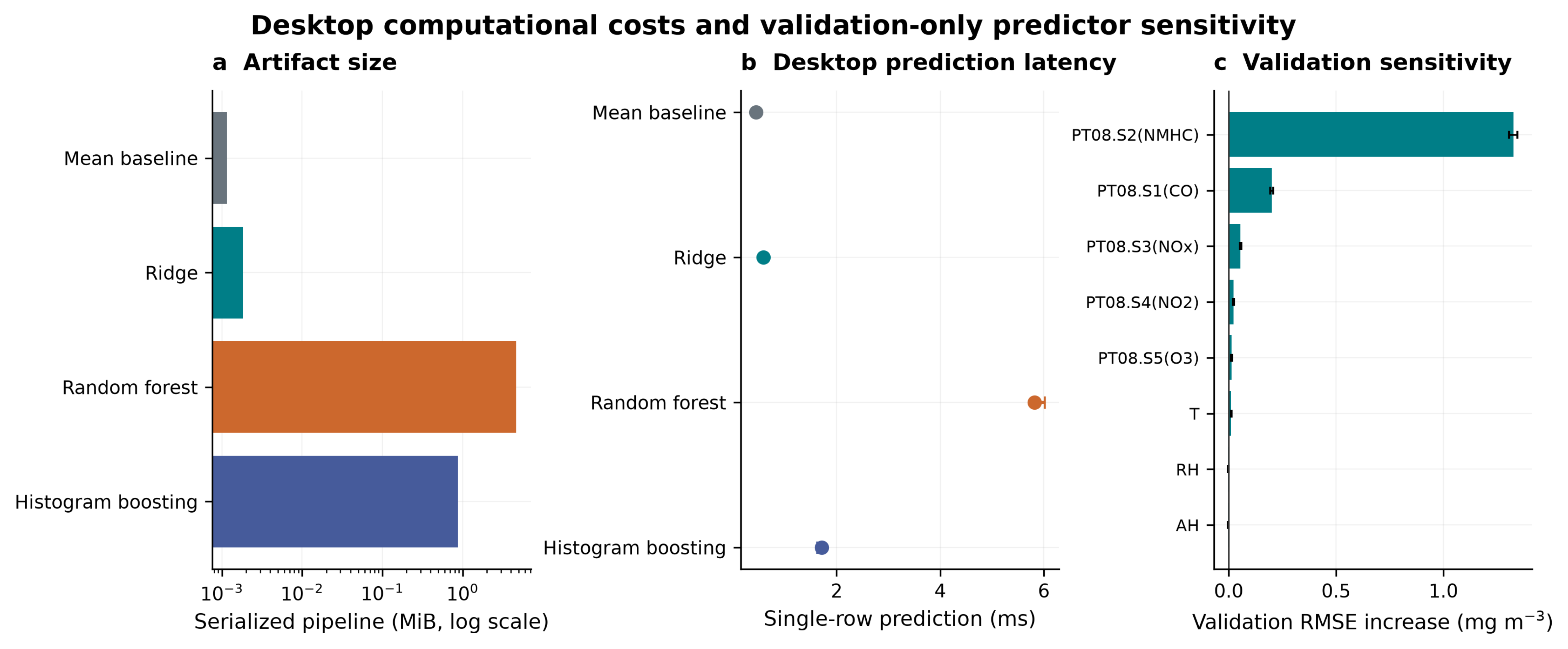}
\caption{External-data model costs and predictor sensitivity: (a) uncompressed pipeline size in MiB; (b) median single-row desktop prediction time and interquartile range over 21 warmed repetitions; (c) validation-selected Ridge RMSE increase after predictor permutation on validation data, with mean and standard deviation over ten repeats. Desktop costs do not establish embedded feasibility. Original analysis of Vito \citep{vito2008_026a72}, CC BY 4.0.}
\label{fig:44}
\end{figure}

\FloatBarrier

\section{Integrated assessment and experimental validation}\label{sec:11}

\FloatBarrier

\subsection{Design tradeoffs exposed by the calculations}\label{sec:11.1}

The calculations show that the most important unknowns are not interchangeable. Mass and rolling resistance affect traction demand; centre-of-mass height affects stability; auxiliary power affects endurance during slow observation; and response time affects the spatial interpretation of moving measurements. A more capable processor may improve perception while reducing runtime. A taller mast may improve visibility while increasing structural loads and vibration sensitivity. An arm redesign that increases reach or allowable payload may also increase moving mass and narrow the support margin. Figure 45 quantifies the sensitivity of the energy and stability calculations to their assumed inputs.

These tradeoffs favour staged development. First establish reliable manual motion, raw sensing and explicit fault handling. Then identify the mechanical and electrical parameters needed for feedback control. Next calibrate environmental channels against appropriate references. Finally evaluate autonomy within the measured mobility and sensing envelope. Adding learning before these foundations increases apparent sophistication while leaving dominant uncertainties unresolved.

\begin{figure}[!htbp]
\centering
\includegraphics[width=6.307in,height=2.800in,keepaspectratio]{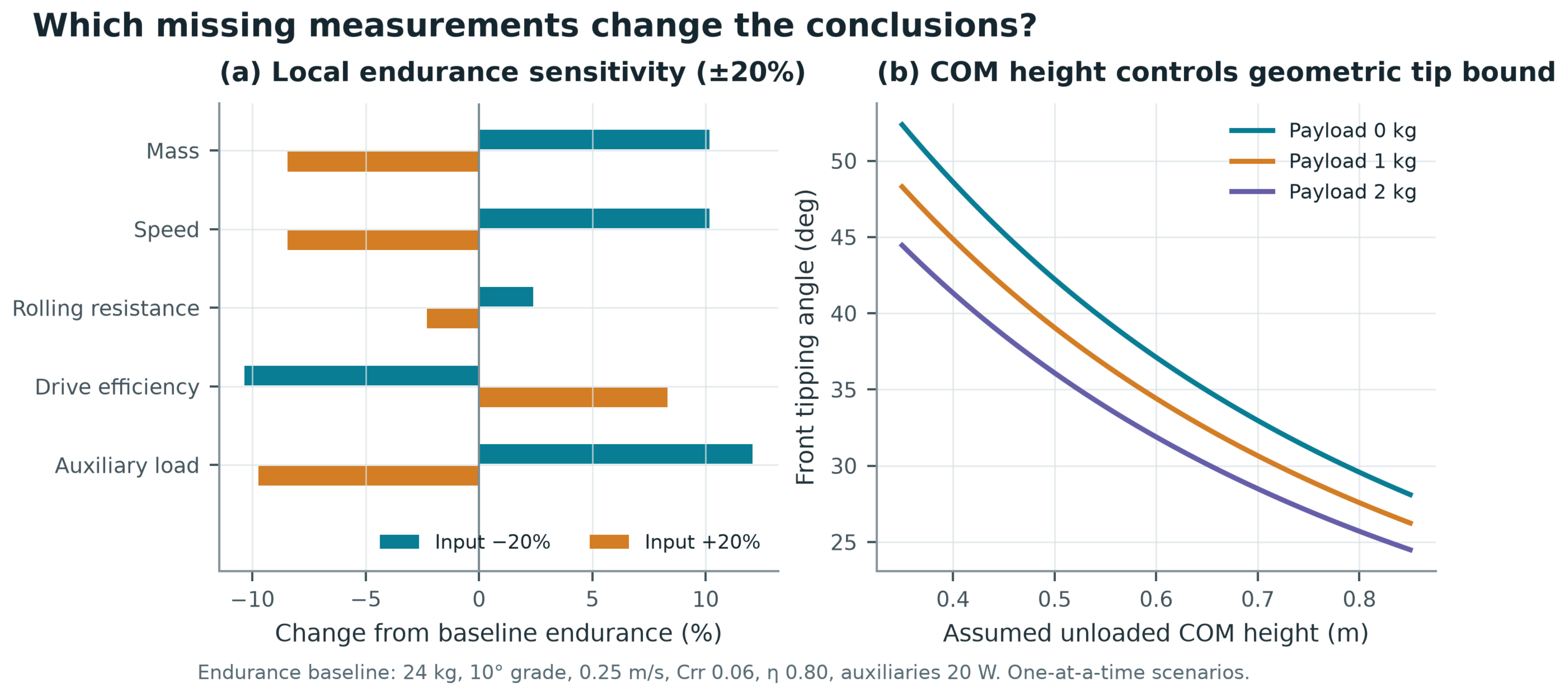}
\caption{Deterministic sensitivity analysis of assumed energy and stability inputs. Panel (a) changes one energy-model input at a time by \ensuremath{\pm}20\%; panel (b) varies unloaded centre-of-mass height from 0.35 to 0.85 m for three payload scenarios. These sweeps show model dependence and are not confidence intervals.}
\label{fig:45}
\end{figure}

\FloatBarrier

\begingroup

\fontsize{9.5}{11.2}\selectfont

\renewcommand{\arraystretch}{1.16}

\setlength{\tabcolsep}{5pt}

\begin{longtable}{>{\raggedright\arraybackslash}p{\dimexpr 0.17\linewidth-2\tabcolsep\relax}>{\raggedright\arraybackslash}p{\dimexpr 0.27\linewidth-2\tabcolsep\relax}>{\raggedright\arraybackslash}p{\dimexpr 0.2\linewidth-2\tabcolsep\relax}>{\raggedright\arraybackslash}p{\dimexpr 0.36\linewidth-2\tabcolsep\relax}}
\caption{Design tradeoffs exposed by the calculations.}\label{tab:5}\\
\toprule
\textbf{Assessment} & \textbf{Scenario or evidence} & \textbf{Result} & \textbf{Interpretation} \\
\midrule
\endfirsthead
\multicolumn{4}{l}{\small Table \thetable\ continued}\\
\toprule
\textbf{Assessment} & \textbf{Scenario or evidence} & \textbf{Result} & \textbf{Interpretation} \\
\midrule
\endhead
\midrule
\multicolumn{4}{r}{\footnotesize Continued on the next page}\\
\endfoot
\bottomrule
\endlastfoot
Wheel speed & 0.25 m/s, radius 0.140 m & 17.05 rpm & Ideal output-wheel speed \\
Grade demand & 24 kg, 20\textdegree{}, Crr 0.06 & 3.282 N\ensuremath{\cdot}m/wheel & Equal-sharing output torque \\
Payload effect & 2 kg at (0.8, 0, 1.0) m & 32.66\textdegree{} front-tip bound & Geometric zero-margin limit \\
Energy & 11.1 V, 10 Ah, stated derating & 84.915 Wh & Assumed usable energy \\
Endurance & 0.25 m/s, 20 W auxiliary, 20\textdegree{} & 1.72 h & Conditional model result \\
Sensor response & Assumed time constant 15 s & 34.54 s to 90\% & Requires measured dynamics \\
Stopping & 0.4 m/s, 0.3 s, 0.5 m/s\textsuperscript{2} & 0.430 m & Includes 0.15 m margin \\
External benchmark & Chronological UCI test, n 1,102 & RMSE 0.502 mg/m\textsuperscript{3} & Ridge, not rover accuracy \\
\end{longtable}

\endgroup

\FloatBarrier

\subsection{Mechanical and control identification}\label{sec:11.2}

Validation should begin with an as-built configuration record. Measure total mass, loaded wheel radius, wheelbase, track, steering limits and linkage geometry. Identify centre of mass in several arm poses using a controlled weighing or suspension method. Record battery location, sample load and optional equipment because a single configuration label is insufficient when the mass distribution changes. Replace visual-model assumptions with these measurements before proposing an operating envelope.

Motor characterization should include output speed, current, voltage and temperature under representative continuous loads. Stall torque is not a continuous operating point. Steering tests should measure zero offset, backlash, tracking error, maximum rate and loaded current. Repeat straight, crab and coordinated-turn trajectories on documented surfaces with an independent position reference where possible. Report path error and command realization separately, distinguishing planner error from actuator saturation or slip.

Repeatability is more useful than an isolated successful demonstration. A record should specify surface, slope, payload, battery state, ambient conditions, controller version and trial order. Repeated trials must include failed starts, protective stops and interventions. Trial-level uncertainty is preferable to treating many correlated samples from one run as independent repetitions. The NIST response-robot programme provides a useful framework for controlled task decomposition, without implying compliance before its methods are applied.

\FloatBarrier

\subsection{Environmental and mission validation}\label{sec:11.3}

Each environmental channel needs an appropriate reference and procedure. Gas channels require known exposure conditions, heater state, environmental context and recovery characterization. Particle sensing requires co-location across relevant aerosol and humidity conditions. UV measurements require a validated optical response and conversion. Water assessment requires standards, handling procedures and repeatability checks. Every calibration should be versioned and specify the domain in which uncertainty was assessed.

Paired stationary observations and repeated traverses at several speeds, supported by synchronized fixed-site references, can help separate ambient changes from response lag and motion-associated effects. Additional airflow and dust controls are needed to identify their individual contributions. Reference and rover clocks need synchronization, and the uncertainty of assigning readings to positions should be reported. A detailed pollution map is not stronger evidence than its timing and calibration support. Unmeasured regions should remain identified as such rather than be filled by unqualified interpolation.

\FloatBarrier

\begingroup

\fontsize{9.5}{11.2}\selectfont

\renewcommand{\arraystretch}{1.16}

\setlength{\tabcolsep}{5pt}

\begin{longtable}{>{\raggedright\arraybackslash}p{\dimexpr 0.25\linewidth-2\tabcolsep\relax}>{\raggedright\arraybackslash}p{\dimexpr 0.29\linewidth-2\tabcolsep\relax}>{\raggedright\arraybackslash}p{\dimexpr 0.46\linewidth-2\tabcolsep\relax}}
\caption{Environmental and mission validation.}\label{tab:6}\\
\toprule
\textbf{Validation stage} & \textbf{Measurements and controls} & \textbf{Primary outcomes} \\
\midrule
\endfirsthead
\multicolumn{3}{l}{\small Table \thetable\ continued}\\
\toprule
\textbf{Validation stage} & \textbf{Measurements and controls} & \textbf{Primary outcomes} \\
\midrule
\endhead
\midrule
\multicolumn{3}{r}{\footnotesize Continued on the next page}\\
\endfoot
\bottomrule
\endlastfoot
As-built identification & Mass, CG, geometry, steering limits, cable sweep & Parameters and configuration uncertainty \\
Drivetrain & Loaded speed, current, temperature, surface & Continuous operating region, slip, saturation \\
Feedback mobility & Repeated paths, independent pose reference & Tracking, repeatability, interventions \\
Stopping and faults & Speed, delay, braking, link loss, low voltage & Stop distance, detection time, safe-state outcome \\
Sensor co-location & Reference, environmental range, calibration ID & Bias, precision, drift, uncertainty, validity domain \\
Manipulation & Known payload, reach, grip, support margin & Grasp/place results, contact forces, recovery \\
Integrated missions & Defined completion and abort criteria & Valid observations, runtime, failures \\
Autonomy & Static/dynamic obstacles, degraded localization & Completion, clearance, recovery, takeover \\
\end{longtable}

\endgroup

\FloatBarrier

\subsection{Threats to validity and limitations}\label{sec:11.4}

The largest limitation is the absence of measured integrated rover performance. Geometry is estimated, mass and energy scenarios are assumed, and rendered tasks demonstrate arrangement rather than operation. The literature census is not exhaustive and depends on database coverage and metadata availability. Selected references support engineering context but do not establish a systematic comparison against every platform. Reused figures preserve their study context and are not pooled into a cross-platform ranking.

The models omit coupled effects. Unequal wheel loading, soft-soil interaction, suspension dynamics, motor thermal limits and battery sag are absent from the simple grade and endurance equations. Stability holds the base centre of mass fixed while adding payload, omitting actual arm-link movement. The response model is first-order and normalized, omitting chemical selectivity and asymmetric recovery. These are screening models for prioritizing measurements; a more detailed simulation would still require identified parameters and experimental comparison.

The public-data benchmark uses a different sensor array, historical deployment and sampling process from NEMESIS. Chronological testing is informative for temporal transfer within that dataset, but does not assess transfer to another device, site or gas-sensor model. The bootstrap interval reflects resampling over observed test days rather than every source of calibration uncertainty. These limits guide subsequent experiments and cannot be condensed into a single general accuracy claim.

\FloatBarrier

\subsection{Research contribution and next experiments}\label{sec:11.5}

The study produces falsifiable engineering expectations. Controlled wheel-torque measurements on documented grades and surfaces can assess the combined force-demand model; isolating rolling resistance also requires control of acceleration, loading and drivetrain losses. Measuring centre-of-mass movement tests the static envelope. Logging power during dwell and motion replaces the constant-load budget. Co-location tests whether environmental corrections reduce error on later days and whether their validity transfers between sites. Each experiment connects to a current uncertainty and a saved analytical baseline.

A paper reporting subsequent field trials should retain this provenance structure while replacing assumptions with measurements where justified. It should state the tested configuration, release raw data or an accessible justified subset, describe exclusions and report negative outcomes. The design can then support an experimental claim about integrated inspection. Until that evidence exists, the supported claim is a documented architecture, detailed geometric model and reproducible analytical evaluation programme.

\FloatBarrier

\subsection{Reproducibility and project support}\label{sec:11.6}

The accompanying research package contains the literature census, source registry, analytical scripts and tables, benchmark outputs, original graphs, Canva design links and Blender mission-study source. The public dataset remains attributable to its repository and authors. Published comparison and empirical figures are reused under the licenses in their individual captions and the figure ledger. Prototype photographs were supplied for this study. No newly collected human-participant or field environmental data are reported.

\textbf{Author contributions.} Shafi Bin Sultan led engineering planning, project coordination, supervision and fundraising, and supported Robot Operating System (ROS) and related system integration. Sabik Bin Sultan and Safwan Sadad developed the robotic platform under his guidance and mentorship. All three authors contributed to project collaboration and fundraising activities.

\textbf{Funding and acknowledgements.} Project support totalled US\$5,000, comprising US\$3,000 from Prime Now and US\$2,000 from Mercedes-Benz Bangladesh. The authors acknowledge their project collaborations with Mercedes-Benz Bangladesh and Prime Bank.

\textbf{Correspondence.} Shafi Bin Sultan, St. Joseph Higher Secondary School; shafibinsultan0207@gmail.com.

\FloatBarrier

\section{Conclusion}\label{sec:12}

This study establishes a reproducible engineering framework for A3P5 NEMESIS by connecting the photographed rover to a detailed geometric reconstruction, subsystem interfaces, mission arrangements and traceable analytical results. The design retains the four independently steered wheel assemblies, carbon-pattern enclosure, folded manipulator, inclined camera mast and side-mounted sampling equipment. Managed wiring and explicit control, sensing and protection interfaces make the proposed integration inspectable. The prototype photographs support the physical arrangement; the rendered missions and analytical models define candidate operating conditions that require experimental confirmation.

The declared 24 kg scenario requires 3.28 N\ensuremath{\cdot}m of output torque per wheel on a 20\textdegree{} grade under equal load sharing. A 2 kg forward payload reduces the modelled front-tipping bound from 38.1\textdegree{} to 32.7\textdegree{}, while the stated battery and load assumptions yield 1.72 h of endurance on that grade. These results identify measurements that most directly constrain mobility, manipulation and observation. Separately, chronological evaluation on the external sensor dataset selected ridge regression by validation error and produced a test RMSE of 0.502 mg/m\textsuperscript{3}, with a 95\% daily-block bootstrap interval of 0.435--0.569 mg/m\textsuperscript{3}. The associated code preserves training-only preprocessing and a frozen test partition. This benchmark demonstrates a reproducible calibration procedure rather than rover-specific measurement accuracy. As-built parameter identification, reference co-location, fault-response tests and integrated field trials are therefore the next steps needed to convert the proposed inspection architecture into quantified operating performance.

\FloatBarrier


\end{document}